\documentclass{article}

     \usepackage[final,nonatbib]{neurips_2022}

\usepackage[utf8]{inputenc} 
\usepackage[T1]{fontenc}    
\usepackage{hyperref}       
\usepackage{url}            
\usepackage{booktabs}       
\usepackage{amsfonts}       
\usepackage{nicefrac}       
\usepackage{microtype}      
\usepackage{xcolor}         

\usepackage{float}
\usepackage{subfig}
\usepackage{dsfont}
\usepackage{multirow}
\usepackage{makecell}
\usepackage{graphicx}
\usepackage{algorithm}
\usepackage[noend]{algpseudocode}
\usepackage{mathrsfs}
\usepackage{bbm}
\usepackage{enumitem}
\usepackage{mathtools}
\usepackage{amsmath, amsthm, amssymb}
\usepackage{bm}
\usepackage[cal=cm]{mathalfa}
\usepackage{wrapfig}

\newtheorem{theorem}{\textbf{Theorem}}
\newtheorem{lemma}{\textbf{Lemma}}

\theoremstyle{definition}

\newtheorem{problem}{\textbf{Problem}}
\newtheorem{remark}{\textbf{Remark}}

\newtheorem{definition}{\textbf{Definition}}

\let\L\relax

\let\define\relax
\let\t\relax

\DeclareMathOperator{\E}{\mathbb{E}}

\DeclareMathOperator{\M}{\mathcal{M}}

\DeclareMathOperator{\Pc}{\mathcal{P}}
\DeclareMathOperator{\Qc}{\mathcal{Q}}

\DeclareMathOperator{\I}{I}
\DeclareMathOperator{\F}{\mathcal{F}}

\DeclareMathOperator{\R}{\mathcal{R}}
\DeclareMathOperator{\N}{\mathcal{N}}
\DeclareMathOperator{\T}{\mathcal{T}}
\DeclareMathOperator{\D}{\mathcal{D}}
\DeclareMathOperator{\X}{\mathcal{X}}
\DeclareMathOperator{\Y}{\mathcal{Y}}
\DeclareMathOperator{\L}{\mathcal{L}}

\DeclareMathOperator*{\argmax}{arg\,max}

\DeclareMathOperator{\define}{\coloneqq}

\DeclareMathOperator{\appro}{approx}
\DeclareMathOperator*{\DE}{DE}
\DeclareMathOperator*{\DC}{DC}

\DeclareMathOperator{\fws}{\overline{w}}
\DeclareMathOperator{\pws}{\widetilde{w}}

\DeclareMathOperator{\fw}{\overline{\vec{w}}}
\DeclareMathOperator{\pw}{\widetilde{\vec{w}}}

\newcommand\norm[1]{\left\lVert#1\right\rVert}
\renewcommand{\vec}[1]{\mathbf{#1}}

\newcommand{\s}{\text{\textcircled{s}}}
\newcommand{\t}{\text{\textcircled{t}}}
\newcommand{\KR}{R_K}

\newcommand{\dal}{\overline{\alpha}}
\newcommand{\ial}{\widetilde{\alpha}}

\renewcommand{\arraystretch}{1.3} 

\title{Social-Inverse: Inverse Decision-making of Social Contagion Management with Task Migrations}

\author{%
  Guangmo~Tong \\
  Department of Computer and Information Science\\
  University of Delaware\\
  \texttt{amotong@udel.edu} \\
}

\begin{document}

\maketitle

\begin{abstract}
Considering two decision-making tasks $A$ and $B$, each of which wishes to compute an effective \textit{decision} $Y$ for a given \textit{query} $X$, {can we solve task $B$ by using query-decision pairs $(X, Y)$ of $A$ without knowing the latent decision-making model?} Such problems, called \textit{inverse decision-making with task migrations}, are of interest in that the complex and stochastic nature of real-world applications often prevents the agent from completely knowing the underlying system. In this paper, we introduce such a new problem with formal formulations and present a generic framework for addressing decision-making tasks in social contagion management. On the theory side, we present a generalization analysis for justifying the learning performance of our framework. In empirical studies, we perform a sanity check and compare the presented method with other possible learning-based and graph-based methods. We have acquired promising experimental results, confirming for the first time that it is possible to solve one decision-making task by using the solutions associated with another one.
\end{abstract}

\section{Introduction}
\label{sec: intro}
\textbf{Social contagion management.} {Social contagion}, in its most general sense, describes the diffusion process of one or more information cascades spreading between a set of atomic entities through the underlying network \cite{christakis2013social,burt1987social,hodas2014simple,iacopini2019simplicial}. Prototypical applications of social contagion management include deploying advertising campaigns to maximize brand awareness \cite{de2008multi,subramani2003knowledge}, broadcasting debunking information to minimize the negative impact of online misinformation \cite{tong2017efficient,yan2019rumor,del2016spreading}, HIV prevention for homeless youth \cite{yadav2018bridging,young2020modeling}, and the prevention of youth obesity \cite{oostenbroek2021link,ricepeer}. In these applications, a central problem is to launch new information cascades in response to certain input queries, with the goal of optimizing the agents' objectives \cite{aral2011commentary,aral2011creating}. In principle, most of these tasks fall into either \textit{diffusion enhancement}, which seeks to maximize the influence of the to-be-generated cascade (e.g., marketing campaign \cite{de2008multi,duaniasua2010mechanisms, chen2010scalable} and public service announcement \cite{oostenbroek2021link,seymour2018impact, valente2007identifying}), or \textit{diffusion containment}, which aims to generate positive cascades to minimize the spread of negative cascades (e.g., misinformation \cite{budak2011limiting,farajtabar2017fake, he2012influence} and violence-promoting messages \cite{fagan2007social, knight2005social}).

\textbf{Inverse decision-making with task migrations.} Traditional research on social contagion management often adopts classic operational diffusion models with known parameters, and focuses on algorithmic development in overcoming the NP-hardness \cite{kempe2003maximizing, chen2010scalable, chen2009efficient,goyal2011celf++,tang2014influence}. However, real-world contagions are often very complicated, and therefore, perfectly knowing the diffusion model is less realistic \cite{du2014influence, he2016learning,goyal2010learning,kalimeris2018learning,gomez2016influence}. When presented with management tasks defined over unknown diffusion models, one can adopt the learn-and-optimize approach in which modeling methodologies and optimization schemes are designed separately; in such methods, the main issue is that the learning process is guided by model accuracy but not by the optimization effect \cite{li2019disco,ko2020monstor}, suggesting that the endeavors dedicated to model construction are neither necessary nor sufficient for successfully handling the downstream optimization problems. This motivates us to explore unified frameworks that can shape the learning pipeline towards effective approximations. Recently, it has been shown that for contagion management tasks like diffusion containment, it is possible to produce high-quality decisions for future queries by using query-decision pairs from the same management task without learning the diffusion model \cite{tong2020stratlearner}. Such findings point out an interesting and fundamental question: with a fixed latent diffusion model, can we solve a target management task by using query-decision pairs from a different management task? This is of interest because the agents often simultaneously deal with several management tasks while it is less likely that they always have the proper empirical evidence concerning the target task. For example, a network manager may need to work on a rumor blocking task, but they only have historical data collected in solving viral marketing tasks. We call such a setting as \textit{inverse decision-making with task migrations}. 

\textbf{Contribution.} This paper presents a formal formulation of inverse decision making where the target task we wish to solve is different from the source task that generates samples, with a particular focus on social contagion management tasks. Our main contribution is a generic framework, called \textit{Social-Inverse}, for handling migrations between tasks of diffusion enhancement and diffusion containment.  For Social-Inverse, we present theoretical analysis to obtain insights regarding how different contagion management tasks can be subtly correlated in order for samples from one task to help the optimization of another task. In empirical studies, we have observed encouraging results indicating that our method indeed works the way it is supposed to. Our main observations suggest that the task migrations are practically manageable to a satisfactory extent in many cases. In addition, we also explore the situations where the discrepancy between the target task and the source task is inherently essential, thereby making the samples from the source task less useful.

\textbf{Roadmap.} In Sec. \ref{sec: pre}, we first provide preliminaries regarding social contagion models, and then discuss how to formalize the considered problem. The proposed method together with its theoretical analysis is presented in Sec. \ref{sec: theory}. In Sec. \ref{sec: exp}, we present our empirical studies. We close our paper with a discussion on limitations and future works (Sec. \ref{sec: further}). The technical proofs, source code, pre-train models, data, and full experimental analysis can be found in the supplementary material. The data and source code is maintained online\footnote{\url{https://github.com/cdslabamotong/social_inverse}}.

\section{Preliminaries}
\label{sec: pre}

\subsection{Stochastic diffusion model}
\label{subsec: model}
A social network is given by a directed graph $G=(V,E)$, with $V$ and $E$ respectively denoting the user set and the edge set. In modeling the contagion process, let us assume that there are $L \in \mathbb{Z}$ information cascades $\{C_i\}_{i=1}^{L}$, each of which is associated with a seed set $S_i \subseteq V$. Without loss of generality, we assume that $S_i\cap S_j = \emptyset$ for $i \neq j$. A diffusion model $\M$ is governed by two sets of configurations: each node $u \in V$ is associated with a distribution $\N_u$ over $2^{N_u^-}$, where $N_u^-$ is the set of the in-neighbors of $u$; each edge $(u,v) \in E$ is associated with a distribution $\T_{(u,v)}$ over $(0,+\infty)$ denoting the transmission time. During the diffusion process, a node can be inactive or $C_i$-active if activated by cascade $C_i$. Given the seed sets, the diffusion process unfolds as follows:
\begin{itemize}
	\item  \textbf{Initialization}: Each node $u$ samples a subset $A_u \subseteq N_u^-$ following $\N_u$, and each edge $(u,v)$ samples a real number $t_{(u,v)} \geq 0 $ following $\T_{(u,v)}$. 
	
	\item \textbf{Time $0$}: The nodes in $S_i$ become $C_i$-active at time $0$, and other nodes are inactive.
	
	\item \textbf{Time $t$}: When a node $u$ becomes $C_i$-active at time $t$, for each inactive node $v$ such that $u$ is in $A_v$, $v$ will be activated by $u$ and become $C_i$-active at time $t+t_{(u,v)}$. Each node will be activated by the first in-neighbor attempting to activate them and never deactivated. When a node $v$ is activated by two or more in-neighbors at the same time, $v$ will be activated by the cascade with the smallest index. 
	
\end{itemize}

\begin{remark}
The considered model is in general highly expressive because $\N_u$ and $\T_{(u,v)}$ can be flexibly designed. For sample, it subsumes the classic independent cascade model \cite{kempe2003maximizing} by making $\N_u$  sample each in-neighbor independently. When there is only one or two cascades, the above model generalizes a few popular diffusion models, including discrete-time independent cascade model \cite{kempe2003maximizing}, discrete-time linear threshold model \cite{kempe2003maximizing}, continuous-time independent cascade model \cite{du2013scalable}, and independent multi-cascade model \cite{tong2021multi, tong2020stratlearner}.
\end{remark}


\begin{definition}[\textbf{Realization}]
\label{def: realization}
Notice that the initialization phase essentially samples a weighted subgraph, and the diffusion process becomes deterministic after the initialization phase. For an abstraction, we call each of such weighted subgraph a \textit{realization}, and use $\R_G$ to denote the space of weighted subgraphs of $G$.	With the concept of realization, we may abstract a concrete stochastic diffusion model $\M$ as a collection of density functions, i.e., $\M=\{\N_u: u \in V\}\cup \{\T_{(u,v)}: (u,v) \in E \}$.  Slightly abusing the notation, we also use $\M: \R_G \rightarrow [0,1]$ to denote the distribution over $\R_G$ induced by the density functions specified by $\M$. On top of a diffusion model $\M$, the distribution of the diffusion outcome depends on the seed sets of the cascades. An example for illustrating the diffusion process is given in Appendix \ref{sec: diffusion_example}. 
\end{definition}

\subsection{Social contagion management tasks}
\label{subsec: task}
In this paper, we focus on the following two classes of social contagion management tasks.

\begin{problem}[\textbf{Diffusion Enhancement }(\textbf{DE})]
	\label{problem: diff_enhancement}
	Given a diffusion model $\M$ and a set of target users $X \subseteq V$, we consider the single-cascade case and let $f_{\M}^{\DE}(X, Y)$ be the expected number of users in $X$ who are activated by a cascade from seed set $Y \subseteq V$. We would like to find a seed set with at most $k \in \mathbb{Z}$ nodes such that the total influence on $X$ can be maximized, i.e.,
	\begin{equation}
	\argmax_{Y \subseteq V, |Y|\leq k} f_{\M}^{\DE}(X, Y).
	\end{equation}
\end{problem}

\begin{problem}[\textbf{Diffusion Containment} (\textbf{DC})]
	\label{problem: diff_containment}
	Given a diffusion model $\M$, we now consider the situation of competitive diffusion where there is a negative cascade $C_1$ with a seed set $X\subseteq V$ and a positive cascade $C_2$ with a seed set $Y\subseteq V$. Let $f_{\M}^{\DC}(X, Y)$ be the expected number of users who are \textit{not} activated by the negative cascade. Given the diffusion model $\M$ and the seed set $X$ of the negative cascade, we would like to find a seed set $Y$ for the positive cascade with at most $k \in \mathbb{Z}$ nodes such that the impact of the negative cascade can be maximally limited, i.e., 
	\begin{equation}
	\argmax_{Y \subseteq V, |Y|\leq k} f_{\M}^{\DC}(X, Y).
	\end{equation}
\end{problem}

Contagion management tasks like Problems \ref{problem: diff_enhancement} and \ref{problem: diff_containment} might be viewed as decision-making problems aiming to infer an effective decision $Y$ in response to a query $X$ . In such a sense, we may abstract such problems in the following way:

\begin{problem}[\textbf{Abstract Contagion Management Tasks}]
	\label{problem: abstract_problem}
	Given a diffusion model $\M$, an abstract management task $T$ is specified by an objective function $f_{\M}^T(X,Y): 2^V \times 2^V \rightarrow \mathbb{R}$, a candidate space $\X_T \subseteq 2^V$ of the queries, and a candidate space $\Y_T \subseteq 2^V$ of the decisions, where we wish to compute 
	\begin{equation}
	\label{eq: problem_abstract}
	Y_{\M, T, X} \define \argmax_{Y \in \Y_T} f_{\M}^T(X, Y)
	\end{equation}
	for each input query $X \in \X_T$. We assume that $\X_T $ and $\Y_T$ are matroids over $V$, subsuming common constraints such as the cardinality constraint or $k$-partition \cite{oxley2006matroid}.  Since such optimization problems are often NP-hard, their approximate solutions are frequently used, and we denote by $Y_{\M, T, X}^{\alpha}$ an $\alpha$-approximation to Equation \ref{eq: problem_abstract}.
\end{problem}

\begin{definition}[\textbf{Linearity over Kernel Functions}]
	\label{def: linearity}
	In addressing the above management tasks, it is worth noting that the objective function is calculated over the possible diffusion outcomes, which are determined in the initialization phase. Specifically, denoting by $f_r^T(X,Y)$ the objective value projected to a single realization $r \in \R_G$, the objective function can be expressed as 
	\begin{align}
	\label{eq: linearity}
	f_{\M}^T(X, Y) = \int_{r \in \R_G} \M(r) \cdot f_r^T(X,Y) d r.
	\end{align}
	The function $f_r^T(X,Y)$  is called \textit{kernel function}, in the sense that it transforms set structures into real numbers. For example, $f_{r}^{\DE}(X, Y)$ denotes the number of users in $X$ who are activated by a cascade generated from $Y$ in a single realization $r$; $f_{r}^{\DC}(X, Y)$ denotes the number of users in $G$ who are not activated by the negative cascade (generated from $X$) in realization $r$ when the positive cascade spreads from $Y$.
\end{definition}

\subsection{Inverse decision-making of contagion management with task migrations}
\label{subsec: inverse}
Supposing that the diffusion model $\M$ is given, DE and DC are purely combinatorial optimization problems, which have been extensively studied \cite{kempe2003maximizing,budak2011limiting}. In the case that the diffusion model is unknown, inverse decision-making is a potential solution, which seeks to solve contagion management tasks by directly learning from query-decision pairs \cite{tong2021usco}. In particular, with respect to a certain management task $f_{\M}^T$ associated with an unknown diffusion model $\M$, the agent receives a collection of pairs $(X_i,Y_i)$ where $Y_i$ is the optimal/suboptimal solution to maximizing $f_{\M}^T(X_i,Y)$. Such empirical evidence can be mathematically characterized as
\begin{align}
\label{eq: dataset}
S_{\M, T, m}^{\dal}=
\Big\{(X_i,Y_{ \M, T, X_i}^{\dal}): f_{\M}^T(X_i, Y_{ \M, T, X_i}^{\dal}) \geq \dal \cdot \max_{Y\in \Y_T} f_{\M}^T(X_i, Y)\Big\}_{i=1}^m
\end{align}
where $\dal \in (0, 1]$ is introduced to measure the optimality of the sample decisions. For the purpose of theoretical analysis, the ratio $\dal$ may be interpreted as the best approximation ratio that can be achieved by a polynomial-time algorithm under common complexity assumptions (e.g., $NP \neq P$). For DE and DC, we have the best ratio as $1-1/e$ due to the well-known fact that their objective functions are submodular \cite{kempe2003maximizing, budak2011limiting}.  Leveraging such empirical evidence, we wish to solve the same or a different management task for future queries:
\begin{problem}[\textbf{Inverse Decision-making with Task Migrations}]
	\label{problem: main}
	Suppose that there is an underlying diffusion model $\M_{true}$.  Consider two management tasks $T_\s$ and $T_\t$ defined in Problem \ref{problem: abstract_problem}, where $T_\s$ is the source task and $T_\t$ is the target task. With a collection $S_{\M_{true}, T_\s, m}^{\dal_\s}$ of samples concerning the source task $T_\s$ for some ratio $\dal_\s \in (0, 1]$, we aim to build a framework $A: \X_{T_\t} \rightarrow \Y_{T_\t}$ that can make a prediction $A(X)$ for each future query $X$ of the target task $T_\t$. Let $l$ be a loss function $l(X, \hat{Y}): \X_{T_\t} \times \Y_{T_\t} \rightarrow [0,1]$ that measures the desirability of $\hat{Y}$ with respect to $X$. We seek to minimize the generalization error $\L$ with respect to an unknown distribution $\D$ over  $\X_{T_\t}$:
	\begin{equation}
	\L(A,\D,l) \define \E_{X \sim \D} \Big[l(X, A(X)) \Big].
	\end{equation}
\end{problem}
Since $\M_{true}$ and $m$ are fixed, we denote $S_{\M_{true}, T_\s, m}^{\dal_\s}$ as $S_{T_\s}^{\dal_\s}$ for conciseness. We will focus on the case where the source task and the target task are selected from DE and DC.

\begin{remark}
In general, the above problem appears to be challenging because the query-decision pairs of one optimization problem do not necessarily shed any clues on effectively solving another optimization problem. What makes our problem tractable is that the source task and the target task share the same underlying diffusion model $\M_{true}$. With the hope that the query-decision pairs of the source task can identify $\M_{true}$ to a certain extent, we may solve the target task with statistical significance, as evidenced in experiments. In such a sense, our setting is called \textit{inverse} as it implicitly infers the structure of the underlying model from solutions, in contrast to the forward decision-making pipeline that seeks solutions based on given models. 
\end{remark}

\section{Social-Inverse}
\label{sec: theory}
In this section, we present a learning framework called Social-Inverse for solving Problem \ref{problem: main}. Our method is inspired by the classic structured prediction \cite{tsochantaridis2005large} coupled with randomized kernels \cite{tong2021usco}, which may be ultimately credited to the idea of Random Kitchen Sink \cite{rahimi2008weighted}. Social-Inverse starts by selecting an empirical distribution $\M_{em}$ over $\R_G$ and a hyperparameter $K \in \mathbb{Z}$, and then proceeds with the following steps:
\begin{itemize}
	\item \textbf{Hypothesis design.} Sample $K$ iid realizations $\KR=\{r_1,...,r_K\} \subseteq \R_G$ following $\M_{em}$, and obtain the hypothesis space $\F_{\KR} \define \{ H_{\KR, \vec{w}}^{\t}(X,Y) :  \vec{w} =(w_1,...,w_K) \in \mathbb{R}^K\}$ where $H_{\KR, \vec{w}}^{\t}$ is the affine combination of $f_{r}^{T_\t}$ over the realizations in $R_K$:
	{
	\begin{align}
	\label{eq: h_space}
	H_{\KR, \vec{w}}^{\t}(X,Y) \define \sum_{i=1}^K w_i \cdot f_{r_i}^{T_\t}(X,Y).
	\end{align}}
	\item \textbf{Training.} Compute a prior vector $\pw$ using the training set $S_{T_\s}^{\dal_\s}$ (Sec. \ref{subsec: train}), and sample the final parameter $\fw$ from an isotropic Gaussian $Q(\gamma \cdot \pw, \mathcal{I})$ with a mean of $\pw$ scaled by $\gamma$. The selection of $\gamma$ will be discussed in Sec. \ref{subsec: consistency}.
	
	\item \textbf{Inference.} Given a future query $X \in \X_{T_{\t}}$ of the target task $T_\t$, the prediction is made by solving the inference problem over the hypothesis in $\F_{\KR}$ associated with final weight $\fw$:
	\begin{align}
	\label{eq: inference}
	\argmax_{Y \in \Y_{T_\t}} H_{\KR, \fw}^{\t}(X,Y).
	\end{align}
	It is often NP-hard to solve the above inference problem in optimal, and therefore, we assume that an $\ial_\t$-approximation to Equation \ref{eq: inference} -- denoted by $Y_{\KR, \fw, X}^{\ial_\t}$ --   is employed for some  $\ial_\t \in (0,1]$. Notice that the ratio $\ial_\t$ herein represents the inference hardness, while the ratio $\dal_\s$ associated with the training set measures the hardness of the source task.
\end{itemize}
In completing the above procedure, it is left to determine a) the prior distribution $\M_{em}$, b) the hyperparameter $K$, c) the scale factor $\gamma$, d) the training method for computing the prior vector $\pw$, and e) the inference algorithm for computing $Y_{\KR, \fw, X}^{\ial_\t}$. In what follows, we will first discuss how they may influence the generalization performance in theory, and then present methods for their selections. For the convenience of reading, the notations are summarized in Table \ref{table: notation} in Appendix \ref{sec: appendix_notation}.

\subsection{Generalization  analysis}
\label{subsec: consistency}
For Social-Inverse, given the fact the generation of $\fw$ is randomized, the generalization error is further expressed as
\begin{align}
\label{eq: true_error}
\L(\text{Social-Inverse},\D,l) \define \E_{X \sim \D, \fw \sim Q} \Big[l(X, Y_{\KR, \fw, X}^{\ial_\t}) \Big].
\end{align}
In deriving an upper bound with respect to the prior vector $\pw$, let us notice that the empirical risk is given by $l(X_i,Y_{\KR, \fw, X_i}^{\ial_\t})$, which is randomized by $\fw \sim Q(\gamma \cdot \pw, \mathcal{I})$. Thus, the prediction $Y_{\KR,  \fw, X_i}^{\ial_\t}$ associated with a training input $X_i$ is most likely one of those centered around $Y_{\KR, \pw, X_i}^{\ial_\t}$, and we will measure such concentration by their difference in terms of a fraction of the empirical risk associated with $\pw$. More specifically, controlled by a hyperparameter $\beta \in (0, \ial_\t)$, for an input query $X_i$, the potential predictions are those within the margin:
\begin{align}
\label{eq: margin}
& \I_{\KR,  \pw,  X_i,\beta}^{\t} \define  \\ \nonumber
&\Big\{ Y \in \Y_\t:  \ial_\t \cdot H_{\KR, \pw}^{\t}(X_i,Y_{\KR,\pw,X_i}^{\ial_\t}) - H_{\KR, \pw}^{\t}(X_i,Y) \leq \beta \cdot  H_{\KR, \pw}^{\t}(X_i,Y_{\KR,\pw,X_i}^{\ial_\t}) \Big\}.
\end{align}
The empirical risk is therefore given via the above margin:
\begin{align}
\label{eq: em_risk}
\L_{em}(\pw, \KR, S_{T_\s}^{\dal_\s}, l)\define \frac{1}{m}\sum_{i=1}^{m} \max_{Y \in \Y_\t}l(X_i,Y) \cdot  \mathds{1}_{ \I_{\KR,  \pw,  X_i,\beta}^{\t} }(Y),
\end{align}
where $\mathds{1}_S(x) \in \{0,1\}$ is the indicator function: $\mathds{1}_S(x) = 1 \iff x \in S$. With the above progressions, we have the following result concerning the generalization error.

\begin{theorem}
\label{theorem: bound}
For each $\widetilde{\vec{w}}=(\pws_1,...,\pws_K)$, $R_K\subseteq \R_G$, $\beta \in (0, \ial_\t)$, and $\delta>0$, with probability at least $1-\delta$, we have
\begin{align*}
\L(\text{Social-Inverse},\D,l) \leq \L_{em}(\pw, \KR, S_{T_\s}^{\dal_\s}, l) +\frac{\norm{\widetilde{\vec{w}}}^2}{m}+\sqrt{\frac{ \gamma^2\norm{\widetilde{\vec{w}}}^2/2 +\ln (m/\delta)}{2(m-1)}}
\end{align*}
provided that
\begin{align}
\label{eq: gamma}
\gamma = \frac{\ial_\t^2+1}{\min_p |\pws_p| \cdot \beta \cdot  \ial_\t}\sqrt{2\ln \frac{2mK}{\norm{\widetilde{\vec{w}}}^2}}
\end{align}
\end{theorem}

The proof follows from the standard analysis of the PAC-Bayesian framework \cite{mcallester2003simplified} coupled with the approximate inference \cite{wu2017learning} based on a multiplicative margin \cite{tong2021usco}; the extra caution we need to handle is that our margin (Equation \ref{eq: margin}) is parameterized by $\beta$. Notice that when $\beta$ decreases, the regularization term becomes larger, while the margin set $\I_{\KR,  \pw,  X_i,\beta}^{\t}$ becomes smaller --  implying a low empirical risk (Equation \ref{eq: em_risk}). In this regard, Theorem \ref{theorem: bound} presents an intuitive trade-off between the estimation error and the approximation error controlled by $\beta$.

Having seen the result for a general loss, we now seek to understand the best possible generalization performance in terms of the approximation loss $l_{\appro}$:
\begin{align}
\label{eq: app_loss}
l_{\appro}(X, \hat{Y}) \define 
1- \frac{f_{\M_{true}}^{T_{\t}}(X, \hat{Y}) }{\max_{Y \in \Y_\t} f_{\M_{true}}^{T_{\t}}(X, Y) } \in [0,1].
\end{align}
Such questions essentially explore the realizability of the hypothesis space $\F_{R_K}$, which is determined by the empirical distribution $\M_{em}$ and the number of random realizations used to construct $\F_{R_K}$. We will see shortly how these factors may impact the generalization performance. By Theorem \ref{theorem: bound}, when infinite samples are available, the empirical risk approaches to 
\begin{equation}
\E_{X \sim \D} \Big[ \max_{Y \in \Y_{T_\t}} l_{\appro}(X, Y) \cdot  \mathds{1}_{ \I_{\KR,  \pw,  X,  \beta}^{\t} }(Y) \Big].
\end{equation} 
The next result provides an analytical relationship between the complexity of the hypothesis space and the best possible generalization performance in terms of $ l_{\appro}$.  
\begin{theorem}
	\label{theorem: realizability}
	Let $\Delta \define \sup_r \frac{\M_{true}(r)}{\M_{em}(r)} \cdot \frac{\max f_r^{T_\t}(X, Y)}{\min f_r^{T_\t}(X, Y)}$ measure the divergence between $\M_{true}$ and $\M_{em}$ scaled by the range of the kernel function. For each $\epsilon>0, \delta_1 >0, \delta_2 >0$, and $\M_{em}$, when $K$ is $O(\frac{\Delta ^2}{ \epsilon^2 \cdot \delta_2^2}(\ln|\Y_{T_\t}|+\ln \frac{1}{\delta_1}))$, with probability at least $1-\delta_1$ over the selection of $R_K$, there exists a desired weight $\pw$ such that 
	\begin{equation}
	\label{eq: realizability}
	\Pr_{X \sim \D} \Big[ \max_{Y \in \Y_{T_\t}} l_{\appro}(X, Y) \cdot  \mathds{1}_{ \I_{\KR,  \pw,  X,  \beta}^{\t} }(Y) \leq 1 - \frac{ \ial_\t \cdot  (\ial_\t - \beta) \cdot(1-\epsilon)}{(1+\epsilon)} \Big]\geq 1-\delta_2.
	\end{equation}
\end{theorem}

\begin{remark}
	The above result has the implication that the best possible ratio in generalization is essentially bounded by $O(\ial_\t \cdot  (\ial_\t - \beta))$. On the other hand, one can easily see that the target task (Equation \ref{eq: problem_abstract}) and the inference problem (Equation \ref{eq: inference}) suffer the same approximation hardness, and therefore, one would not wish for a true approximation error that is better than $\ial_\t$; in this regard, the result in Theorem \ref{theorem: realizability} is not very loose. 
\end{remark}

The results in this section demonstrate how the selections of $K$, $\M_{em}$, $\pw$, and $\beta$ may affect the generalization performance in theory. Since the true model $\M_{true}$ is unknown, the prior distribution $\M_{em}$ can be selected to be uniform or Gaussian distribution. $K$ and $\beta$ can be taken as hyperparameters determining the model complexity. In addition, since the true loss $l_{\appro}$ is not accessible, one can take general loss functions. Given the fact that we are concerned with set structures rather than real numbers, we employ the zero-one loss, which is adopted also for the convenience of optimization. Therefore, it remains to figure out how to compute the prior vector $\pw$ from training samples as well as how to solve the inference problem (Equation \ref{eq: inference}), which will be discussed in the next part.

\subsection{Training method}
\label{subsec: train}
In computing the prior vector $\pw$, the main challenge caused by the task migration is that the target task on which we performance inference is different from the source task that generates training samples. Theorem \ref{theorem: bound} suggests that, ignoring the low-order terms, one may find the prior vector by minimizing the regularized empirical risk $\L_{em}(\pw, \KR, S_{T_\s}^{\dal_\s}, l) +\frac{\norm{\widetilde{\vec{w}}}^2}{m}$. Directly minimizing such a quantity would be notoriously hard because the optimization problem is bilevel: optimizing over $\pw$ involves the term  $Y_{\KR,\pw,X_i}^{\ial_\t}$ which is obtained by solving another optimization problem depending on $\pw$ (Equations \ref{eq: margin} and \ref{eq: em_risk}). Notably, since $H_{\KR, \pw}^{\t}(X_i,Y_{\KR,\pw,X_i}^{\ial_\t})$ is lower bounded by $\ial_\t \cdot H_{\KR, \pw}^{\t}(X_i, Y_{\M_{true}, T_\t,X_i})$, replacing $Y_{\KR,\pw,X_i}^{\ial_\t}$ with $Y_{\M_{true}, T_\t,X_i}$ would allow for us to optimize an upper bound of the empirical risk. Seeking a large-margin formulation, this amounts to solving the following mathematical program under the zero-one loss \cite{tsochantaridis2005large,taskar2005learning}:  
\begin{align}
\label{eq: programming_1}
& {\text{min}}  & & \norm{\vec{w}}^2+ C\sum_{i=1}^{m}\xi_i / m\nonumber \\
& \text{s.t.}  & & \ial_\t (\ial_\t - \beta) \cdot H_{\KR, \vec{w}}^{\t}(X_i, Y_{\M_{true}, T_\t,X_i}) - H_{\KR, \vec{w}}^{\t}(X_i,Y) \geq \xi_i,~\forall i \in [m], ~\forall ~Y \in \Y_\t \nonumber\\
&  & & \vec{w} \geq 0
\end{align}
where $\xi_i$ is the slack variable and $C$ is a hyperparameter \cite{muller2014pystruct}. However, our dataset concerns only about the source task $T_\s$ without informing $Y_{\M_{true}, T_\t,X_i}$ or its approximation. In order to see where we could feed the training samples into the training process, let us notice that the constraints in Equation \ref{eq: programming_1} have an intuitive meaning: with respect to the target the task $T_\t$, a desired weight $\pw$ should lead to a score function $H_{\KR, \pw}^{\t}$ that can assign highest scores to the optimal solutions $Y_{\M_{true}, T_\t,X_i}$. Similar arguments also apply to the source task $T_\s$, as the weight $\pw$ implicitly estimates the true model $\M_{true}$, which is independent of the management tasks. This enables us to reformulate the optimization problem with respect to the source task $T_\s$ by using the following constraints:
\begin{align}
\label{eq: new_constraint}
\ial_\t (\ial_\t - \beta) \cdot H_{\KR, \vec{w}}^{\s}(X_i, Y_{\M_{true},  T_\s, X_i}^{\dal_\s}) - H_{\KR, \vec{w}}^{\s}(X_i,Y) \geq \xi_i,~\forall i \in [m],  ~\forall Y \in \Y_\t \
\end{align}
where  $H_{\KR, \vec{w}}^{\s}(X,Y) \define \sum_{i=1}^K w_i \cdot f_{r_i}^{T_\s}(X,Y)$ is the score function corresponding to the source task $T_\s$. As desired, pairs of $(X_i, Y_{\M_{true},  T_\s, X_i}^{\dal_\s})$ are the exactly the information we have in the training data $S_{T_\s}^{\dal_\s}$. One remaining issue is that the acquired program (Equation \ref{eq: new_constraint}) has an exponential number of constrains \cite{taskar2005learning}, which can be reduced to linear (in sample size) if the following optimization problem can be solved for each $\vec{w}$ and $X_i$:
\begin{align}
\label{eq: lai}
\max_{Y \in \Y_\s } H_{\KR, \vec{w}}^{\s}(X_i,Y).
\end{align}
Provided that the above problem can be addressed, the entire program can be solved by several classic algorithms, such as the cutting plane algorithm \cite{kelley1960cutting} and the online subgradient algorithm \cite{lucchi2013learning}. Therefore, in completing the entire framework, it remains to solve Equations \ref{eq: inference} and \ref{eq: lai}. For tasks of DE and DC, we delightedly have the following results.
\begin{theorem}
	\label{theorem: algorithm}
	When $T_\s$ and $T_\t$ are selected from DE and DC, Equations \ref{eq: inference} and \ref{eq: lai} are both NP-hard to solve in optimal, but both can be approximated within a ratio of $1-1/e$ in polynomial time. 
\end{theorem}
A concrete example of using Social-Inverse to solve Problem \ref{problem: main} is provided in Appendix \ref{sec: example_si}.

\section{Empirical studies}
\label{sec: exp}
Although some theoretical properties of our framework can be justified (Sec. \ref{subsec: consistency}), it remains open whether or not the proposed method is practically effective, especially given the fact that no prior work has attempted to solve one optimization problem by using the solutions to another one. In this section, we present our empirical studies. 

\subsection{Experimental settings}

We herein present the key logic of our experimental settings and provide details in Appendix \ref{subsec: full_setting}.

\textbf{The latent model $\M_{true}$ and samples (Appendix \ref{subsubsec: appendix_model}).} To generate a latent diffusion model $\M_{true}$, we first determine the graph structure and then fix the distributions $\N_u$ and $\T_{(u,v)}$ by generating random parameters. We adopt four graphs: a Kronecker graph \cite{leskovec2010kronecker}, an Erdős-Rényi graph \cite{hagberg2008exploring}, a Higgs graph \cite{de2013anatomy}, and a Hep graph \cite{leskovec2007graph}). Given the underlying diffusion model $\M_{true}$, for each of DE and DC, we generate a pool of query-decision pairs $(X_i,Y_i)$ for training and testing, where $X_i$ is selected randomly from $V$ and $Y_i$ is the approximate solution associated with $X_i$ (Theorem \ref{theorem: algorithm}). As for Problem \ref{problem: main}, there are four possible target-source pairs: DE-DE, DC-DE, DE-DC, and DC-DC.

\begin{table}[t]
	\renewcommand{\arraystretch}{1.1} 
	\small
	\caption{\textbf{Results on Kronecker (Kro) and Erdős-Rényi (ER).} Each cell shows the mean of performance ratio with std. For Social-Inverse under each empirical distribution, the table shows the results with $K\in \{15,30,60\}$ and $\beta = 1$. The training size is $270$, and the testing size is $540$.}
	\centering
	\label{table: main}
	\begin{tabular}{  @{}  c  @{\hspace{4mm}} c  @{\hspace{4mm}}     l @{\hspace{2mm}}  c  @{\hspace{2mm}} c @{\hspace{2mm}} c  @{\hspace{4mm}} c  @{\hspace{4mm}}   l @{\hspace{2mm}} c @{\hspace{2mm}} c @{\hspace{2mm}} c  @{}  }
		\toprule
		& &  &  K = 15 &  K = 30   & K = 60 & & &  K = 15 &  K = 30   & K = 60 	\\
		\multirow{8}{*}{\textbf{Kro}}   &\multirow{4}{*}{\textbf{\makecell[c]{DC\\|\\DE}}} & $\M_{\infty}$  &	  0.541 {\tiny (0.014)} &      0.567 {\tiny (0.016)} 	   		&   0.584 {\tiny (0.006)}  	& &$\M_{1}$ &  0.748 {\tiny (0.022)}  &     	 0.794 {\tiny (0.016)} 		&    0.811 {\tiny (0.011)} 	\\
		& & $\M_{0.5}$ &  0.767 {\tiny (0.028)}  &    	 0.837 {\tiny (0.004)} 		&    0.851 {\tiny (0.004)} & 	&   $\M_{0.1}$ &  0.770 {\tiny (0.025)}  &     	 0.850 {\tiny (0.018)} 		&    0.853 {\tiny (0.040)}  \\
		\cmidrule{3-11}
		& & \multicolumn{9}{l}{\textbf{NB}: 0.64 {\tiny (0.01)} \hspace{1mm} \textbf{GNN}: 0.51 {\tiny (0.05)} \hspace{1mm} \textbf{DSPN}: 0.46 {\tiny (0.14)} \hspace{1mm} \textbf{HD}: 0.59 {\tiny (0.01)} \hspace{1mm} \textbf{Random}: 0.15 {\tiny (0.01)} }\\
		\cmidrule{2-11}
		
		&\multirow{4}{*}{\textbf{\makecell[c]{DE\\|\\DC}}} & $\M_{\infty}$  &	 0.846 {\tiny (0.005)} &      0.913 {\tiny(0.005)} 	   		&   0.985 {\tiny (0.014)} 	& &$\M_{1}$ &  0.796 {\tiny (0.056)}  &   	 0.845 {\tiny (0.040)} 		&    0.899 {\tiny (0.030)} 	    	\\
		& & $\M_{0.5}$ &  0.850 {\tiny (0.036)}  &     0.937 {\tiny (0.058)} 		&    1.021 {\tiny (0.040)} & 	&   $\M_{0.1}$ &  0.862 {\tiny (0.013)}  &   	 0.953 {\tiny (0.018)} 		&    1.041 {\tiny (0.026)}   \\
		\cmidrule{3-11}
		& & \multicolumn{9}{l}{\textbf{NB}: 0.88 {\tiny (0.01)} \hspace{1mm} \textbf{GNN}: 0.76 {\tiny (0.05)} \hspace{1mm} \textbf{DSPN}: 0.57 {\tiny (0.14)} \hspace{1mm} \textbf{HD}: 0.89 {\tiny (0.01)} \hspace{1mm} \textbf{Random}: 0.26 {\tiny (0.01)} }\\
			\midrule

		\multirow{8}{*}{\textbf{ER}}   &\multirow{4}{*}{\textbf{\makecell[c]{DC\\|\\DE}}} &  $\M_{\infty}$  &	 0.541 {\tiny (0.010)}   &  0.585 {\tiny (0.018)} 	   	   & 0.577 {\tiny (0.007)} 	   	  	& & $\M_{1}$ &  0.744 {\tiny (0.005)}  &  0.752 {\tiny (0.003)} 		     &  0.754 {\tiny (0.003)} 		\\
		& & $\M_{0.5}$ &  0.825 {\tiny (0.005)}  &   	 0.830 {\tiny (0.005)} 	&	  0.830 {\tiny (0.005)}  & 	&   $\M_{0.1}$ &   0.829 {\tiny (0.002)}  & 	 0.833 {\tiny (0.004)} 	 &	0.833 {\tiny (0.004)}	  \\
		\cmidrule{3-11}
		& & \multicolumn{9}{l}{\textbf{NB}: 0.78 {\tiny (0.01)} \hspace{1mm} \textbf{GNN}: 0.05 {\tiny (0.02)} \hspace{1mm} \textbf{DSPN}: 0.06 {\tiny (0.03)} \hspace{1mm} \textbf{HD}: 0.32 {\tiny (0.01)} \hspace{1mm} \textbf{Random}: 0.06 {\tiny (0.01)} }\\
		
		\cmidrule{2-11}
		&\multirow{4}{*}{\textbf{\makecell[c]{DE\\|\\DC}}} & $\M_{\infty}$  &	 0.526 {\tiny (0.020)} &    0.677 {\tiny (0.016)} 	&  0.750 {\tiny (0.010)} 		& &$\M_{1}$ &  0.773 {\tiny (0.016)}  & 	 0.796 {\tiny (0.010)} 	& 	 0.800 {\tiny (0.009)}  	\\
		& & $\M_{0.5}$ &  0.879 {\tiny (0.002)}  &   	 0.900 {\tiny (0.002)} 		     & 0.892 {\tiny (0.002)} & 	&   $\M_{0.1}$ &   0.886 {\tiny (0.005)} &   0.902 {\tiny (0.003)} 	 &  0.895 {\tiny (0.003)} \\
		\cmidrule{3-11}
		& & \multicolumn{9}{l}{\textbf{NB}: 0.04 {\tiny (0.01)} \hspace{1mm} \textbf{GNN}: 0.05 {\tiny (0.02)} \hspace{1mm} \textbf{DSPN}: 0.04 {\tiny (0.01)} \hspace{1mm} \textbf{HD}: 0.09 {\tiny (0.01)} \hspace{1mm} \textbf{Random}: 0.04 {\tiny (0.01)} }\\
		\bottomrule
	\end{tabular}
	\vspace{-3mm}
\end{table}

\textbf{Social-Inverse (Appendix \ref{subsubsec: appendix_SI}).} With $K$ and $\beta$ being hyperparameters, to set up Social-Inverse, we need to specify the empirical distribution $\M_{em}$. We construct the empirical distribution by building three diffusion models $\M_{q}$ with $q \in \{0.1, 0.5, 1\}$, where a smaller $q$ implies that  $\M_{q}$ is closer to $\M_{true}$. In addition, we construct an empirical distribution $\M_{\infty}$ which is totally random and not close to $\M_{true}$ anywhere.  For each empirical distribution, we generate a pool of realizations.

\textbf{Competitors (Appendix \ref{subsubsec: appendix_other}).} Given the fact that Problem \ref{problem: main} may be treated as a supervised learning problem with the ignorance of task migration, we have implemented  Naive Bayes (NB) \cite{rish2001empirical}, graph neural networks (GNN) \cite{kipf2016semi}, and a deep set prediction network ({DSPN}) \cite{zhang2019deep}. In addition, we consider the High-Degree (HD) method, which is a popular heuristic believing that selecting the high-degree users as the seed nodes can decently solve DE and DC. A random (Random) method is also used as the baseline.

\textbf{Training, testing, and evaluation (Appendix \ref{subsubsec: appendix_train})} The testing size is $540$, and the training size $m$ is selected from $\{90, 270, 1350\}$. Given the training size and the testing size, the samples are randomly selected from the pool we generate; similarly, given $K$, the realizations used in Social-Inverse are also randomly selected from the pool we generate. For each method, the entire process is repeated five times, and we report the average performance ratio together with the standard deviation. The performance ratio is computed by comparing the predictions with the decisions in testing samples; larger is better. 


\subsection{Main observations}
The main results on the Kronecker graph and the Erdős-Rényi graph are provided in Table \ref{table: main}. According to Table \ref{table: main}, it is clear that Social-Inverse performs better when $K$ becomes larger or when the discrepancy between $\M_{em}$ and $\M_{true}$ becomes smaller (i.e., $q$ is small), which suggests that Social-Inverse indeed works the way it is supposed to. In addition, while all the methods are non-trivially better than Random, one can also observe that Social-Inverse easily outperforms other methods by an evident margin as long as sufficient realizations are provided. We also see that learning-based methods do not perform well in many cases; this is not very surprising because the effectiveness of learning-based methods hinges on the assumption that different tasks share similar decisions for the same query, which however may not be the case especially on the Erdős-Rényi graph. Furthermore, Social-Inverse appears to be more robust than other methods in terms of standard deviation. Finally, the performance of standard learning methods (e.g., NB and GNN) are sensitive to graph structures; they are relatively good on the Kronecker graph but less satisfactory on the Erdős-Rényi graph, while Social-Inverse is consistently effective on all datasets.

\begin{figure}[t]
	\centering
	\subfloat{\label{fig: lagend}\includegraphics[width=0.4\textwidth]{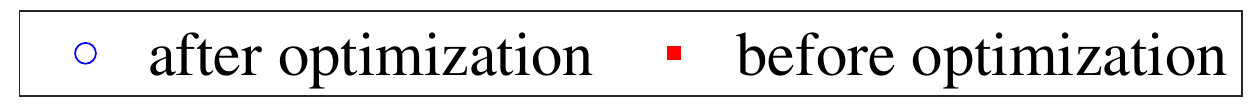}}
	\addtocounter{subfigure}{-1}
	
	\vspace{-2mm}
	\subfloat[{[DC-DE, $ \M_{0.1}$, $m=90$]}]{\label{fig: kro_dc_de_10_30_90}\includegraphics[width=0.31\textwidth]{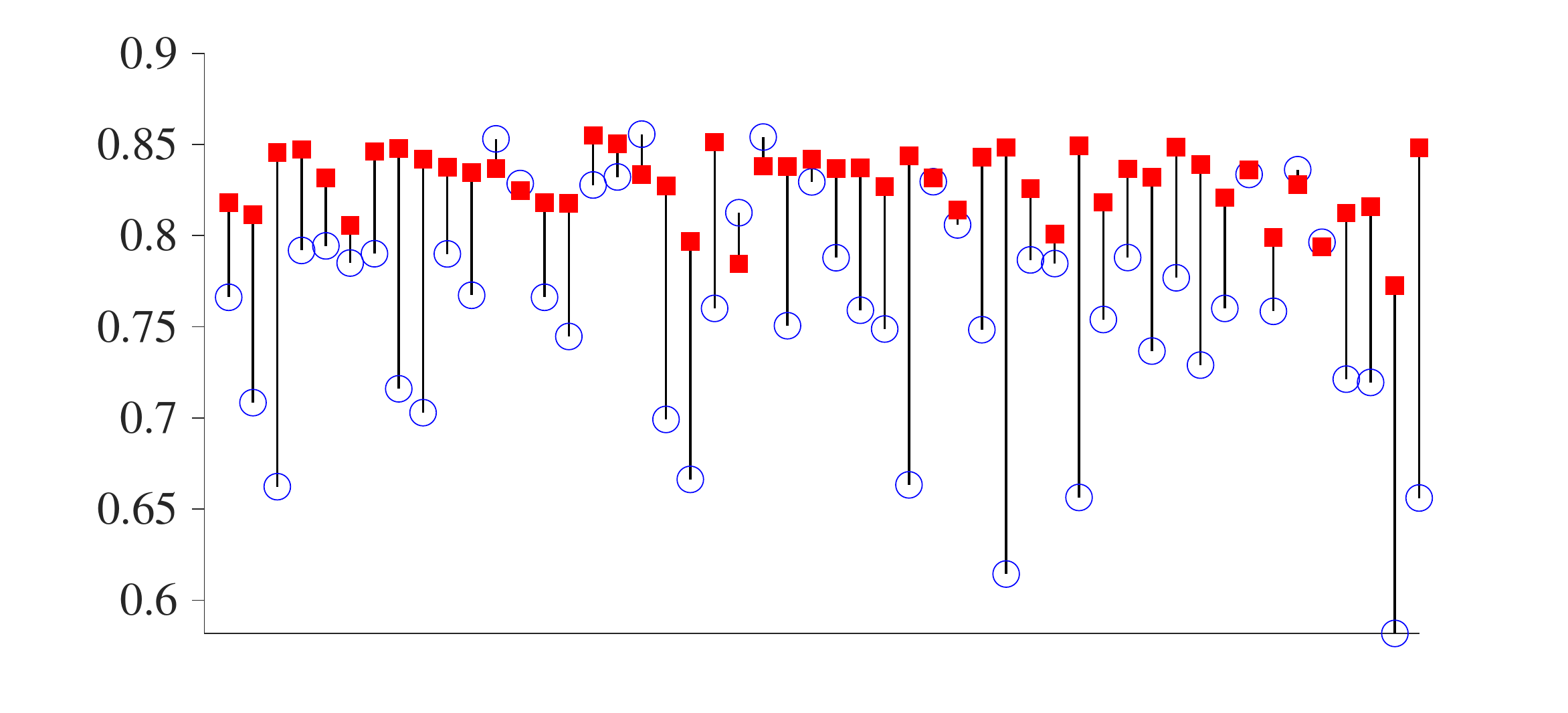}}\hspace{4mm}
	\subfloat[{[DC-DE, $\M_{0.1}$, $m=270$]}]{\label{fig: kro_dc_de_10_30_270}\includegraphics[width=0.31\textwidth]{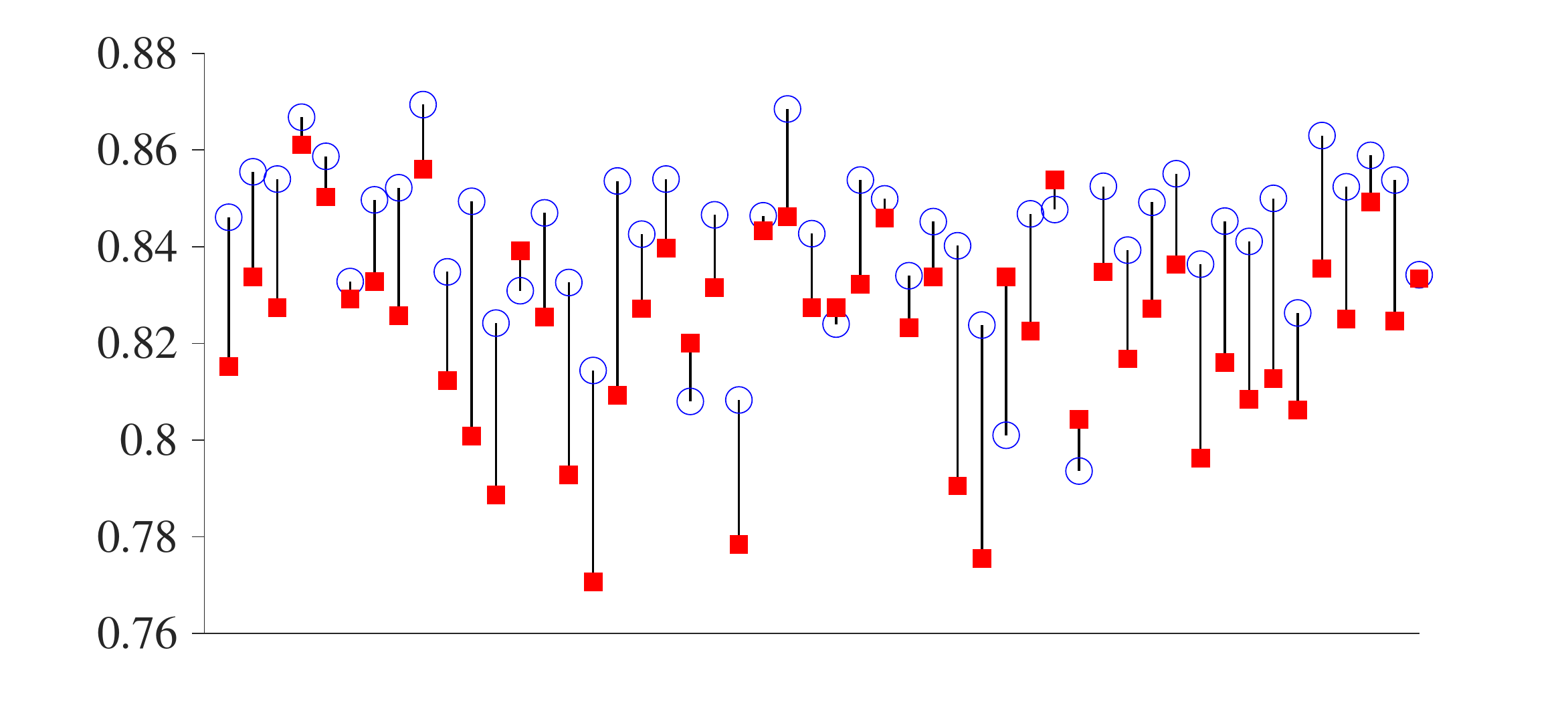}} \hspace{4mm}
	\subfloat[{[DC-DE, $\M_{0.1}$, $m=1350$]}]{\label{fig: kro_dc_de_10_30_1350}\includegraphics[width=0.31\textwidth]{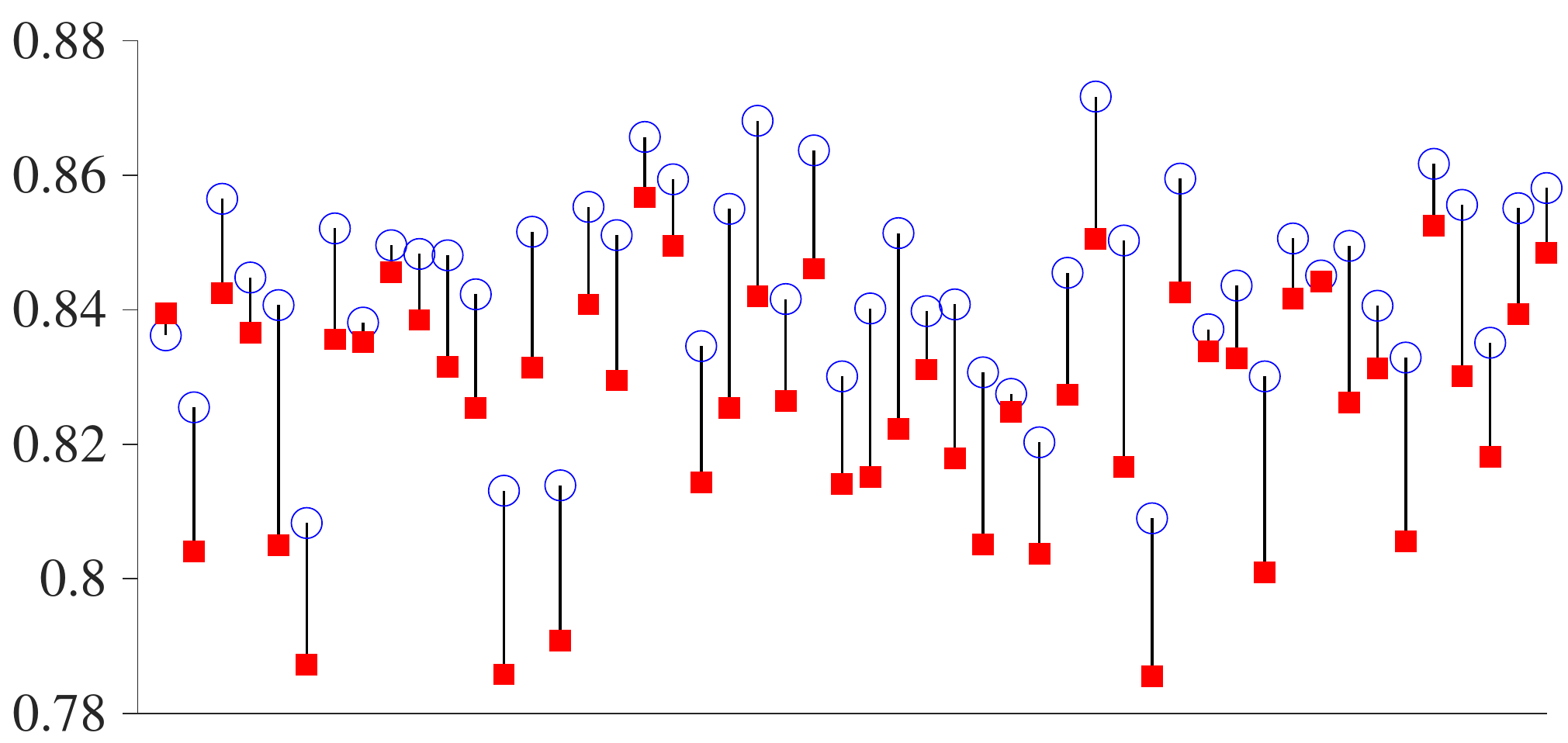}}
	
	\vspace{-2mm}
	\subfloat[{[DC-DE, $\M_{\infty}$, $m=90$]}]{\label{fig: kro_dc_de_uni_30_90}\includegraphics[width=0.31\textwidth]{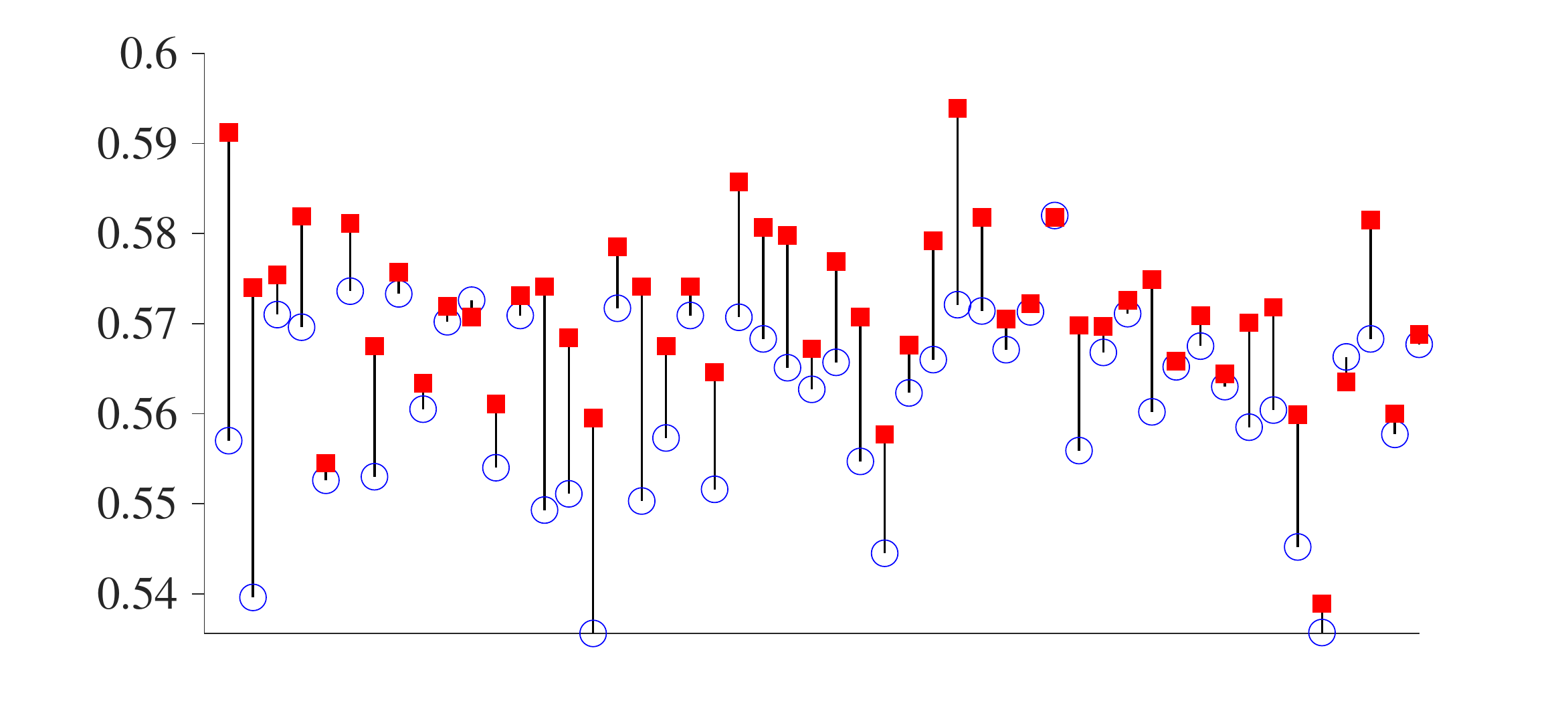}}\hspace{4mm}
	\subfloat[{[DC-DE, $\M_{\infty}$, $m=270$]}]{\label{fig: kro_dc_de_uni_30_270}\includegraphics[width=0.31\textwidth]{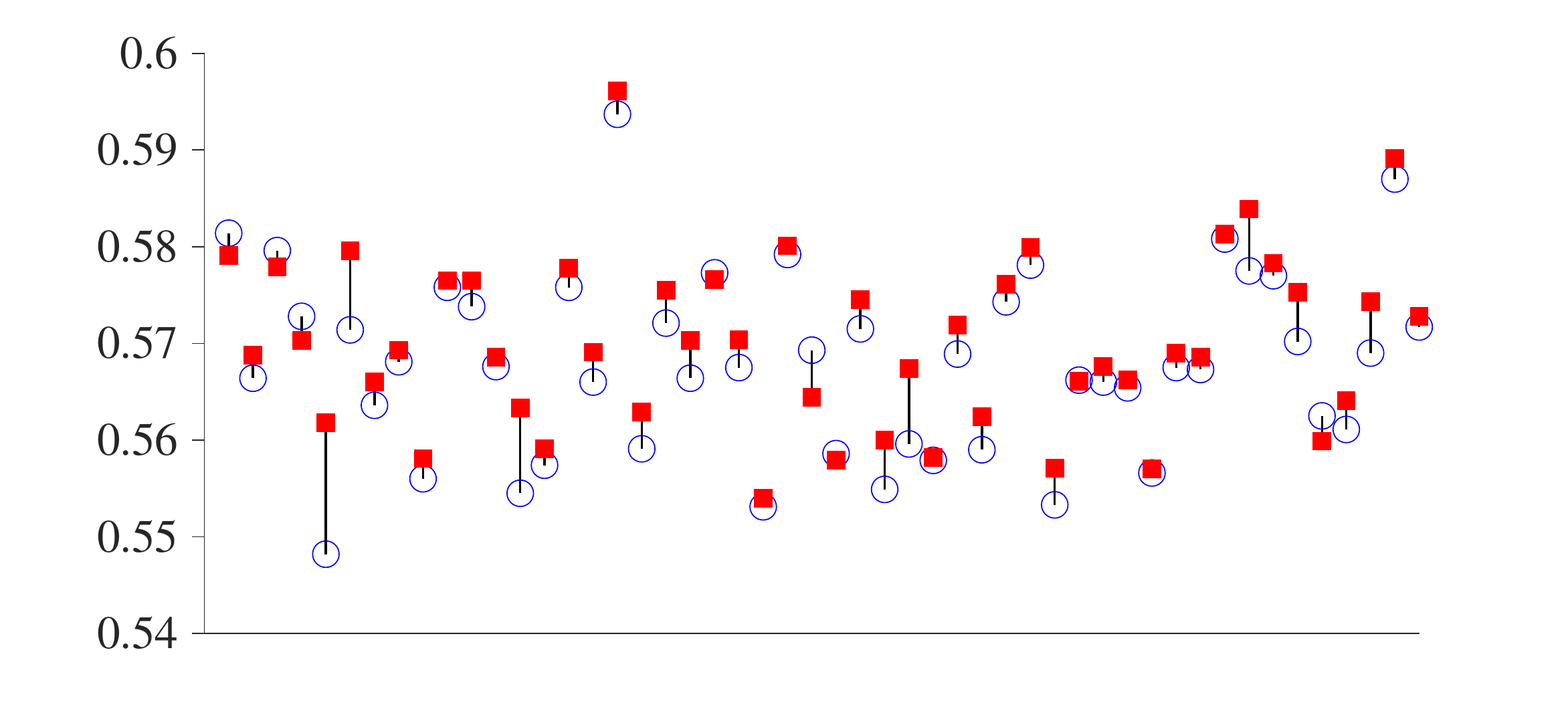}}\hspace{4mm}
	\subfloat[{[DC-DE, $\M_{\infty}$, $m=1350$]}]{\label{fig: kro_dc_de_uni_30_1350}\includegraphics[width=0.31\textwidth]{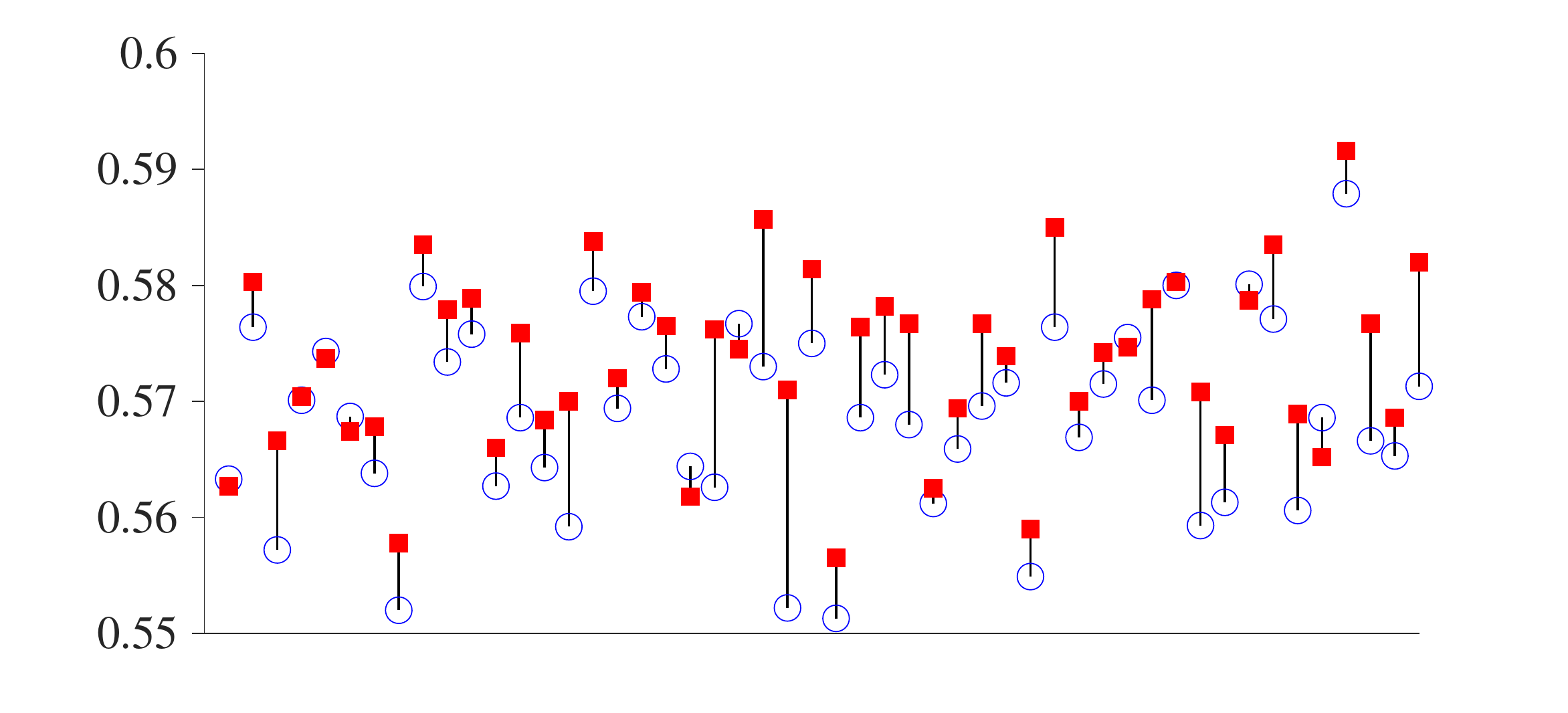}}
	\caption{Each subgraph shows the results of 50 testing queries on the Kronecker graph with $K=30$. For each query, we report the quality (i.e. performance ratio) of the predictions made based on the initial weight (before optimization) and final weight (after optimization).}
	\label{fig: kro_batch}
	\vspace{-5mm}
\end{figure}


\subsection{An in-depth investigation on task migration}

Notably, the effectiveness of Social-Inverse depends not only on the training samples $S_{T_\t}^{\dal_\t}$ (for tuning the weight $\fw$) but also on the expressiveness of the hypothesis space (determined by $\M_{em}$ and $K$). Therefore, with solely the results in Table \ref{table: main}, we are not ready to conclude that samples of DE (resp., DC) are really useful for solving DC (resp., DE). In fact, when $\M_{em}$ is identical to $\M_{true}$, no samples are needed because setting $\fw=1$ can allow for us to perfectly recover the best decisions as long as $K$ is sufficiently large. As a result, the usefulness of the samples should be assessed by examining how much they can help in delivering a high-quality $\fw$. To this end, for each testing query, we report the quality of two predictions made based, respectively, on the initial weight (before optimization) and on the final weight (after optimization). 

Such results for DC-DE on the Kronecker graph are provided in Figure \ref{fig: kro_batch}. As seen from Figure \ref{fig: kro_dc_de_10_30_270}, the efficacy of DC samples in solving DE is statistically significant under $\M_{0.1}$, which might be the first piece of experimental evidence confirming that it is indeed possible to solve one decision-making task by using the query-decision pairs of another one. In addition, according to Figures \ref{fig: kro_dc_de_10_30_90}, \ref{fig: kro_dc_de_10_30_270}, and \ref{fig: kro_dc_de_10_30_1350}, the efficacy of the samples is limited when the sample size is too small, and it does not increase too much after sufficient samples are provided.  On the other hand, with Figure \ref{fig: kro_dc_de_uni_30_270}, we also see that the samples of DC are not very useful when the empirical distribution  (e.g., $\M_{\infty}$) deviates too much from the true model, and in such a case, providing more samples can even cause performance degeneration (Figure \ref{fig: kro_dc_de_uni_30_1350}), which is an interesting observation calling for further investigations. 

The full experimental study can be found in Appendix \ref{subsec: appendix_time}, \ref{subsec: appendix_beta} and \ref{subsec: appendix_full_observation}, including the results on Higgs and Hep, results with more realizations, results of DE-DE and DC-DC, results of GNN under different seeds, a discussion on the impact of $\beta$, and a discussion of training efficiency.

\section{Further discussion}
\label{sec: further}
We close our paper with a discussion on the limitations and future directions of the presented work, with the related works being discussed in Appendix \ref{sec: appendix_related}.

\textbf{Future directions.} 
Following the discussion on Figure \ref{fig: kro_batch}, one immediate direction is to systemically investigate the necessary or sufficient condition in which the task migrations between DE and DC are manageable. In addition, settings in Problem \ref{problem: main} can be carried over to other contagion management tasks beyond DE and DC, such as effector detection \cite{lappas2010finding} and diffusion-based community detection \cite{ramezani2018community}. Finally, in its most general sense, the problem of inverse decision-making with task migrations can be conceptually defined over any two stochastic combinatorial optimization problems \cite{tong2021usco} sharing the same underlying model (e.g., graph distribution). For instance, considering the stochastic shortest path problem \cite{bertsekas1991analysis} and the minimum Steiner tree problem \cite{gupta2005stochastic}, with the information showing the shortest paths between some pairs of nodes, can we infer the minimum Steiner tree of a certain group of nodes with respect to the same graph distribution? Such problems are interesting and fundamental to many complex decision-making applications \cite{ramanathan1997stochastic,celik2019stochastic}.

\textbf{Limitations.} While our method offers promising performance in experiments, it is possible that deep architectures can be designed in a sophisticated manner so as to achieve improved results. In addition, while we believe that similar observations also hold for graphs that are not considered in our experiment, more experimental studies are required to support the universal superiority of Social-Inverse. In another issue, the assumption that the training set contains approximation solutions is the minimal one for the purpose of theoretical analysis, but in practice, such guarantees may never be known. Therefore, experimenting with samples of heuristic query-decision pairs is needed to further justify the practical utility of our method.  Finally, we have not experimented with graphs of extreme scales (e.g., over 1M nodes) due to the limit in memory. We wish to explore the above issues in future work.

\begin{ack}
We thank the reviewers for their time and insightful comments. This work is supported by a) National Science Foundation under Award IIS-2144285 and b) the University of Delaware.
\end{ack}

\bibliographystyle{IEEEtran}
\bibliography{bib_amo}

\section*{Checklist}

The checklist follows the references.  Please
read the checklist guidelines carefully for information on how to answer these
questions.  For each question, change the default \answerTODO{} to \answerYes{},
\answerNo{}, or \answerNA{}.  You are strongly encouraged to include a {\bf
justification to your answer}, either by referencing the appropriate section of
your paper or providing a brief inline description.  For example:
\begin{itemize}
  \item Did you include the license to the code and datasets? \answerYes{See Section~\ref{gen_inst}.}
  \item Did you include the license to the code and datasets? \answerNo{The code and the data are proprietary.}
  \item Did you include the license to the code and datasets? \answerNA{}
\end{itemize}
Please do not modify the questions and only use the provided macros for your
answers.  Note that the Checklist section does not count towards the page
limit.  In your paper, please delete this instructions block and only keep the
Checklist section heading above along with the questions/answers below.

\begin{enumerate}

\item For all authors...
\begin{enumerate}
  \item Do the main claims made in the abstract and introduction accurately reflect the paper's contributions and scope?
    \answerYes{}
  \item Did you describe the limitations of your work?
    \answerYes{See Sec. \ref{sec: further}}
  \item Did you discuss any potential negative societal impacts of your work?
    \answerNo{This work focuses on learning and algorthimic foundations of social contagion management, without being tied to specific applications}
  \item Have you read the ethics review guidelines and ensured that your paper conforms to them?
    \answerYes{}
\end{enumerate}

\item If you are including theoretical results...
\begin{enumerate}
  \item Did you state the full set of assumptions of all theoretical results?
    \answerYes{}
        \item Did you include complete proofs of all theoretical results?
    \answerYes{They can be found in the appendix.}
\end{enumerate}

\item If you ran experiments...
\begin{enumerate}
  \item Did you include the code, data, and instructions needed to reproduce the main experimental results (either in the supplemental material or as a URL)?
    \answerYes{They are provided in the supplemental material. Due to the 100M limitation, we only provide partial data at the submission time and will release the complete dataset (> 6GB) after the review period.}
  \item Did you specify all the training details (e.g., data splits, hyperparameters, how they were chosen)?
    \answerYes{The details are provided in Appendix \ref{subsec: full_setting}}.
        \item Did you report error bars (e.g., with respect to the random seed after running experiments multiple times)?
    \answerYes{Yes, we report standard deviations.}
        \item Did you include the total amount of compute and the type of resources used (e.g., type of GPUs, internal cluster, or cloud provider)?
    \answerYes{See Appendix \ref{subsec: appendix_time}.}
\end{enumerate}

\item If you are using existing assets (e.g., code, data, models) or curating/releasing new assets...
\begin{enumerate}
  \item If your work uses existing assets, did you cite the creators?
    \answerYes{}
  \item Did you mention the license of the assets?
    \answerYes{}
  \item Did you include any new assets either in the supplemental material or as a URL?
    \answerYes{}
  \item Did you discuss whether and how consent was obtained from people whose data you're using/curating?
    \answerYes{We use public dataset and source code following their licenses.}
  \item Did you discuss whether the data you are using/curating contains personally identifiable information or offensive content?
    \answerNo{Our dataset is purely numerical and does not have any personally identifiable information or offensive content.}
\end{enumerate}

\item If you used crowdsourcing or conducted research with human subjects...
\begin{enumerate}
  \item Did you include the full text of instructions given to participants and screenshots, if applicable?
    \answerNA{}
  \item Did you describe any potential participant risks, with links to Institutional Review Board (IRB) approvals, if applicable?
    \answerNA{}
  \item Did you include the estimated hourly wage paid to participants and the total amount spent on participant compensation?
    \answerNA{}
\end{enumerate}

\end{enumerate}


\newpage
\appendix

\section*{Social-Inverse: Inverse Decision-making of Social Contagion Management with Task Migrations (Appendix)}
\section{Diffusion process}
\label{sec: diffusion_example}
\begin{figure}[h]
	\centering
	\vspace{3mm}
	\includegraphics[width=0.98\textwidth]{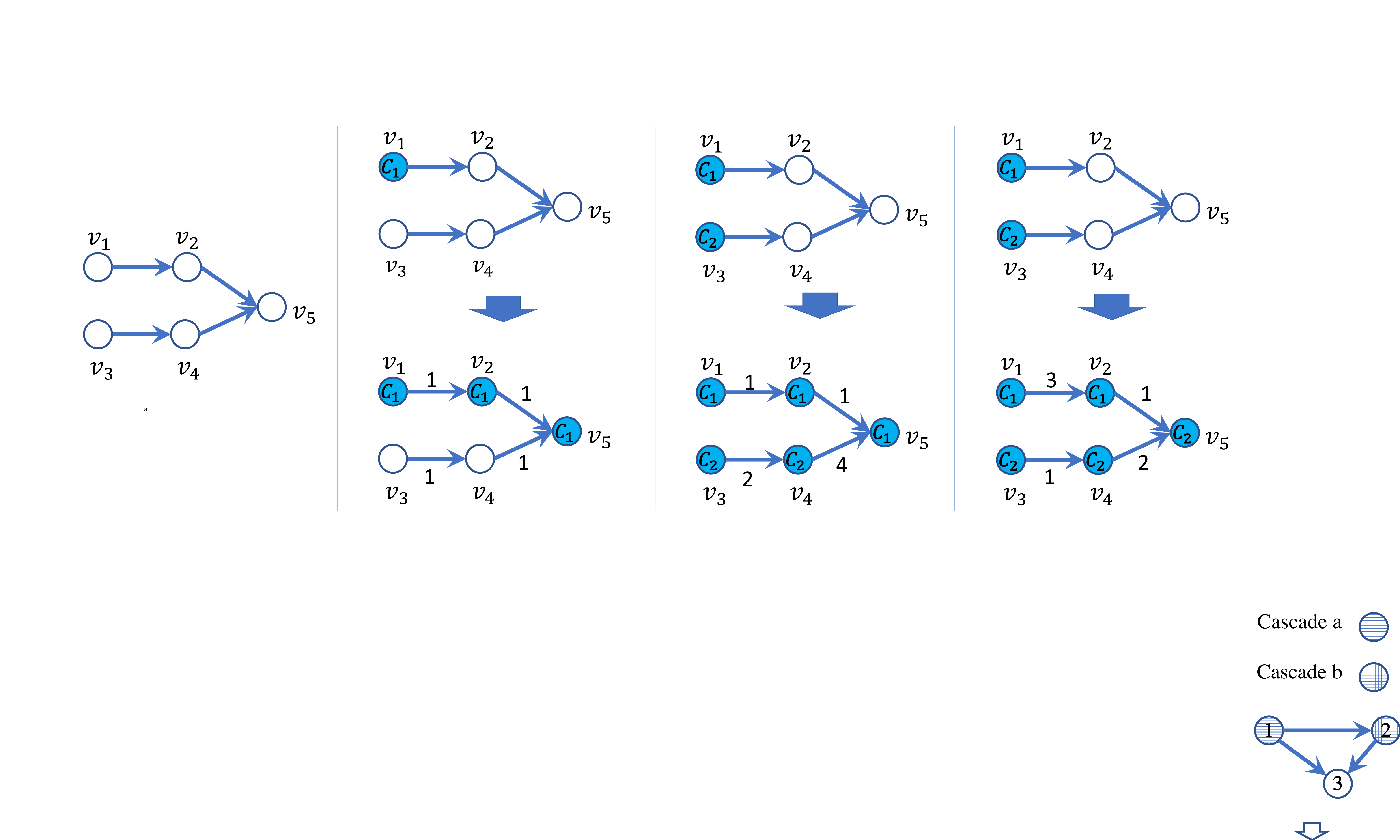} 
	\caption{The first subgraph shows an example social network. The second subgraph presents a possible realization as well as a diffusion process of one cascade spreading from $\{v_1\}$. The third and fourth subgraphs depict a competitive diffusion under different realizations, where the cascades $C_1$ and $C_2$ start to spread from $\{v_1\}$ and $\{v_3\}$, respectively; node $v_5$ is activated by cascade $C_1$ at time $2$ in the third subgraph, while it is activated by cascade $C_2$ at time $3$ in the fourth subgraph.}
	\label{fig: diffusion_example}
\end{figure}

An example of the diffusion process can be found in Figure \ref{fig: diffusion_example}.

\section{Notations}
\label{sec: appendix_notation}
\begin{table}[h]
\renewcommand{\arraystretch}{1.2} 
\vspace{-3mm}
\small
\caption{Notations.}
\centering

\label{table: notation}
\begin{tabular}{  @{}  l      l   l   @{}  }
\toprule
Symbol & Description & Reference\\
\midrule
$G=(V,E)$ & social graph & Sec. \ref{subsec: model} \\
$N_u^-$ & the set of in-neighbors of node $u$ & Sec. \ref{subsec: model} \\ 
$\N_u$ & a distribution over $2^{N_u^-}$ for a node $u$& Sec. \ref{subsec: model} \\
$\T_{(u,v)}$ & a distribution over $(0,+\infty]$ for a pair $(u,v)$ & Sec. \ref{subsec: model}  \\
$\R_G$ & the set of all possible retaliations (i.e., weighted subgraphs) & Def. \ref{def: realization}\\
$\M: \R_G \rightarrow [0,1]$ & the distribution over $\R_G$ specified by model $\M$ &  Def. \ref{def: realization}\\
$f_{\M}^{\DE}$ and $f_{\M}^{\DC}$ & the objective functions of DE and DC associated with model $\M$ & Problems \ref{problem: diff_enhancement}  and \ref{problem: diff_containment}\\
$f_{\M}^{T}$ & the objective function of an abstract task $T$ associated model $\M$ & Problem \ref{problem: abstract_problem} \\
$Y_{\M, T, X}^\alpha$ & an $\alpha$-approximation to  $\argmax_{Y \in \Y_T} f_{\M}^T(X, Y)$ & Problem \ref{problem: abstract_problem} \\
$f_{r}^{T}$ & the objective function projected to a realization $r$ & Def. \ref{def: linearity} \\
$S_{\M, T, m}^\alpha$ & dataset associated with task $T$ and a ratio $\alpha$ & Equation \ref{eq: dataset}\\
$T_\s$ and $T_\t$ & source task and target task & Problem \ref{problem: main}\\
$\M_{true}$  & the true model that defines the optimization tasks & Problem \ref{problem: main}\\
$\dal_\s$ & approximation ratio of the samples regarding the source task & Problem \ref{problem: main}\\
$\M_{em}$ & an empirical distribution over $\R_G$ & Sec. \ref{sec: theory}\\
$\F_{R_K}$ & the hypothesis space defined over a set $R_K$ of $K$ realizations & Sec. \ref{sec: theory}\\
$H_{\KR, \vec{w}}^{\t}(X,Y)$ & the hypothesis associated with vector $\vec{w}$ and $\KR$ & Sec. \ref{sec: theory} \\
$\pw$ and $\fw$ & the prior weight and final weight & Sec. \ref{sec: theory}\\
$\gamma$ & the parameter controlling the distribution for sampling  $\fw$ & Sec. \ref{sec: theory}  \\
$Y_{\KR, \fw, X}^{\ial_\t}$ & an $\ial_\t$-approximation for inference & Equation \ref{eq: inference}\\
$ \I_{\KR,  \pw,  X, \beta}^{\t}$ & the margin set specifying potential predictions associated with $X$ & Equation \ref{eq: margin}\\
$\beta$ & the parameter that controls the strictness of margin & Equation \ref{eq: margin}\\
$H_{\KR, \vec{w}}^{\s}(X,Y)$ & the score function associated with the source task & Equation \ref{eq: new_constraint} \\
\bottomrule
\end{tabular}
\vspace{-5mm}
\end{table}

\section{An illustration of Social-Inverse}
\label{sec: example_si}
In what follows, we present concrete steps in Social-Inverse for solving Problem \ref{problem: main}, using an example with $T_\s=\DE$, $T_\t=\DC$, $K=30$, and $M_{em}=\M_{\infty}$, where $\M_{\infty}$ samples random subgraphs with unit weights. In such a case, the best ratio one can have is $\dal_\s=\ial_\t=1-1/e \approx 0.618$. 
\begin{itemize}
	\item \textbf{Instance setting.} A social graph $G=(V, E)$ and an underlying diffusion model $\M_{true}$.
	\item \textbf{Input.} A collection of pairs $S_{\DE}^{\dal_\s}=\Big\{(X_i, Y_{\M_{true},  \DE, X_i}^{\dal_\s})\Big\}_{i=1}^m$, where $X_i$ is a subset of $V$ and $Y_{\M_{true},  \DE, X_i}^{\dal_\s}$ is a $(1-1/e)$-approximation to $\argmax_{Y \subseteq V, |Y|\leq k} f_{\M_{true}}^{\DE}(X_i, Y)$ for some predefined budget $k \in \mathbb{Z}$.
	\item \textbf{Step 1.} Sample a collection of $K=30$ iid realizations $R_K=\{r_1,...,r_{30}\} \subseteq \R_G$ following $\M_{\infty}$.
	\item \textbf{Step 2.} With $S_{\DE}^{\dal_\s}$ and $R_K$, calculate a prior vector $\pw$ by solving the following program using, for example, the n-slack cutting plane algorithm \cite{tsochantaridis2005large}:
	{\small
	\begin{align*}
	& {\text{min}}  & & \norm{\vec{w}}^2+ C\sum_{i=1}^{m}\xi_i / m\nonumber \\
	& \text{s.t.}  & &\ial_\t (\ial_\t - \beta) \cdot H_{\KR, \vec{w}}^{\s}(X_i, Y_{\M_{true},  \DE, X_i}^{\dal_\s}) - H_{\KR, \vec{w}}^{\s}(X_i,Y) \geq \xi_i,~\forall i \in [m],  ~\forall Y \in \Y_\s \\
	&  & & \vec{w} \geq 0
	\end{align*}}
	In solving the above program, we need to repeatedly deal with some intermediate weight $\vec{w}=(w_1,...,w_{30})$ and solve 
	\begin{align*}
	\argmax_{Y \subseteq V, |Y| \leq \big|Y_{\M_{true},  \DE, X_i}^{\dal_\s}\big|}  H_{\KR, \vec{w}}^{\s}(X_i,Y) =\sum_{i=1}^{30} {w}_i \cdot f_{r_i}^{\DE}(X_i,Y),
	\end{align*}
	which can be approximated within in a ratio of $1-1/e$ (Theorem \ref{theorem: algorithm}). 
	\item \textbf{Step 3.} Acquire $\gamma$ through Equation \ref{eq: gamma}, and sample the final vector $\fw$ following $Q(\gamma \cdot \pw, \mathcal{I})$. Notice that this step is optional due to the facts that a) every entry of $\pw$ is scaled by the same factor and b) the prediction is invariant to scaling $\pw$.
	\item \textbf{Step 4.} With the final vector $\fw = (\overline{{w}}_1,...,\overline{w}_{30}) $ and $R_K$, given a new query $X^* \subseteq V$ of the target task DC and a budget $k$, make a prediction by solving the following problem using the greedy algorithm (Theorem \ref{theorem: algorithm}):
	\begin{equation*}
	\argmax_{Y \subseteq V, |Y|\leq k} H_{\KR, \fw}^{\t}(X^{*}, Y) = \sum_{i=1}^K \overline{w}_i \cdot f_{r_i}^{\DC}(X^{*}, Y).
	\end{equation*}
	
\end{itemize}

\section{Proofs}
\subsection{Proof of Theorem \ref{theorem: bound}}
The proof of Theorem \ref{theorem: bound} is derived based on the following two results. The first lemma is the standard PAC-Bayesian bound for structured prediction \cite{mcallester2003simplified}. The second lemma (proved in Sec. \ref{subsubsec: proof_margin_chance}) states that with a high chance, the prediction based on the sampled vector $\fw$ will fall into the margin defined by the prior vector $\pw$.
\begin{lemma}
	\label{lemma: pac-bayesian}
	For each $R_K$, let $\Pc$ be a prior distribution over the hypothesis space (i.e., $\mathbb{R}^K$) and let $\delta \in (0,1)$. Then, with probability of at least $1-\delta$ over the choice of an iid training set $S=\{z_1,...,z_m\}$ sampled according to $\D$, for all distributions $\Qc$ over $\mathbb{R}^K$, we have
	\begin{align*}
	\E_{\overline{\vec{w}} \sim \Qc, X \sim \D} \Big[ l(X, Y_{\KR, \fw, X}^{\ial_\t}\Big] 
	\leq \frac{1}{m} \sum_{i=1}^{m} \E_{\overline{\vec{w}} \sim \Qc }\Big[l(X_i, Y_{\KR, \fw, X}^{\ial_\t} )\Big]  + \sqrt{\frac{D(\Qc|| \Pc)+\ln (m/\delta)}{2(m-1)}}
	\end{align*}
	where $D(\Qc || \Pc)$ is the KL divergence.
\end{lemma}

\begin{lemma}
	\label{lemma: margin_chance}
	For each $\widetilde{\vec{w}}$, $R_K$ and $\phi_{true}$, with probability at least $1-\frac{\norm{\widetilde{\vec{w}}}^2}{m}$ over the selection of $\overline{\vec{w}}$, we have
	\begin{equation}
	\label{eq: margin_chance}
	l(X, Y_{\KR, \fw, X}^{\ial_\t})  \leq \max_{Y \in \Y_\t } l(X,Y) \cdot  \mathds{1}_{ \I_{\KR,  \pw,  X,  \beta}^{\t}}(Y)
	\end{equation}
	holding simultaneously for all $X \in \X_\t$, provided that $\gamma = \frac{\ial_\t^2+1}{\min_p |\pws_p| \cdot \beta \cdot  \ial_\t}\sqrt{2\ln \frac{2mK}{\norm{\widetilde{\vec{w}}}^2}}$.
\end{lemma}

With the above two lemmas, we now prove Theorem \ref{theorem: bound}. Let the prior distribution $\Pc$ in Lemma \ref{lemma: pac-bayesian} be the isotropic Gaussian with an identity covariance and a zero mean, and $\Qc$ be $Q(\gamma\cdot \pw, \mathcal{I})$. We have the following inequality,
\begin{align}
\label{eq: bound_1}
\E_{\overline{\vec{w}} \sim Q, X \sim \D} \Big[ l(X, Y_{\KR, \fw, X}^{\ial_\t}\Big] 
\leq \frac{1}{m} \sum_{i=1}^{m} \E_{\overline{\vec{w}} \sim Q}\Big[l(X_i, Y_{\fw, \KR, X_i}^{\ial_\t} )\Big]  +\sqrt{\frac{\frac{\gamma^2\norm{\widetilde{\vec{w}}}^2}{2} +\ln \frac{m}{\delta}}{2(m-1)}}.
\end{align}
According to Lemma \ref{lemma: margin_chance}, for each $X_i$, we have 

\begin{align}
\label{eq: bound_2}
&\E_{\overline{\vec{w}} \sim Q }\Big[l(X_i, Y_{\fw, \KR, X_i}^{\ial_\t} )\Big] \nonumber\\\
\leq &(1-\frac{\norm{\widetilde{\vec{w}}}^2}{m}) \max_{Y \in \Y_\t } l(X_i,Y) \cdot  \mathds{1}_{  \I_{\KR,  \pw,  X_i,\beta}^{\t}}(Y)+\frac{\norm{\widetilde{\vec{w}}}^2}{m} \max_{Y \in \Y_{\t}} l(X_i,Y) \nonumber\\
\leq& \max_{Y \in \Y } l(X_i,Y) \cdot  \mathds{1}_{ \I_{\KR,  \pw,  X_i,\beta}^{\t}}(Y)+\frac{\norm{\widetilde{\vec{w}}}^2}{m}.
\end{align}
Combining Equations \ref{eq: bound_1} and \ref{eq: bound_2}, we have 
\begin{align*}
&\E_{\overline{\vec{w}} \sim Q, X \sim \D_{\X}} \Big[ l(X, Y_{\KR, \fw, X}^{\ial_\t}\Big] \\
\leq& \frac{1}{m} \sum_{i=1}^{m}  \max_{Y \in \Y_\t } l(X_i,Y) \cdot  \mathds{1}_{ \I_{\KR,  \pw,  X_i,\beta}^{\t}}(Y)+\frac{\norm{\widetilde{\vec{w}}}^2}{m}  +\sqrt{\frac{\frac{\gamma^2\norm{\widetilde{\vec{w}}}^2}{2} +\ln \frac{m}{\delta}}{2(m-1)}}.
\end{align*}

\subsubsection{Proof of Lemma \ref{lemma: margin_chance}}
\label{subsubsec: proof_margin_chance}

Let $\pw=(\pws_1,...,\pws_K)$, $\fw=(\fws_1,...,\fws_K)$ and $\epsilon \define \frac{\beta\cdot \ial_\t}{\ial_\t^2+1}$. We first prove the following result:
\begin{align}
\label{eq: bound_3}
|\overline{{w}}_p-\gamma \cdot \widetilde{{w}}_p |\leq \epsilon \cdot  \gamma \cdot \widetilde{{w}}_p, \forall p\in\{1,...,K\} \implies Y_{\KR, \fw, X}^{\ial_\t} \in \I_{\KR,  \pw,  X,  \beta}^{\t} \implies \text{Equation \ref{eq: margin_chance}}.
\end{align}
The second $\implies$ holds by definition, so it is left to prove the first part. With the LHS of Equation\ref{eq: bound_3}, by rearrangements, we have 
{
\begin{align*}
&\ial_\t \cdot H_{\KR, \pw}^{\t}(X_i,Y_{\KR,\pw,X_i}^{\ial_\t}) - H_{\KR, \pw}^{\t}(X_i,Y_{\KR,  \fw, X_i}^{\ial_\t}) \\
=& \frac{1}{\gamma} \Big(\ial_\t \cdot H_{\KR,\gamma \cdot \pw}^{\t}(X_i,Y_{\KR,\pw,X_i}^{\ial_\t}) - H_{\KR,\gamma \cdot \pw}^{\t}(X_i,Y_{\KR,  \fw, X_i}^{\ial_\t})\Big)\\
=& \frac{1}{\gamma} \Big(\ial_\t \cdot H_{\KR, \gamma \cdot \pw -\fw}^{\t}(X_i,Y_{\KR,\pw,X_i}^{\ial_\t})+\ial_\t \cdot H_{\KR, \fw}^{\t}(X_i,Y_{\KR,\pw,X_i}^{\ial_\t}) \\
&\hspace{2cm} - H_{\KR, \gamma \cdot \pw - \fw}^{\t}(X_i,Y_{\KR, \fw, X_i}^{\ial_\t}) - H_{\KR, \fw}^{\t}(X_i,Y_{\KR,  \fw, X_i}^{\ial_\t})\Big)\\
&\{\text{by the LHS of Equation\ref{eq: bound_3}}\}\\
=& \frac{1}{\gamma} \Big(\ial_\t \cdot H_{\KR, \epsilon \cdot \gamma  \cdot \pw}^{\t}(X_i,Y_{\KR,\pw,X_i}^{\ial_\t})+\ial_\t \cdot H_{\KR, \fw}^{\t}(X_i,Y_{\KR,\pw,X_i}^{\ial_\t}) \\
&\hspace{2cm} + H_{\KR, \epsilon \cdot \gamma  \cdot \pw}^{\t}(X_i,Y_{\KR,  \fw, X_i}^{\ial_\t}) - H_{\KR, \fw}^{\t}(X_i,Y_{\KR,  \fw, X_i}^{\ial_\t})\Big).
\end{align*}}
Since $H_{\KR, \fw}^{\t}(X_i,Y_{\KR,  \fw, X_i}^{\ial_\t}) \geq \ial_\t \max_{Y \in \Y} H_{\KR, \fw}^{\t}(X_i, Y) \geq \ial_\t \cdot H_{\KR, \fw}^{\t}(X_i,Y_{\KR,\pw,X_i}^{\ial_\t})$, the above inequality is further upper bounded by
\begin{align*}
& \frac{1}{\gamma} \Big( \ial_\t \cdot H_{\KR, \epsilon \cdot \gamma  \cdot \pw}^{\t}(X_i,Y_{\KR,\pw,X_i}^{\ial_\t})+H_{\KR, \epsilon \cdot \gamma  \cdot \pw}^{\t}(X_i,Y_{\KR,  \fw, X_i}^{\ial_\t}) \Big),
\end{align*}
which equals to 
\begin{align}
\label{eq: t1}
\ial_\t \cdot H_{\KR, \epsilon \cdot  \pw}^{\t}(X_i,Y_{\KR,\pw,X_i}^{\ial_\t})+H_{\KR, \epsilon \cdot \pw}^{\t}(X_i,Y_{\KR,  \fw, X_i}^{\ial_\t}).
\end{align}
By the definition of $Y_{\KR,\pw,X_i}^{\ial_\t}$, Equation \ref{eq: t1} is no more than 
\begin{align}
\label{eq: t2}
& \ial_\t \cdot H_{\KR, \epsilon \cdot  \pw}^{\t}(X_i,Y_{\KR,\pw,X_i}^{\ial_\t})+\frac{1}{\ial_\t} \cdot H_{\KR, \epsilon \cdot \pw}^{\t}(X_i,Y_{\KR,\pw,X_i}^{\ial_\t}) \nonumber \\
&= (\epsilon \cdot \ial_\t+\frac{\epsilon}{\ial_\t})\cdot H_{\KR, \pw}^{\t}(X_i,Y_{\KR,\pw,X_i}^{\ial_\t}) 
\end{align}
By the selection of $\epsilon$, the RHS of Equation \ref{eq: t2} is upper bounded by $\beta \cdot H_{\KR, \pw}^{\t}(X_i,Y_{\KR,\pw,X_i}^{\ial_\t})$.  which implies $Y_{\KR, \fw, X}^{\ial_\t} \in \I_{\KR,  \pw,  X,  \beta}^{\t}$ by the definition of $\I_{\KR,  \pw,  X,  \beta}^{\t}$ (Equation \ref{eq: margin}). Therefore, Equation  \ref{eq: bound_3} is proved.

With Equation  \ref{eq: bound_3}, to prove Lemma \ref{lemma: margin_chance}, it suffices to prove 
\begin{align}
\label{eq: bound_4}
\Pr_{\fw \sim Q} \Big[ |\overline{{w}}_p-\gamma \cdot \widetilde{{w}}_p |\leq \epsilon \cdot  \gamma \cdot \widetilde{{w}}_p, \forall p\in [K] \Big] \geq 1- \frac{\norm{\widetilde{\vec{w}}}^2}{m}
\end{align}
By the property of Gaussian, we have 
\begin{align}
\Pr\Big[|\overline{{w}}_p-\gamma \cdot \widetilde{{w}}_p |\geq \epsilon \cdot  \gamma \cdot \widetilde{{w}}_p\Big] \leq 2\exp(\frac{- (\gamma \cdot \widetilde{{w}}_p \cdot \epsilon)^2}{2}),
\end{align}
which is no more than $ \frac{\norm{\widetilde{\vec{w}}}^2}{K\cdot m}$ by the selection of $\epsilon$ and $\gamma$. This implies Equation \ref{eq: bound_4} due to the union bound over $p \in [K]$.

\subsection{Proof of Theorem \ref{theorem: realizability}}
\label{subsec: proof_theorem: realizability}
The next lemma (proved in Sec. \ref{subsubsec: proof_lemma: approximation}) relates the margin (Equation \ref{eq: margin}) with the latent objective function (i.e., $f_{\M_{true}}^{T_\t}$) by stating that our choice of $K$ ensures that the hypothesis space contains a score function that can approximate the latent objective function to a sufficient extent.
\begin{lemma}
	\label{lemma: approximation}
	For each $\epsilon, \delta_1 , \delta_2 \in \mathbb{R}$ and $\M_{em}$, when $K=\frac{4 \Delta ^2}{ \epsilon^2 \cdot \delta_2^2}(\ln|\Y_{T_\t}|+\ln \frac{1}{\delta_1}+1)$, with probability at least $1-\delta_1$ over the selection of $R_K$, there exist a $\widetilde{\vec{w}}$ such that we have
	\begin{equation}
	\label{eq: approximation}
	\Pr_{X \sim \D} \Big[ \big| H_{\KR, \pw}^{\t}(X,Y)-f_{\M_{true}}^{T_\t}(X,Y) \big| \leq \epsilon \cdot  f_{\M_{true}}^{T_\t}(X,Y), \forall~Y \in \Y \Big] \geq 1-\delta_2.
	\end{equation}
\end{lemma}
To prove Theorem \ref{theorem: realizability}, it suffices to prove that the event in Equation \ref{eq: approximation} implies the event in Equation \ref{eq: realizability}. With the event in Equation \ref{eq: approximation}, we have for each $Y \in \I_{\KR,  \pw,  X,  \beta}^{\t}$ that  
\begin{align*}
&\ial_\t \cdot H_{\KR, \pw}^{\t}(X,Y_{\KR,\pw,X}^{\ial_\t}) - H_{\KR, \pw}^{\t}(X,Y) \leq \beta \cdot  H_{\KR, \pw}^{\t}(X,Y_{\KR,\pw,X}^{\ial_\t})\\
&\{\text{by the definition of $Y_{\KR,\pw,X}^{\ial_\t}$}\}\\
\implies& \ial_\t \cdot  (\ial_\t - \beta) \cdot H_{\KR, \pw}^{\t}\big(X,\argmax_{Y\in \Y_{\t}} f_{\M_{true}}^{T_\t}(X,Y) \big) - H_{\KR, \pw}^{\t}(X,Y) \leq 0\\
&\{\text{by the event in Equation \ref{eq: approximation}}\}\\
\implies& \ial_\t \cdot  (\ial_\t - \beta) (1-\epsilon)\cdot  f_{\M_{true}}^{T_\t}\big(X,\argmax_{Y\in \Y_\t} f_{\M_{true}}^{T_\t}(X,Y) \big) - (1+\epsilon) \cdot f_{\M_{true}}^{T_\t}(X,Y) \leq 0\\
\implies& \ial_\t \cdot  (\ial_\t - \beta) (1-\epsilon) \cdot  \max_{Y \in \Y_\t} f_{\M_{true}}^{T_\t}\big(X,Y \big) - (1+\epsilon) \cdot f_{\M_{true}}^{T_\t}(X,Y) \leq 0.
\end{align*}
This implies the desired results as follows:
\begin{align*}
\max_{Y \in \Y_\t} l_{\appro}(X, Y) \cdot  \mathds{1}_{  \I_{\KR,  \pw,  X,  \beta}^{\t} }(Y) \leq 1 - \frac{ \ial_\t \cdot  (\ial_\t - \beta) \cdot(1-\epsilon)}{(1+\epsilon)}
\end{align*}

\subsubsection{Proof of Lemma \ref{lemma: approximation}}
\label{subsubsec: proof_lemma: approximation}
The proof basically concerns about the concentration of averages towards the mean in a Hilbert space, which is similar to the proof of Lemma 4 in \cite{rahimi2008weighted}. For completeness, we give the proof below. 

For each $X \in \X_\t$ and $Y \in \Y_\t$, we have the observation that
\begin{align}
\E_{r \sim \M_{em}} \Big[\frac{M_{true}(r)}{\M_{em}(r)} f_{r}^{T_{\t}}(X,Y) \Big] = f_{\M_{true}}^{T_{\t}}(X,Y)
\end{align}
Noticing that $R_K=\{r_1,...,r_K\}$ are iid samples from $\M_{em}$, we consider its normalized average function 
\begin{align*}
\hat{h}_{(X,Y)}(r_1,...,r_K)=\frac{1}{ f_{\M_{true}}^{T_{\t}}(X,Y)} \sum_{i=1}^{K}\frac{\M_{true}(r_i)}{K\cdot \M_{em}(r_i)}f_{r_i}^{T_{\t}}(X,Y).
\end{align*}
For each $Y \in \Y_\t$, we denote by $g_Y(r_1,...,r_K)$ the L2 norm of $|\hat{h}_{(X,Y)}(r_1,...,r_K)-1|$ with respect to $X \sim \D$:
\begin{align}
g_Y(r_1,...,r_K)= \sqrt{\int_{X \in \X_\t} \Big( \hat{h}_{(X,Y)}(r_1,...,r_K)-1 \Big)^2 d \D(X)}.
\end{align}
Taking $g_Y(r_1,...,r_K)$ as a function of the variables $r_i$, its stability is given by the fact that for each $\{r_1,...,r_K, r_i^*\} \subseteq \R_G$, we have
\begin{align*}
\big|g_Y(r_1,..., r_i,..., r_K) -  g_Y(r_1,..., r_i,..., r_K)\big| \leq \sup_r \frac{\M_{true}(r)}{\M_{em}(r)} \frac{2 \max f_r^{T_\t}(X, Y)}{K \min f_r^{T_\t}(X, Y)}  = 2\Delta/K.
\end{align*}
In addition, an upper bound of the expectation of $g_Y(r_1,...,r_K)$  is found by
\begin{align}
\label{eq: temp1}
&\E_{R_K \sim \M_{em}} [g_y(r_1,...,r_K)]\leq \sqrt{\E_{R_K \sim \M_{em}} [g_y(r_1,...,r_K)^2]} \nonumber \\
= &\sqrt{ \int_{X \in \X_\t}  \E_{R_K \sim \M_{em}}[\Big( \hat{h}_{(X, Y)}(r_1,...,r_K)-1 \Big)^2] d \D(X)} \nonumber \\
= &\sqrt{ \int_{X \in \X_\t} \frac{1}{K} \Big(\E_{r \sim \M_{em}}[\Big(\frac{1}{ f_{\M_{true}}^{T_{\t}}(X,Y)} \frac{\M_{true}(r)}{\M_{em}(r)}f_{r}^{T_{\t}}(X,Y)\Big)^2]-1 \Big) d \D(X) } \nonumber\\
\leq & \sqrt{ \int_{X \in \X_\t } \frac{1}{K}  \Big(\sup_r \frac{\M_{true}(r)}{\M_{em}(r)}  \frac{\max f_r^{T_\t}(X, Y)}{\min f_r^{T_\t}(X, Y)} \Big)^2 d \D(X) } =  \frac{\Delta}{\sqrt{K}}
\end{align}
With the stability result and a bounded expectation, we have the following results:
\begin{align*}
&\Pr_{R_K \sim \M_{em}}[g_Y(r_1,...,r_K) - \frac{\epsilon \cdot \delta_2}{2} \geq  \frac{\epsilon \cdot \delta_2}{2} ]\\
\leq& \Pr_{R_K \sim \M_{em}}[g_Y(r_1,...,r_K) -\frac{\Delta}{\sqrt{K}} \geq  \frac{\epsilon \cdot \delta_2}{2} ]\\
\leq& \Pr_{R_K \sim \M_{em}}[g_Y(r_1,...,r_K) - \E_{R_K \sim \M_{em}} [g_y(r_1,...,r_K)] \geq  \frac{\epsilon \cdot \delta_2}{2} ]\\
\leq& \exp(\frac{-2 (\frac{\epsilon \cdot \delta_2}{2} )^2 }{K(2\Delta/K)^2}) = \delta_1/|\Y_\t|,
\end{align*}
where the first and the third inequalities follow from selection the selection of $K$ and the second inequality follows from Equation \ref{eq: temp1}. Taking the union bound over $Y \in \Y_\t$, we have the followings holding  simultaneously for each $Y \in \Y_\t$ with probability at least $1-\delta_1$,
\begin{align*}
&\E_{X \sim \D} \Big[|\hat{h}_{(X,Y)}(r_1,...,r_K)-1 | \Big] \\
\leq &\sqrt{\E_{X \sim \D} \Big[|\hat{h}_{(X,Y)}(r_1,...,r_K)-1 |^2 \Big] } = g_Y(r_1,...,r_K) \leq \epsilon \cdot \delta_2
\end{align*}
By the Markov's inequality, we have
\begin{align*}
&\Pr_{X \sim \D} \Big[ | \hat{h}_{(X,Y)}(r_1,...,r_K) -1  | \leq \epsilon, \forall~Y \in \Y_\t \Big] \leq  \frac{\epsilon \cdot \delta_2}{\epsilon}\implies\\
&\Pr_{X \sim \D} \Big[ | \sum_{i=1}^{K}\frac{\M_{true}(r_i)}{K\cdot \M_{em}(r_i)}f_{r_i}^{T_{\t}}(X,Y) - f_{\M_{true}}^{T_{\t}}(X,Y)  | \leq \epsilon f_{\M_{true}}^{T_{\t}}(X,Y), \forall~Y \in \Y_\t \Big]\leq \delta_2,
\end{align*}
which has suggested that $(\frac{\M_{true}(r_1)}{K\cdot \M_{em}(r_1)},..., \frac{\M_{true}(r_K)}{K\cdot \M_{em}(r_K)})$ is the desired weight. 

\subsection{Proof of Theorem \ref{theorem: algorithm}}
A set function $f$ over the subsets of a set $U$ is submodular if it has a diminishing marginal return \cite{fujishige2005submodular}: 
\begin{align*}
f(\{v\} \cup U_1)-f(U_1) \geq f(\{v\} \cup U_2)-f(U_2), ~ \forall~ U_1 \subseteq U_2 \subseteq~U, v \notin U_2. 
\end{align*}
Maximizing a nonnegative monotone submodular function over a matroid can be approximated within a ratio of $1-1/e$ \cite{fujishige2005submodular,calinescu2011maximizing}. It is well-known that for DE and DC, the kernel function is Np-hard to optimize in optimal  over a general realization $r \in \R_G$, but it is submodular \cite{kempe2003maximizing,budak2011limiting}. Since submodularity is preserved under non-negative linear combination, $H_{\KR, \vec{w}}^{\s}(X_i,Y)$ and $H_{\KR, \vec{w}}^{\t}(X,Y)$ are both submodular for any $\vec{w}\geq 0$, thereby admitting $(1-1/e)$-approximations. In particular, when the matroid reduces to the cardinality constraint, the simple greedy algorithm provides an $(1-1/e)$-approximation \cite{nemhauser1978analysis}.

\newpage

\begin{table}[h]
	\renewcommand{\arraystretch}{1.2} 
	\small
	\caption{\textbf{Datasets.} }
	\centering
	\label{table: data}
	\begin{tabular}{  @{}  l   @{\hspace{4mm}}   l @{\hspace{2mm}}  l @{\hspace{2mm}}  l@{\hspace{2mm}} l @{\hspace{2mm}}  l @{\hspace{2mm}}  l }
		& Kronecker &    Erdős-Rényi  & Higgs & Hep  	\\
		\midrule
		Nodes & 1024 & 	512 		&   10000 &  15233       \\
		
		Edges & 2655 & 	 650	&  22482 & 32235	      \\
		
		Size of the pool of query-decision pairs & 2700 & 2700	 &  810	 & 810 	     \\
		
		Size of the pool of realizations & 10000 & 10000	 &  1000	 & 1000 	     \\
		
		Default training size & 270 & 270	 & 270	 & 270 	     \\
		Default testing size & 540& 540	 & 540 & 540	     \\
		Graph structure source & \cite{leskovec2010kronecker} & \cite{hagberg2008exploring}	 & \cite{de2013anatomy}	 & \cite{leskovec2007graph} 	   \\
		\bottomrule
		\vspace{-5mm}
	\end{tabular}
\end{table}

\section{Empirical studies}
\label{sec: full_exp}

In this section, we first provide details in experimental settings (Sec. \ref{subsec: full_setting}), and then present the full experimental results (Secs. \ref{subsec: appendix_time}, \ref{subsec: appendix_beta} and \ref{subsec: appendix_full_observation}). The data and source code can be found in the supplementary material, including the sample pools and pre-train models. Due to the 100M limit, we only include the realizations of Kronecker and Erdős-Rényi under $\M_{\infty}$. The complete dataset (>6GB) will be released after the review period.

\subsection{Detailed experimental settings}
\label{subsec: full_setting}


\subsubsection{The latent model $\M_{true}$ and samples}
\label{subsubsec: appendix_model}
To set up a latent diffusion model $\M_{true}$, we first fix the graph structure and then generate the distributions $\N_u$ and $\T_{(u,v)}$. 
\begin{itemize}
	\item \textbf{Graphs.} We adopt four graphs: Kronecker,  Erdős-Rényi, Higgs, and Hep. Kronecker and Erdős-Rényi are two representative graphs for studying social networks \cite{seshadhri2011depth, bodine2010distance,seshadhri2012community}. Higgs is a subgraph of the network built after monitoring the spreading process of the messages on Twitter regarding the discovery of the elusive Higgs boson \cite{de2013anatomy}. Hep is a collaboration network of Arxiv High Energy Physics category. More information can be found in Table \ref{table: data}.
	
	\item \textbf{Distribution $\N_u$.} Following the independent cascade model \cite{kempe2003maximizing}, for each node $u$, the distribution $\N_u$ is designed by assuming that each in-neighbor $v$ of $u$ will be selected independently with a probability of $p_{(v,u)}$. In particular, following the weighted cascade model \cite{hong2019seeds}, we have $p_{(v,u)}^{true}  = 1/|N_u^-|$ in $\M_{true}$.

	\item \textbf{Distribution $\T_{(u,v)}$}. For each edge $e$, the distribution $\T_{(u,v)}$ is selected to be a Weibull distribution with two parameters $\alpha_{(u,v)}$ and $\beta_{(u,v)}$. In the underlying model $\M_{true}$,  $\alpha_{(u,v)}^{true}$ and $\beta_{(u,v)}^{true}$ are randomly selected from $\{1,...,10\}$, which is standard practice for having a diverse transmission time \cite{du2014influence, tong2020stratlearner}.
\end{itemize}
With the latent diffusion model $\M_{true}$, the objective functions of DE and DC become concrete. There are four possible settings of Problem \ref{problem: main}: DE-DE, DE-DC, DC-DE, and DC-DC. 

\textbf{Samples.} Given the latent diffusion model $\M_{true}$, we generate random samples for DE and DC, as follows. In generating one query-decision pair $(X, Y_{\M_{true}, T, X}^{\alpha})$ where $T$ is DE (resp., DC), the size of $X$ is sampled from a power law distribution with a parameter of $2.5$ scaled by $40$ (resp., 10), and the elements in $X$  are then randomly selected from $V$; the size of $Y_{\M_{true}, T, X}^{\alpha}$ is sampled from a power law distribution with a parameter of $2.5$ scaled by $10$, and the elements in $Y_{\M_{true}, T, X}^{\alpha}$ are computed by solving Equation \ref{eq: problem_abstract} using the greedy algorithm, which is a $(1-1/e)$-approximation (Theorem \ref{theorem: algorithm}). For each of DE and DC, we generate a pool of samples for training and testing; the size of the pool can be found in Table \ref{table: data}.

\begin{remark}
For DE, the rationale behind the selection of the size of $X$  is that $X$ cannot be either too small or too large. When $X$ is too small, selecting the neighbor of $X$ would result in a perfect influence to $X$, which makes the problem trivial; when $X$ is too large, solving the influence maximization problem over the entire graph would always produce a near-optimal solution, which again makes the problem trivial. 
\end{remark}

\subsubsection{Social-Inverse}
\label{subsubsec: appendix_SI}
To set up Social-Inverse, with $K$ and $\beta$ being the hyperparameters, we are left to specify the empirical distribution $\M_{em}$ for generating realizations. We consider the realizations generated by the following four diffusion models.
\begin{itemize}
	\item $\M_{q}$. $\M_{q}$ is obtained from $\M_{true}$ by using a different set of parameters (i.e., $p_{(u,v)}$, $\alpha_{(u,v)}$, and $\beta_{(u,v)}$ for each $(u,v) \in E$). In particular, for each $q \in \{0.1, 0.5, 1.0\}$, $p_{(u,v)}$, $\alpha_{(u,v)}$, and $\beta_{(u,v)}$ are sampled from uniform distributions, respectively, over \[ \Big[p_{(u,v)}^{true} \cdot (1-q), q \cdot p_{(u,v)}^{true} (1+q) \Big], \]  \[ \Big[\alpha_{(u,v)}^{true} \cdot (1-q), q \cdot \alpha_{(u,v)}^{true} (1+q)\Big], \] and \[ \Big[\beta_{(u,v)}^{true} \cdot (1-q), q \cdot \beta_{(u,v)}^{true} (1+q)\Big]. \]
	Clearly, smaller $q$ implies that $\M_{q}$ is closer to $\M_{true}$.
	\item $\M_{\infty}$. Similarly, $\M_{\infty}$ is obtained from $\M_{true}$ by using a different set of parameters. In particular, all the parameters are selected uniformly at random from $[0,1]$.
\end{itemize}
For each empirical distribution, we generate a pool of realizations (Table \ref{table: data}); given $K$, the realizations used in Social-Inverse are randomly selected from the pool we generate. The training process is done by the n-slack cutting plane algorithm \cite{taskar2005learning} via the implementation of Pystruct \cite{muller2014pystruct}. The hyperparameter $C$ is set as $0.01$ by default.

\subsubsection{Other methods}
\label{subsubsec: appendix_other}
Conceptually, when $T_\s = T_\t$, Problem \ref{problem: main} can be taken as a supervised learning task that leverages $S_{T_\s}^{\dal_\s}$ to build mapping from $\X_{\s}$ to $\Y_{\s}$. When the source task is different from the target task, the success of doing so in mitigating the task migrations hinges on the hypothesis that, for each $X\subseteq V$, the cascade that can maximally influence the nodes in $X$ may also be the cascade that can effectively block the negative cascade from $X$, which is however not true in some cases. Therefore, as seen from the experimental results, standard learning methods can perform poorly, especially when $T_\s \neq T_\t$.

\textbf{Naive Bayes (NB)}. NB solves this problem by learning $\Pr[X|Y]\Pr[Y]$ for each $Y \in \Y_{\t}$ and $X \in \X_{\t}$. In addition to the Naive Bayes assumption, we assume the nodes in $Y$ are also independent, and therefore, we are to learn $ \prod_{u\in X} \prod_{v \in Y} \Pr[u|v]\Pr[v]$ by learning a collection of Bernoulli distributions.

\textbf{Graph neural network (GNN).} GNN is a reasonable choice because the graph structure of $G$ is known to us. For GNN, we adopt the standard implementation  \cite{kipf2016semi} including two graph convolution
layers followed by a dropout layer with a ratio of $0.4$ and a sequence of MLP of size (512, 512, 256) with ReLU as the activation function. The input and output are coded as one-hot vectors. The parameters are computed by minimizing the binary cross-entropy, using an Adam optimizer with a learning rate of $0.01$ and an L2 regularizer. The seeds are selected from $\{7,11,17,23,43\}$.

\textbf{Deep set prediction network (\textbf{DSPN})}. DSPN \cite{zhang2019deep} is a deep architecture designed specifically for dealing with set structures, and it is therefore suitable for our purpose as Problem \ref{problem: main} seeks a mapping between subsets. In implementing DSPN, the best configuration observes an outer learning rate of $0.01$, an inner learning rate of $10^7$, and an inner iteration of $10$, with the set encoder being an MLP encoder of dimension $1024$ and the input encoder being an MLP encoder of dimension $256$.

\textbf{High-Degree (HD)}. Given the size of the decision, the HD method selects the nodes with the highest degree. 

\textbf{Random.} This is a baseline method that outputs a decision by randomly selecting nodes.

\subsubsection{Training, testing, and evaluation}
\label{subsubsec: appendix_train}
Given the training size and the testing size, the samples are randomly selected from the pool we generate, and each method is executed once. The entire process is repeated five times, and we report the average performance ratio together with the standard deviation. 

\textbf{Performance ratio.} When the target task is DE, for a testing sample $(X,Y)$ and a prediction $\hat{Y}$, the performance is measured by the ratio \[f_{\M_{true}}^{\DE}(X,\hat{Y})/f_{\M_{true}}^{\DE}(X,Y);\] when the target task is DC, the performance is measured by the ratio 
\begin{align*}
\frac{f_{\M_{true}}^{\DC}(X,\hat{Y})-f_{\M_{true}}^{\DC}(X,\emptyset)}{f_{\M_{true}}^{\DC}(X,Y)-f_{\M_{true}}^{\DC}(X,\emptyset)},
\end{align*} 
which means that the performance is measured by the improvement based on the empty decision. In any case, a larger ratio implies a better performance. It is possible that the ratio is occasionally larger than $1$ because the sample decisions are approximate solutions.

\subsection{Time complexity of Social-Inverse}
\label{subsec: appendix_time}
The training complexity of Social-Inverse follows immediately from the analysis of the n-slack cutting plane algorithm \cite{joachims2009cutting,tsochantaridis2005large}. Experimentally, the training cost of Social-Inverse is presented in Table \ref{table: time1}, and the cost of making one prediction is less than 1s in all cases. Our experiments are executed on Amazon EC2 C5 Instance with 96 virtual CPUs and a memory of 192G. As we can see from the tables, the training cost is almost linear in $K$ as well as linear in the training size. The training is reasonably fast in all cases. Under the same setting, DC-DE takes more time on $M_{0,1}$ than DE-DC does, but DC-DE is slightly faster than DE-DC on $\M_{\infty}$. 

\begin{table}[h]
	\renewcommand{\arraystretch}{1.3} 
	\small
	\caption{\textbf{Training time.} The training time denotes the cost (measured by sec) used for computing $\fw$ under one setting.}
	\centering
	\label{table: time1}
	\begin{tabular}{  @{}  c  @{\hspace{2mm}}     c @{\hspace{2mm}}     c @{\hspace{2mm}}  r @{\hspace{2mm}} r @{\hspace{2mm}} r @{\hspace{2mm}}   r  @{\hspace{8mm}}   r @{\hspace{2mm}} r @{\hspace{2mm}} r @{\hspace{2mm}}   r    @{}  }
		\toprule
		& & & \multicolumn{4}{@{} c @{\hspace{0mm}} }{\textbf{DC-DE}} &  \multicolumn{4}{@{} c @{\hspace{0mm}} }{\textbf{DE-DC}}\\
		& &   &  $5$ &  $15$   & $30$ & $60$  &   $5$ &  $15$   & $30$ & $60$ \\
		\midrule
		\midrule
		 \multirow{6}{*}{\textbf{Kro}}  & \multirow{3}{*}{$\M_{0.1}$} & 90  &	  13.839&    33.331	  &  69.675	   		&   133.554 	   & 12.207	& 23.723 & 49.753 & 115.068  \\
		& & 270& 37.151 &   75.557& 	 152.075	&    333.557 & 20.674&  40.084 &88.198 & 217.873  \\
		& & 1350 &  138.313 &   369.822& 	 766.713	&    1519.132&    109.309 &292.122 &382.564 & 906.334 \\
			\cmidrule{2-11}
		& \multirow{3}{*}{$\M_{\infty}$} & 90  &	 8.545& 14.579  & 23.599	  & 47.149  	 	&  7.990     &12.488 & 19.890& 39.480	\\
		& 	& 270& 18.278& 31.937  &55.734& 	 112.496    	&   16.300  &  26.043 & 44.581  &90.024 \\
		&	& 1350 &  86.981& 160.363  &259.810& 	 497.938	&   93.215   & 135.816 & 252.990 & 474.851 \\		
		\midrule
		\multirow{6}{*}{\textbf{ER}}  & \multirow{3}{*}{$\M_{0.1}$} & 90  &	 5.707 & 9.197& 20.420  & 29.963 & 5.324 & 7.235 & 13.739 & 21.746 \\
		& & 270& 11.977 & 21.634& 38.295 & 71.406 & 9.618 &16.201 & 26.176 & 58.331 \\
		& & 1350 & 88.852 & 103.064 &167.285 & 320.701 & 84.946 & 72.091 & 132.034 & 218.223\\
		\cmidrule{2-11}
		& \multirow{3}{*}{$\M_{\infty}$} & 90  & 4.834	& 7.695 & 12.402 & 21.230 & 4.460 & 6.544 & 10.689 & 18.063\\
		& 	& 270&  9.581 & 17.026 & 28.196 & 50.213 & 8.836 & 14.301 & 22.206 & 40.461 \\
		&	& 1350 & 39.967  & 82.883 & 133.520  & 233.780 & 50.439 & 119.346 &112.180 & 190.187 \\
		
				\midrule
		\multirow{6}{*}{\textbf{Higgs}}  & \multirow{3}{*}{$\M_{0.1}$} & 90  & 89.026	& 187.430 & 364.440 & 709.585 &84.301  & 174.703 & 328.194 & 676.772 \\
		& & 270&  201.856  & 433.293 & 811.732 & 1626.409 & 186.414 & 394.361 & 732.861 & 1462.641   \\
		& & 1350 &  869.795 & 1897.335 & 3685.378 & 7206.658 & 929.341 & 1698.253 & 3301.660 & 6289.700  \\
		\cmidrule{2-11}
		& \multirow{3}{*}{$\M_{\infty}$} & 90  & 	85.827 & 182.209 & 337.944 & 652.565 & 81.637 & 168.980 & 312.979 & 600.219\\
		& 	& 270&  193.269 & 413.589 & 785.807 & 1528.932 & 184.091 & 381.032 & 704.073 & 1363.695\\
		&	& 1350 &  913.599 & 1852.769 &3474.103 & 6807.414 & 800.248  & 1721.938 & 3058.026 & 5941.948 \\
		
		\midrule
		\multirow{3}{*}{\textbf{Hep}}  & \multirow{3}{*}{$\M_{\infty}$} & 90  & 145.346 & 309.781 & 573.465 & 1113.464 & 140.602 & 293.864 & 534.508 & 1039.320 \\
		& & 270& 332.663  &700.047  & 1307.256 & 2567.882 & 318.912 & 651.294 & 1212.271 & 2346.511 \\
		& & 1350 &   911.897  & 1898.069 & 3538.037 & 7042.835 & 843.958 & 1782.247 & 3180.850 & 6251.398 \\
		
		\bottomrule
	\end{tabular}
\end{table}

\subsection{The impact of $\beta$. }
\label{subsec: appendix_beta}
\begin{figure}[h]
	\centering
	
	\subfloat[{[Kro, DC-DE, $\M_{0.1}$, $30$, $270$]}]{\label{fig: kro_dc_de_10_30_270_beta}\includegraphics[width=0.33\textwidth]{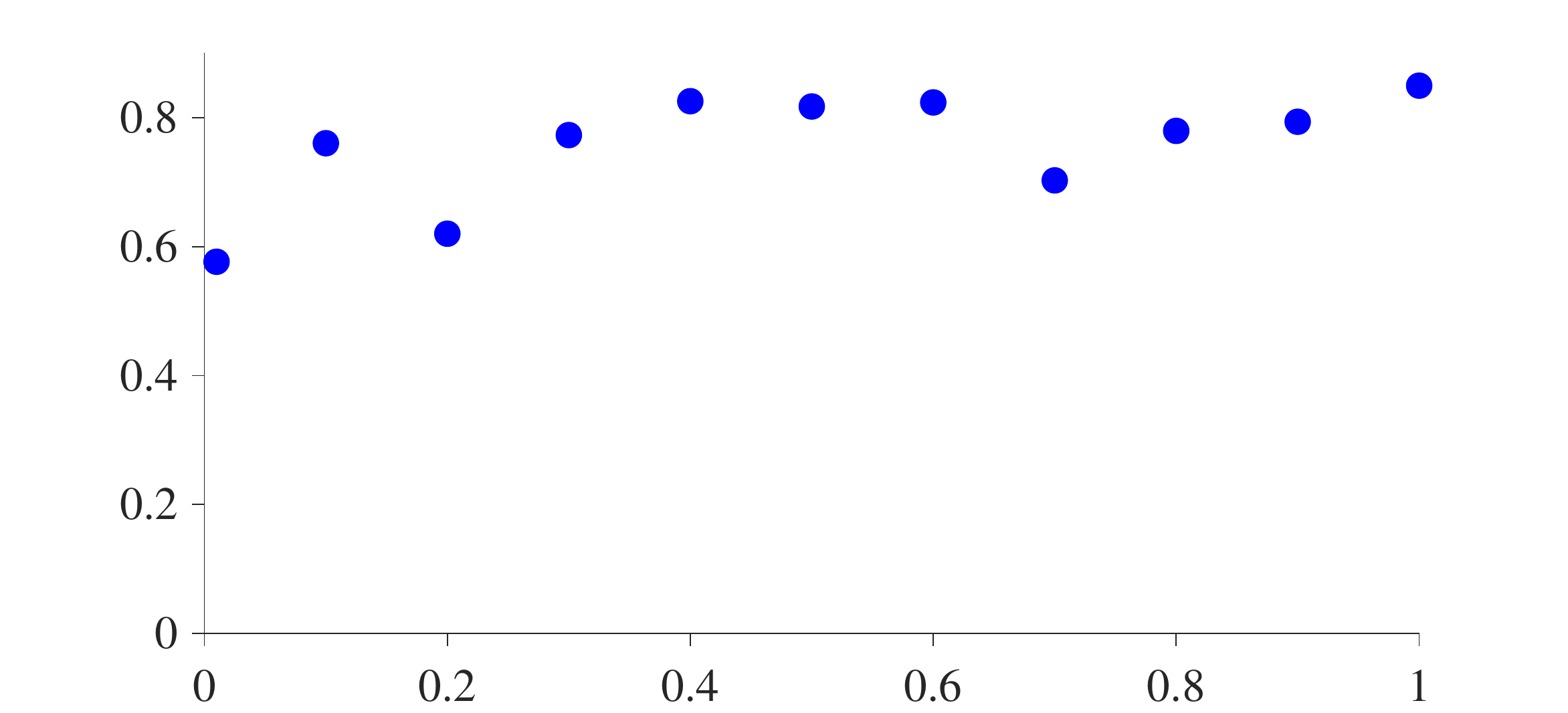}}\hspace{5mm} 
	\subfloat[{[Kro, DC-DE, $\M_{0.5}$, $30$, $270$]}]{\label{fig: kro_dc_de_50_30_270_beta}\includegraphics[width=0.33\textwidth]{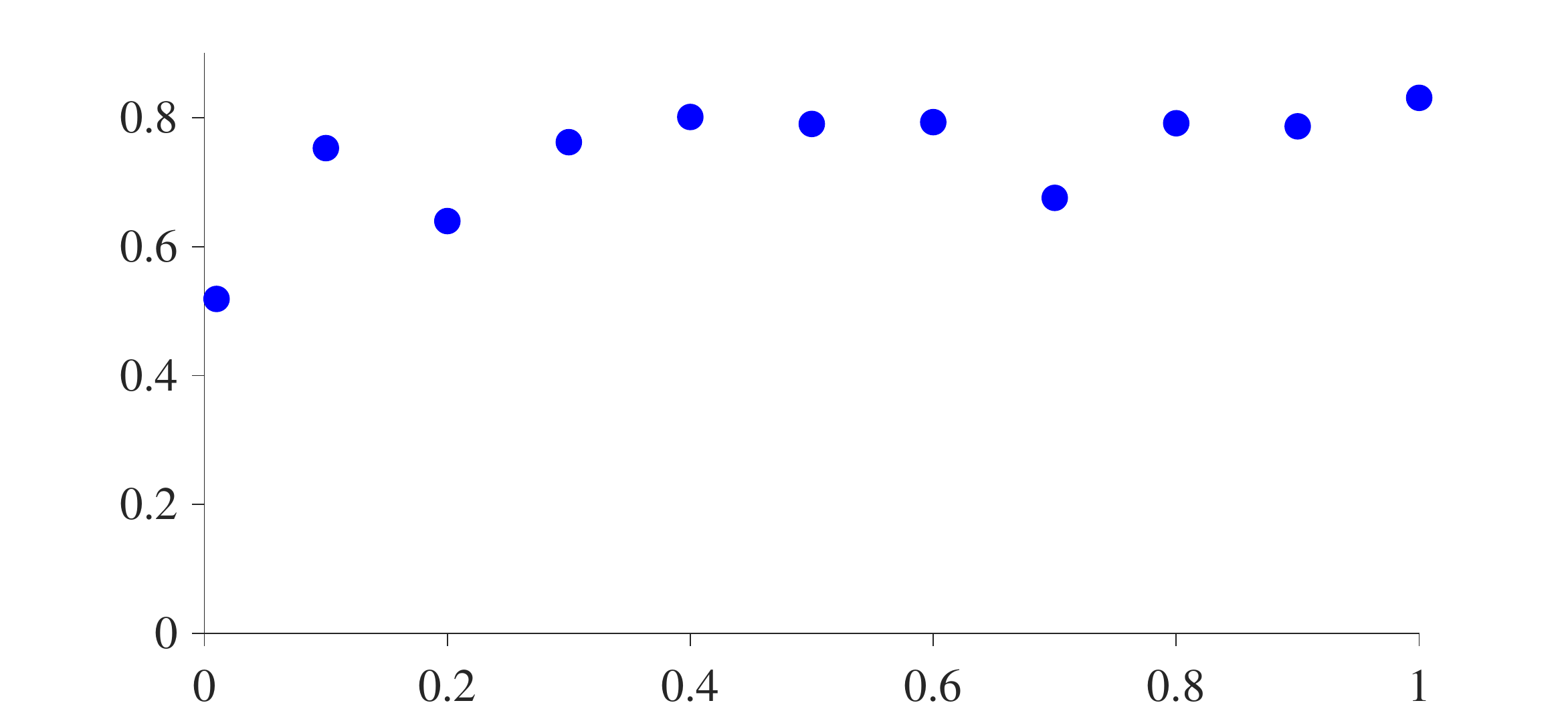}}\hspace{0mm}
	
	\subfloat[{[ER, DC-DE, $\M_{\infty}$, $30$, $1350$]}]{\label{fig: er_dc_de_uni_30_270_beta}\includegraphics[width=0.33\textwidth]{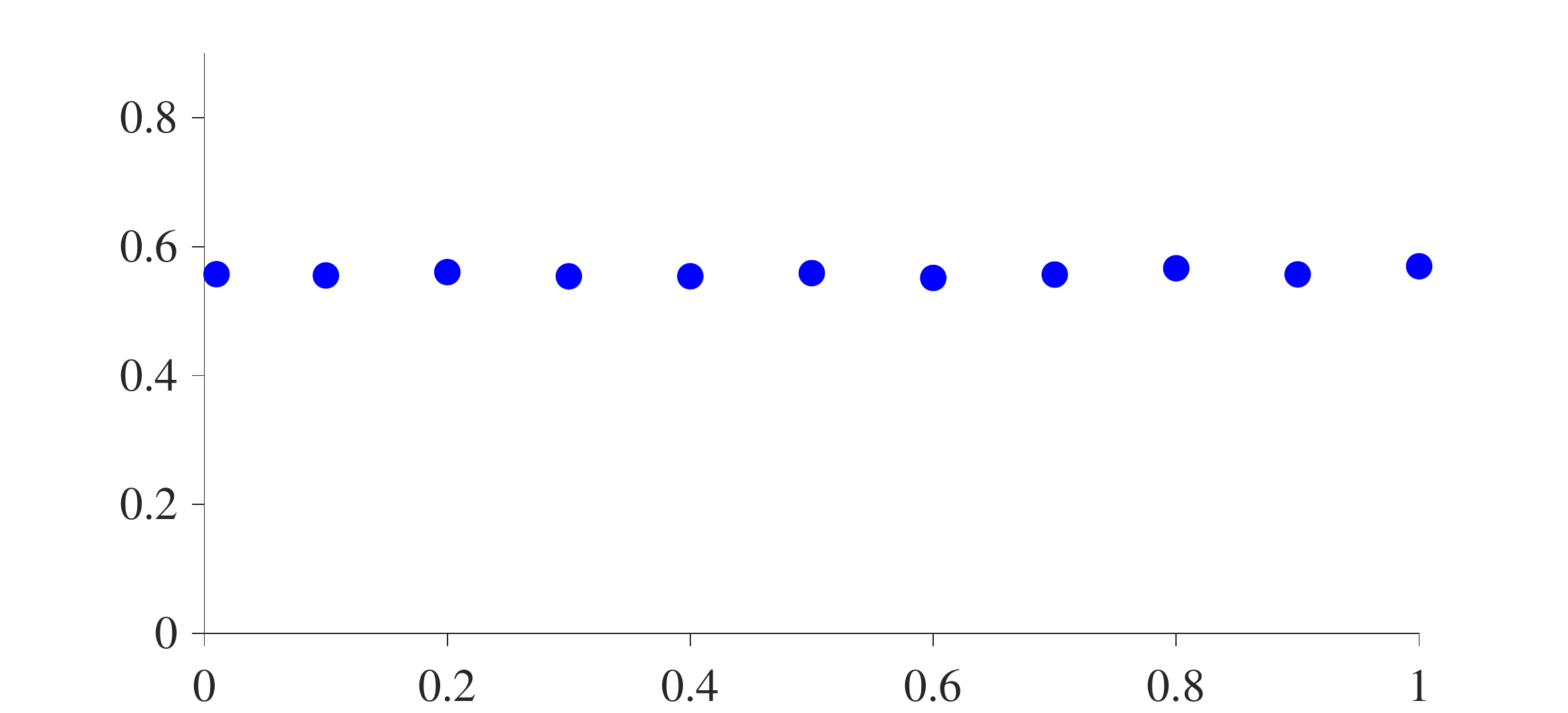}}\hspace{5mm} 
	\subfloat[{[Higgs, DE-DC, $\M_{\infty}$, $5$, $270$]}]{\label{fig: higgs_de_dc_uni_5_270_beta}\includegraphics[width=0.33\textwidth]{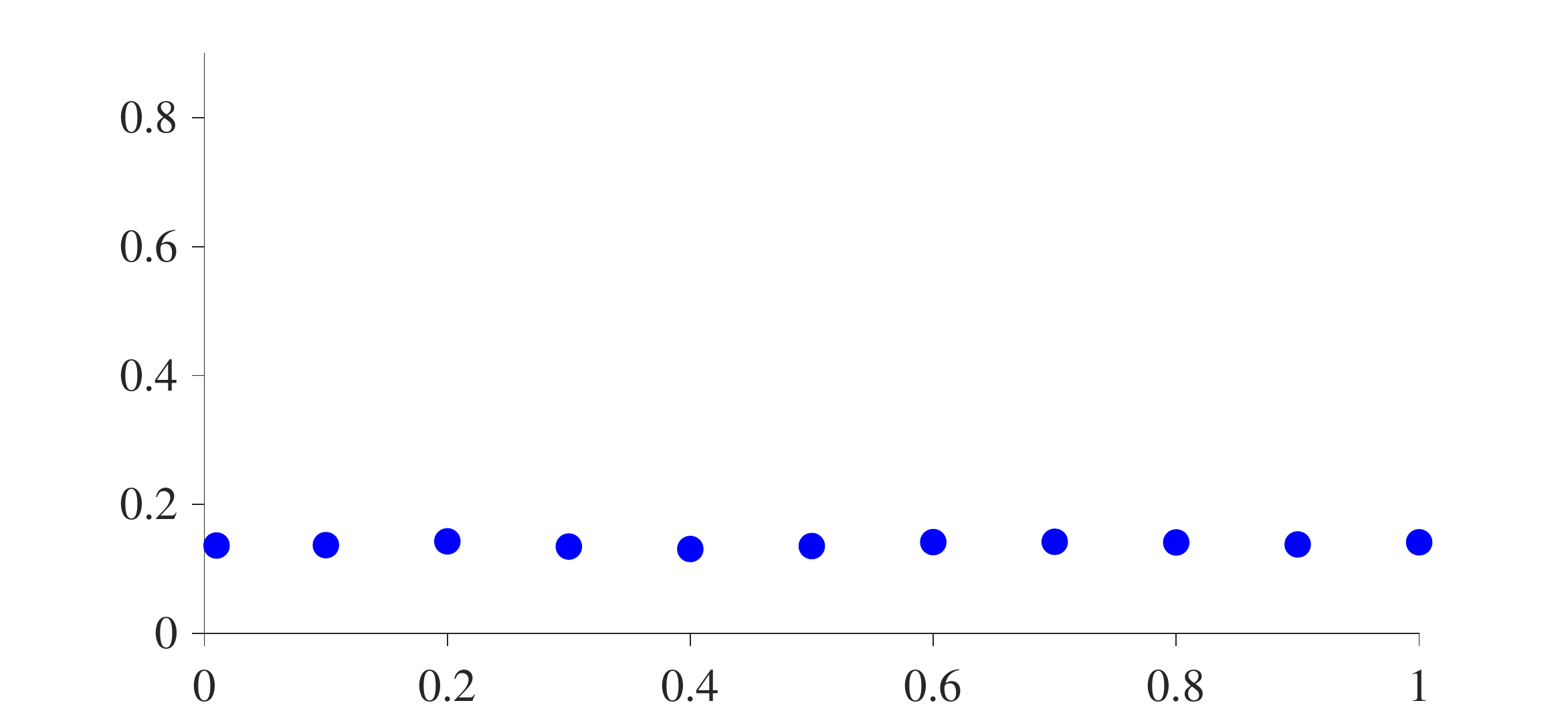}}\hspace{0mm} 
	
	\caption{Each subgraph is labeled by [graph, source-target, empirical distribution, K, training size]. The x-axle denotes $C^{*}$ and the y-axle denotes the performance ratio.}
	\label{fig: beta}
\end{figure}

To explore the impact of $\beta$, we integrate the coefficient $\ial_\t (\ial_\t - \beta)$ as a constant $C^*$ in running the n-slack cutting plane algorithm, and examine the performance of Social-Inverse with different $C^*$ selected from $\{0.01, 0.1, 0.2, 0.3, 0.4 ,0.5, 0.6 , 0.7, 0.8, 0.9 ,1\}$. We are particularly interested in the case where the efficacy of samples is evident in mitigating the task migrations. The results of this part are given in Figure \ref{fig: beta}. According to the figures, the performance on the Kronecker graph is slightly sensitive to $C^*$, and larger $C^*$ in general gives better results. On Erdős-Rényi and Higgs, the performance does not vary much with $C^*$ as long as $C^*$ is larger than zero. 

\newpage

\subsection{More experimental results}
\label{subsec: appendix_full_observation}

\subsubsection{Full results on Kronecker}
\begin{figure}[h]
	\centering
	\subfloat{\includegraphics[width=0.4\textwidth]{images/legend.pdf}}
	\vspace{-2mm}
	\addtocounter{subfigure}{-1}
	
	\subfloat[{[DC-DE, $\M_{0.5}$, $m=90$]}]{\label{fig: kro_dc_de_50_30_90}\includegraphics[width=0.31\textwidth]{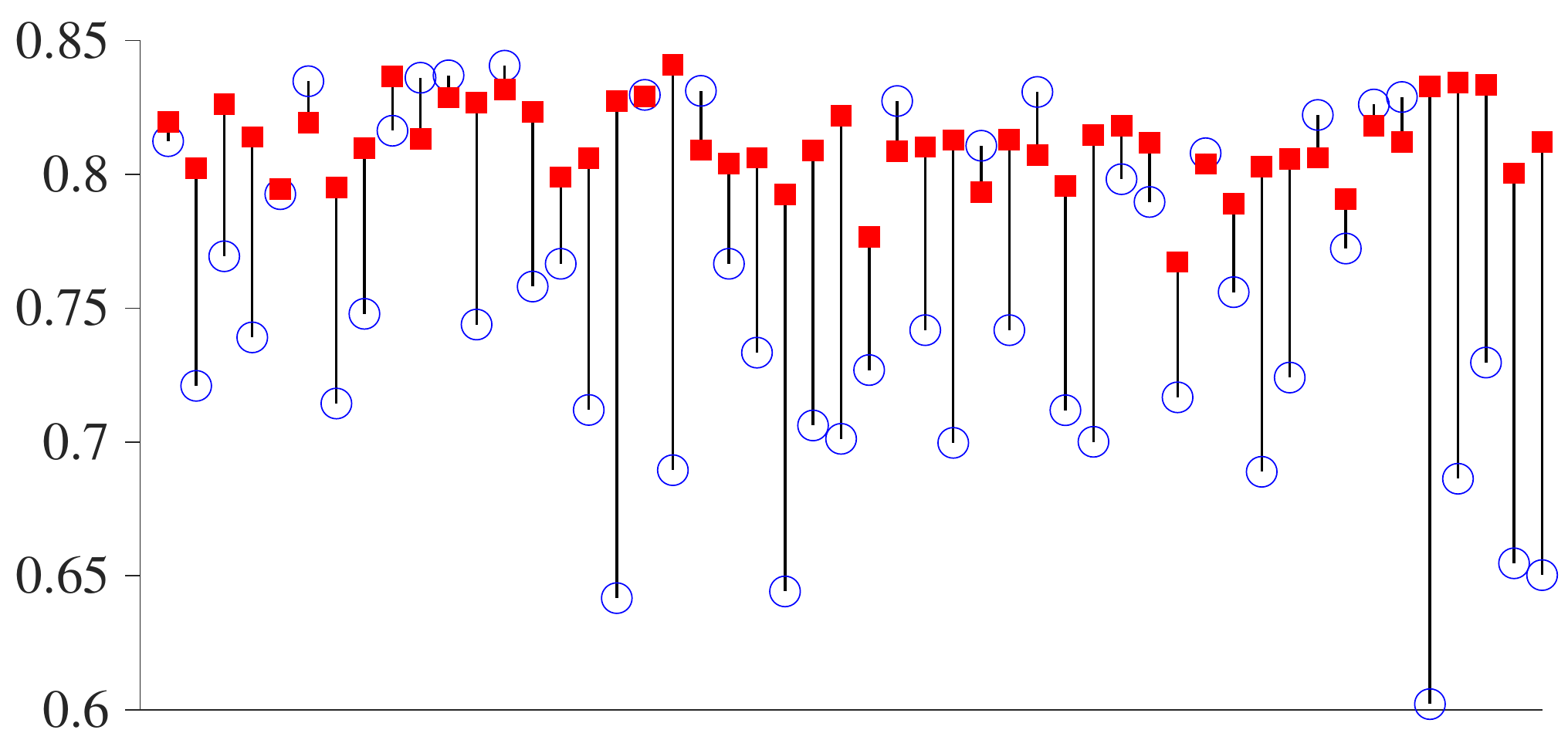}}\hspace{2mm} 
	\subfloat[{[DC-DE, $\M_{0.5}$, $m=270$]}]{\label{fig: kro_dc_de_50_30_270}\includegraphics[width=0.31\textwidth]{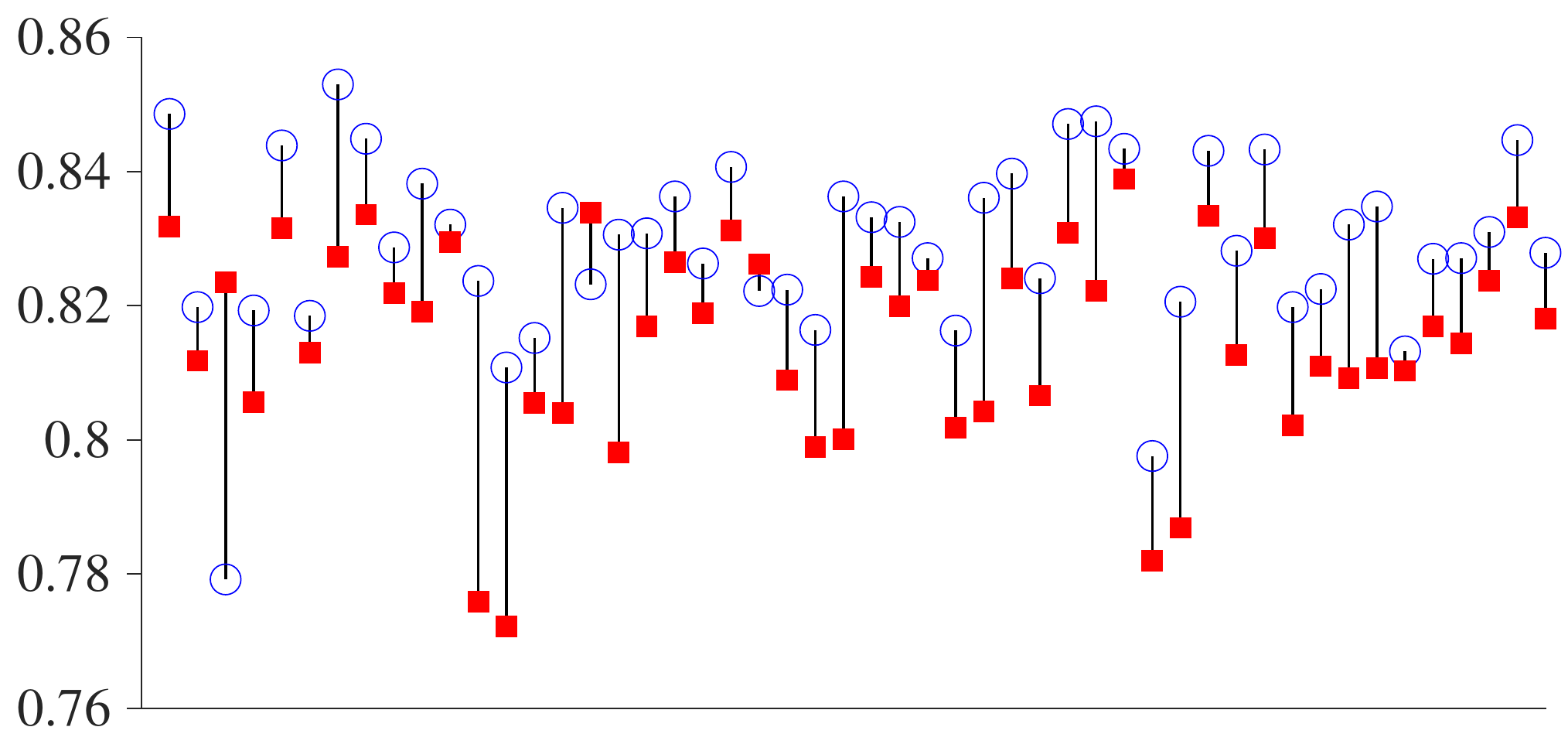}}\hspace{2mm}
	\subfloat[{[DC-DE, $\M_{0.5}$, $m=1350$]}]{\label{fig: kro_dc_de_50_30_2700}\includegraphics[width=0.31\textwidth]{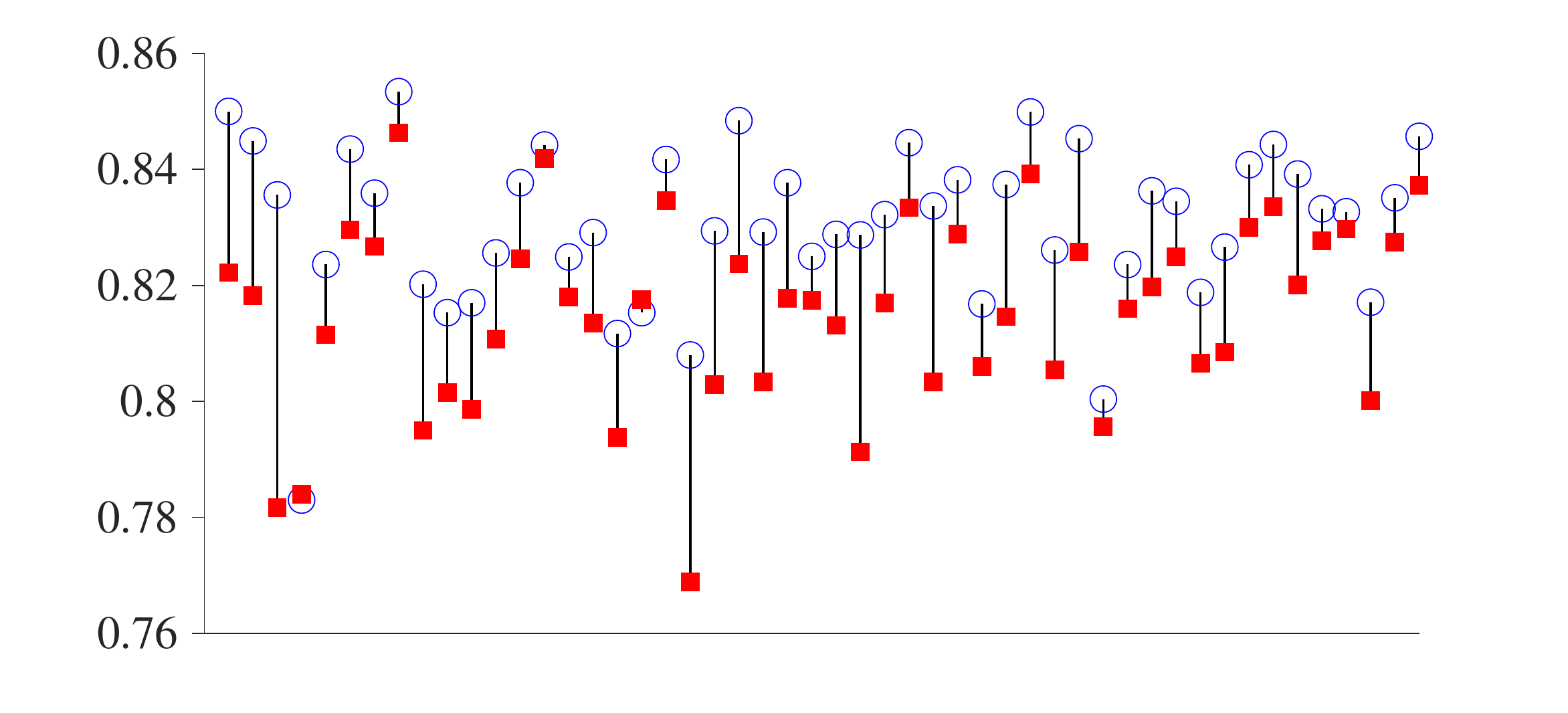}}

	\subfloat[{[DE-DC, $\M_{0.1}$, $m=270$]}]{\label{fig: kro_de_dc_10_30_270}\includegraphics[width=0.31\textwidth]{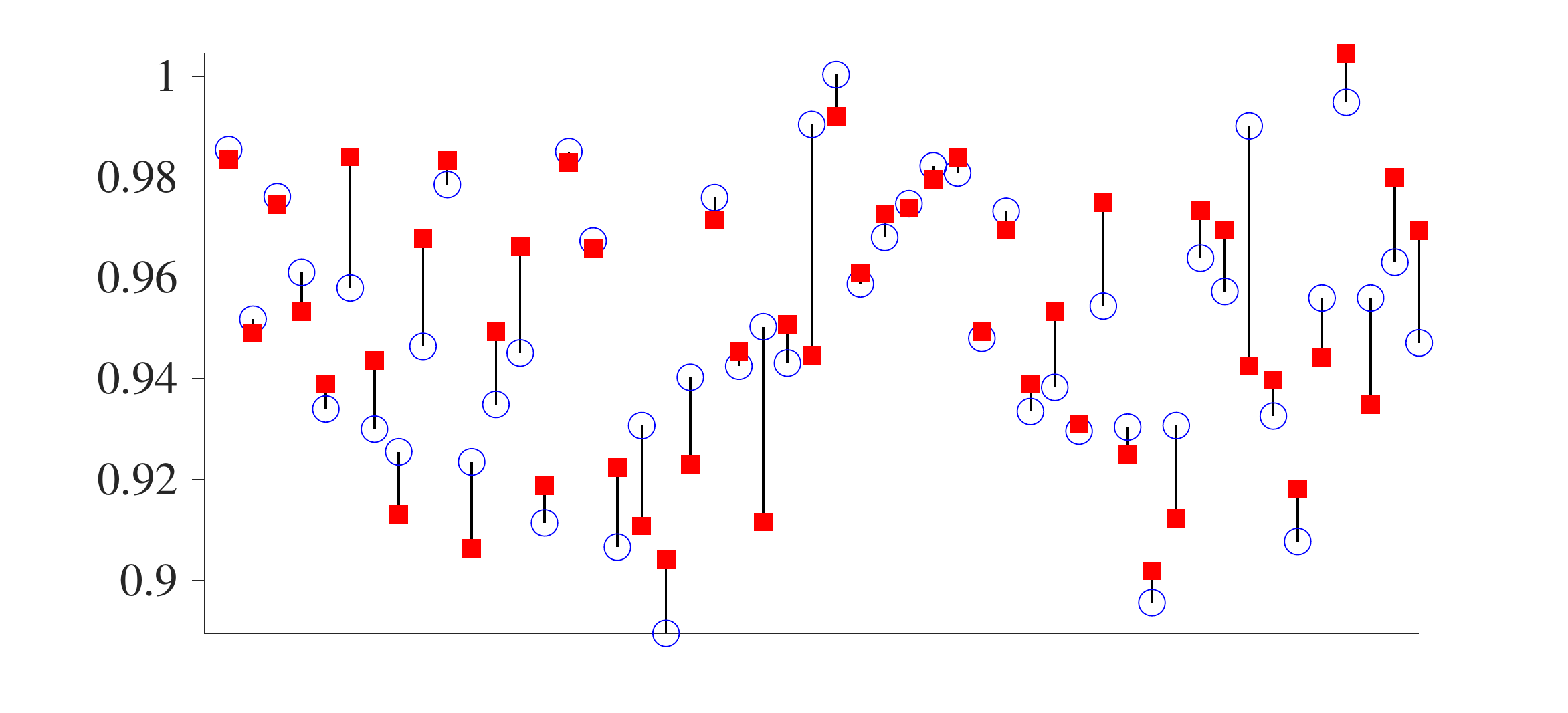}}\hspace{2mm} 
	\subfloat[{[DE-DC, $\M_{0.5}$, $m=270$]}]{\label{fig: kro_de_dc_50_30_270}\includegraphics[width=0.31\textwidth]{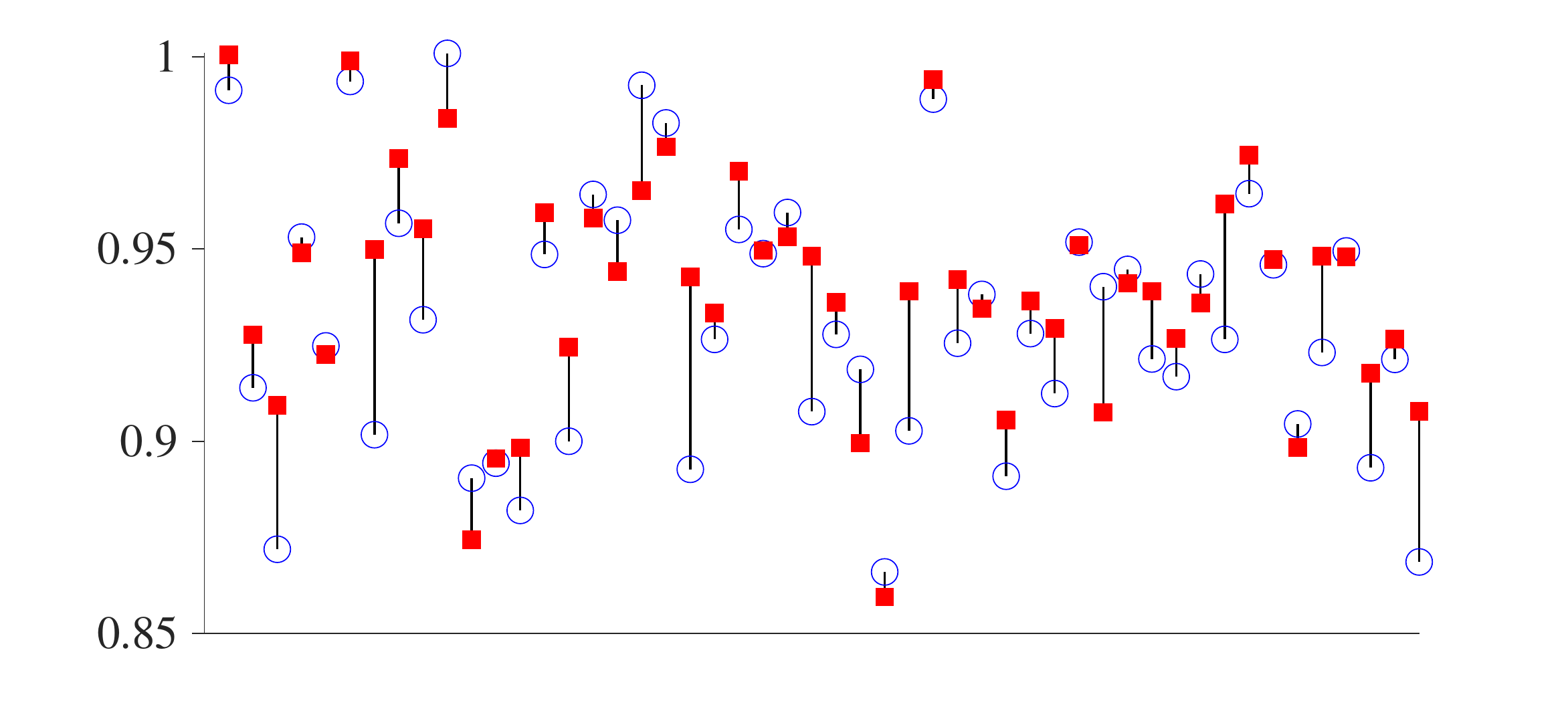}}\hspace{2mm} 
	\subfloat[{[DE-DC, $\M_{\infty}$, $m=270$]}]{\label{fig: kro_de_dc_uni_30_270}\includegraphics[width=0.31\textwidth]{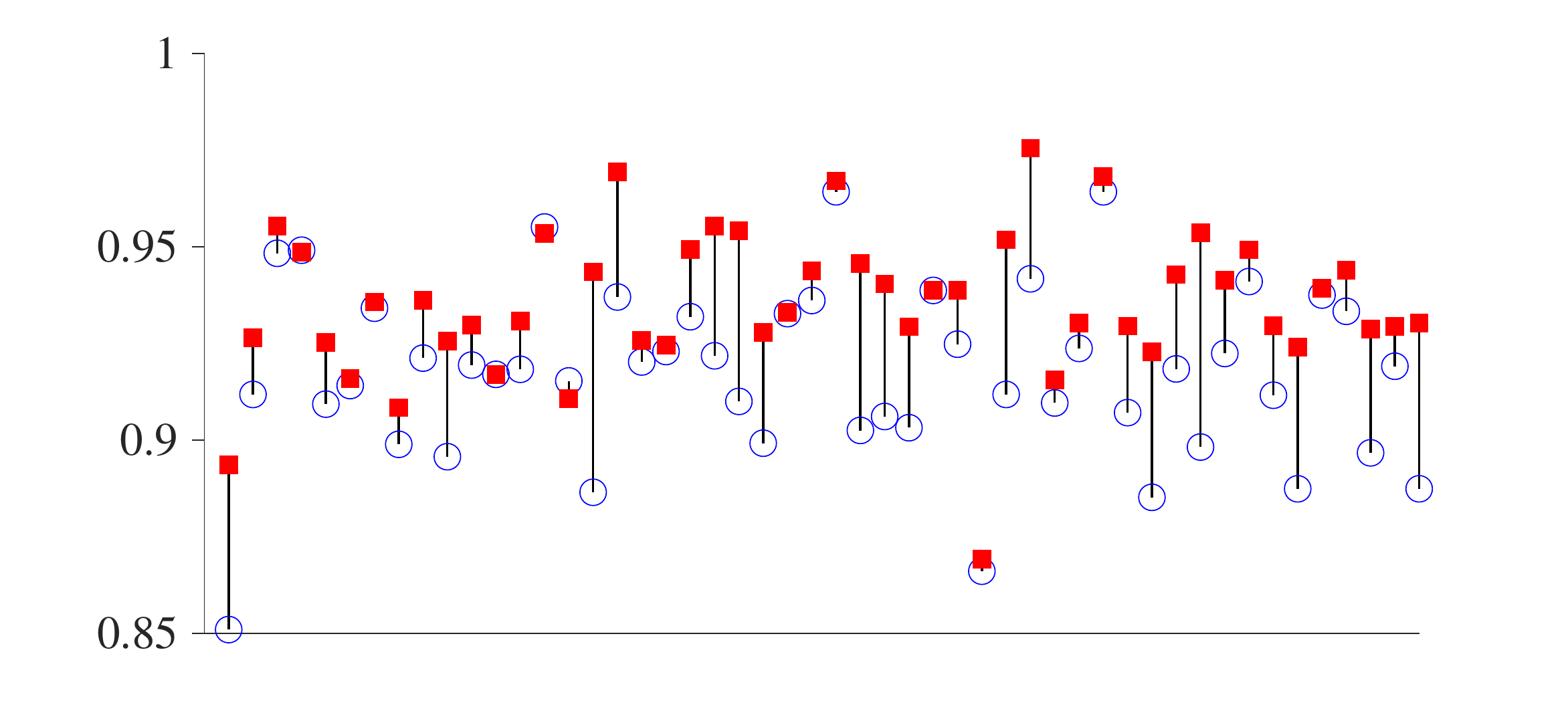}}
	
	\caption{Each subgraph shows the results of 50 testing queries on the Kronecker graph with $K=30$. For each query, we report the quality (i.e. performance ratio) of the predictions made based on the initial weight (before optimization) and final weight (after optimization).}
	\label{fig: kro_more}
\end{figure}

The full results on the Kronecker graph are given in Table \ref{table: kro} and Figure \ref{fig: kro_more}. The output of HD and Random are independent of the training samples, and they therefore have the results for DE-DE and DC-DE; similar arguments apply to DC-DC and DE-DC. The last column shows the results of GNN under different seeds.  In addition to the observations mentioned in the main paper, we see that the ratios of DE-DE (resp., DC-DC) are clearly better than those in DC-DE (resp., DE-DC), which coincides with the expectation that samples from the same task are naturally more effective than those from another task. It is also worth noting that NB is effective in the case of DC-DC. According to the figures, the efficacy of DC samples in solving DE is quite evident. However, under the same setting, the DE samples are not very useful in providing a high-quality weight $\fw$; one plausible reason is that the performance ratio is already too high (when $K=30$) so that there is little room for improvement.

\begin{table}[h]
	\renewcommand{\arraystretch}{1.2} 
	\small
	\caption{\textbf{Full results on Kronecker.} Each cell presents the mean of performance ratio with std.}
	\centering
	\label{table: kro}
	\begin{tabular}{  @{}  c  @{\hspace{2mm}}     c @{\hspace{2mm}}  c @{\hspace{1mm}} c @{\hspace{1mm}} c @{\hspace{1mm}}  c  @{\hspace{1mm}} c  @{\hspace{4mm}}   c @{\hspace{1mm}} c @{\hspace{2mm}} c    @{}  }
		\toprule
		&\multicolumn{6}{@{} c @{\hspace{0mm}} }{\textbf{Social-Inverse}} &  \multicolumn{3}{c}{ \multirow{1}{*}{\textbf{Other Methods}} } \\
		&  $K$ &  $5$ &  $15$   & $30$ & $60$ & $180$ & & &\multicolumn{1}{c}{GNN}  	\\
		\midrule
		\midrule
		\multirow{4}{*}{\textbf{\makecell[c]{DE\\|\\DE}}} & $\M_{\infty}$  &	  0.482{\tiny (0.015)}&    0.528{\tiny (0.018)} 	  &  0.556{\tiny (0.011)} 	   		&   0.578{\tiny (0.010)}   	&   0.612{\tiny (0.002)}     & NB	& DSPN &  0.431 (23) 	\\
		& $\M_{1}$ &  0.592{\tiny (0.022)}  &   0.748{\tiny (0.015)} & 	 0.776{\tiny (0.008)} 		&    0.810{\tiny (0.004)} 	&    0.802{\tiny (0.012)}   &    0.644{\tiny (0.006)} &  0.456{\tiny (0.140)} &   0.545 (17)\\
		& $\M_{0.5}$ &  0.644{\tiny (0.038)}  &   0.784{\tiny (0.012)} & 	 0.824{\tiny (0.016)} 		&    0.846{\tiny (0.011)} &   0.856{\tiny (0.003)}  & HD & Random   &  0.486 (11)\\
		& $\M_{0.1}$ &  0.665{\tiny (0.049)}  &   0.781{\tiny (0.040)} & 	 0.817{\tiny (0.025)} 		&    0.872{\tiny (0.008)} & 0.880{\tiny (0.005)}   &    0.589{\tiny (0.003)}   &    0.150{\tiny (0.014)} & 0.521 (43)\\
		
		\cmidrule{1-10}

	\multirow{4}{*}{\textbf{\makecell[c]{DC\\|\\DE}}} & $\M_{\infty}$  &	 0.489{\tiny (0.010)} &    0.540{\tiny (0.014)} 	  &  0.567{\tiny (0.016)} 	   		&   0.584{\tiny (0.006)}   	&   0.606{\tiny (0.008)}     & NB	& DSPN &   0.459 (23)  	\\
		& $\M_{1}$ &  0.654{\tiny (0.031)}  &   0.748{\tiny (0.022)} & 	 0.794{\tiny (0.016)} 		&    0.811{\tiny (0.011)} 	&    0.842{\tiny (0.026)}    &    0.584{\tiny (0.005)} & 0.434{\tiny (0.056)} &  0.516 (17) \\
		& $\M_{0.5}$ &  0.653{\tiny (0.035)}  &   0.767{\tiny (0.028)} & 	 0.837{\tiny (0.004)} 		&    0.851{\tiny (0.004)} & 0.850{\tiny (0.021)} & HD & Random   & 0.515 (11)  \\
		& $\M_{0.1}$ &  0.591{\tiny (0.036)}  &   0.770{\tiny (0.025)} & 	 0.850{\tiny (0.018)} 		&    0.853{\tiny (0.040)} &  0.888{\tiny (0.007)} &    0.589{\tiny (0.003)}   &    0.150{\tiny (0.014)} & 0.509 (43)\\
		
		\cmidrule{1-10}

	\multirow{4}{*}{\textbf{\makecell[c]{DE\\|\\DC}}}  & $\M_{\infty}$  &	 0.759{\tiny (0.013)} &    0.846{\tiny (0.005)} 	  &  0.913{\tiny(0.005)} 	   		&   0.985{\tiny (0.014)}   	&   1.069{\tiny (0.014)}     & NB	& DSPN &    0.660 (23)	\\
		& $\M_{1}$ &  0.720{\tiny (0.030)}  &   0.796{\tiny (0.056)} & 	 0.845{\tiny (0.040)} 		&    0.899{\tiny (0.030)} 	&    0.935{\tiny (0.028)}   &    0.877{\tiny (0.006)} & 0.567{\tiny (0.237)}&   0.835 (17) \\
		& $\M_{0.5}$ &  0.708{\tiny (0.039)}  &   0.850{\tiny (0.036)} & 	 0.937{\tiny (0.058)} 		&    1.021{\tiny (0.040)} &  1.104{\tiny (0.025)}  & HD & Random & 0.731 (11)    \\
		& $\M_{0.1}$ &  0.682{\tiny (0.037)}  &   0.862{\tiny (0.014)} & 	 0.953{\tiny (0.018)} 		&    1.041{\tiny (0.026)} & 1.137{\tiny (0.026)}  &    0.886{\tiny (0.006)}     &   0.258{\tiny (0.003)} & 0.794 (43)    \\
		\cmidrule{1-10}

		\multirow{4}{*}{\textbf{\makecell[c]{DC\\|\\DC}}} & $\M_{\infty}$  &	 0.760{\tiny (0.020)} &    0.860{\tiny (0.021)} 	  &  0.933{\tiny (0.021)} 	   		&   1.002{\tiny (0.018)}   	&   1.063{\tiny (0.005)}     & NB	& DSPN  &   0.751 (23)	\\
		& $\M_{1}$ &  0.706{\tiny (0.008)}  &   0.853{\tiny (0.026)} & 	 0.959{\tiny (0.014)} 		&    1.052{\tiny (0.012)} 	&    1.099{\tiny (0.010)}   &    0.922{\tiny (0.009)} & 0.585{\tiny (0.131)}&  0.815 (17)  \\
		& $\M_{0.5}$ &  0.732{\tiny (0.025)}  &   0.892{\tiny (0.029)} & 	 0.995{\tiny (0.014)} 		&    1.097{\tiny (0.018)} & 1.169{\tiny (0.010)} & HD & Random & 0.779 (11)    \\
		& $\M_{0.1}$ &  0.720{\tiny (0.022)}  &   0.872{\tiny (0.024)} & 	 1.002{\tiny (0.013)} 		&    1.093{\tiny (0.012)} &  1.173{\tiny (0.012)}  &     0.886{\tiny (0.006)}     &   0.258{\tiny (0.003)}  & 0.806 (43) \\
		\bottomrule
	\end{tabular}
	\vspace{-2mm}
\end{table}

\newpage
\subsubsection{Full results on Erdős-Rényi}
\begin{figure}[h]
	\centering
	\subfloat{\includegraphics[width=0.4\textwidth]{images/legend.pdf}}
	\addtocounter{subfigure}{-1}
	
	\subfloat[{[DC-DE, $\M_{0.1}$, $m=270$]}]{\label{fig: er_dc_de_10_30_270}\includegraphics[width=0.33\textwidth]{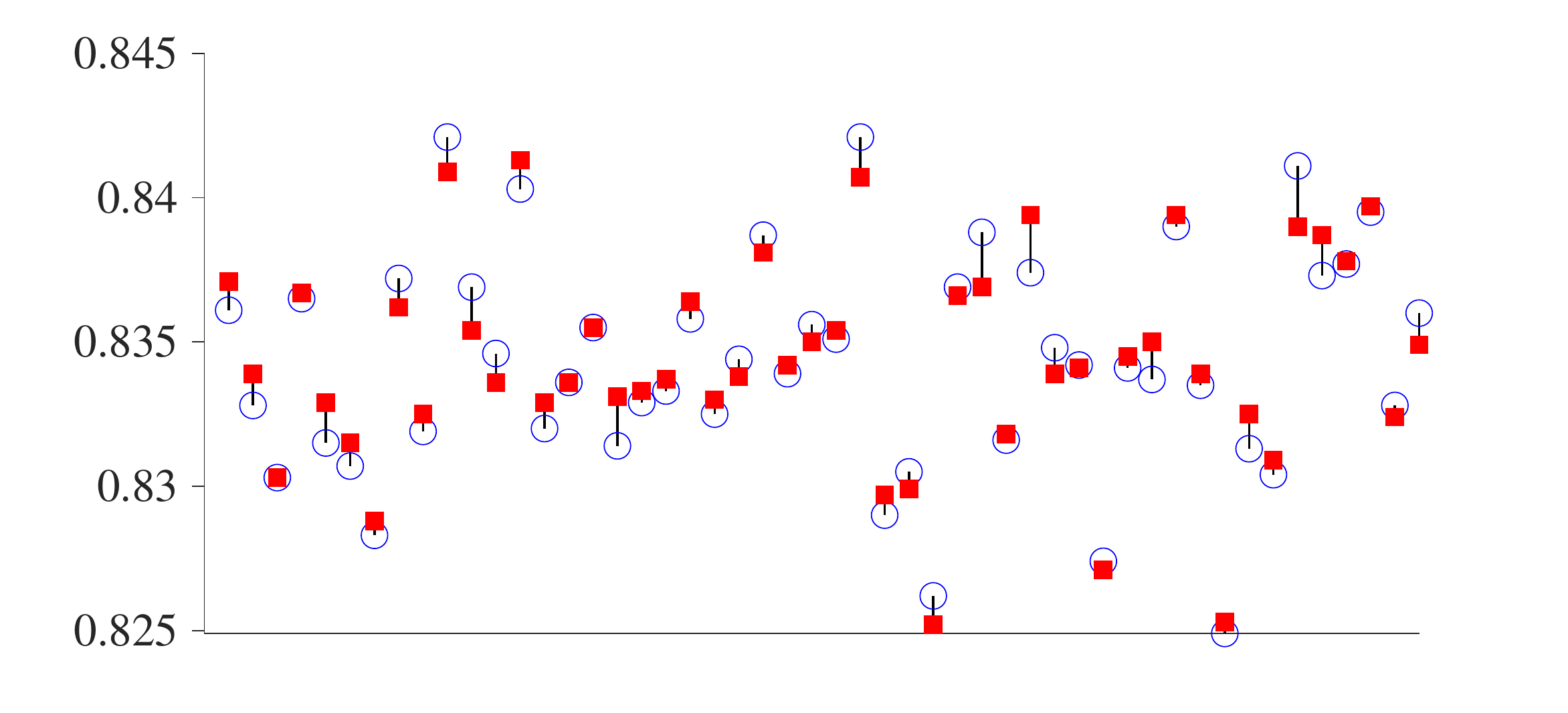}} 
	\subfloat[{[DE-DC, $\M_{0.1}$, $m=270$]}]{\label{fig: er_de_dc_10_30_270}\includegraphics[width=0.33\textwidth]{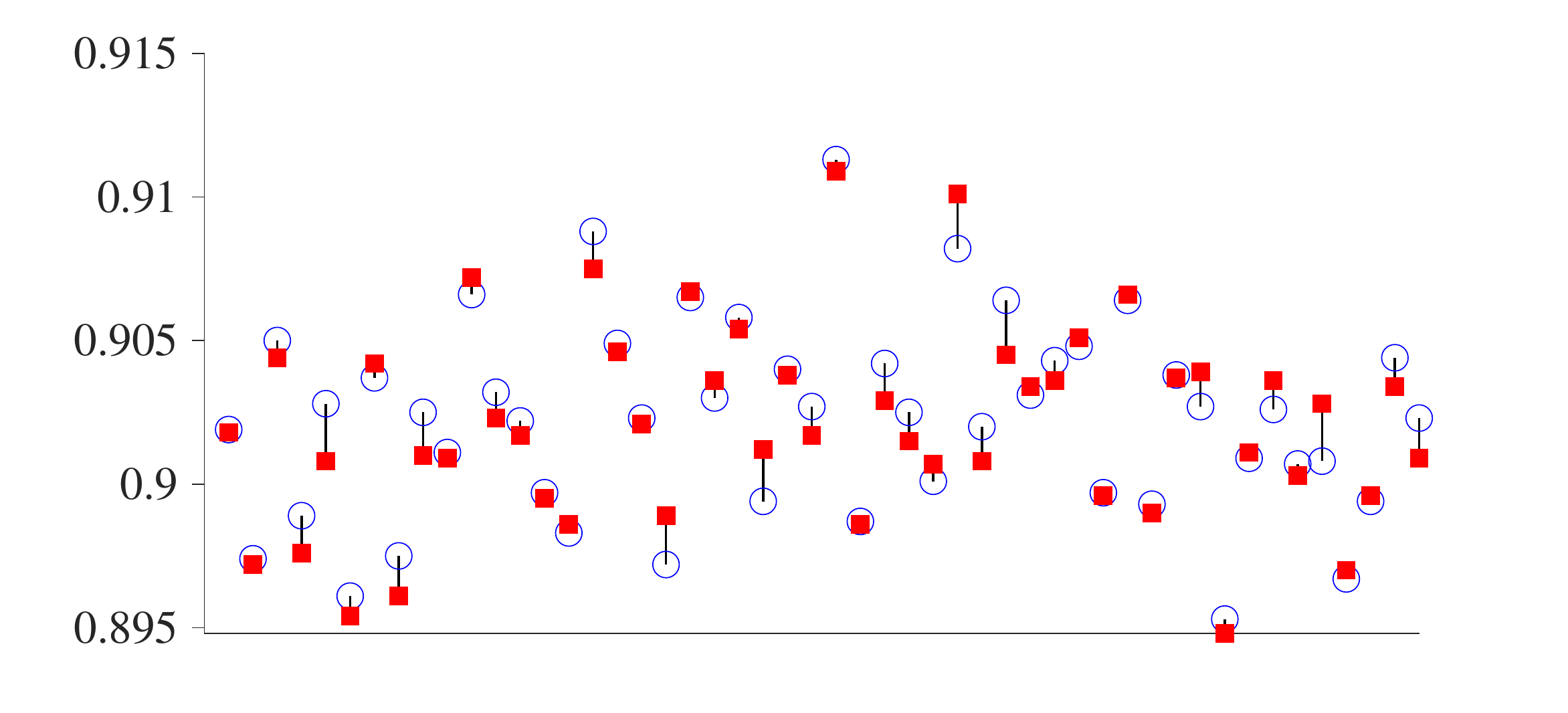}}
	\subfloat[{[DC-DE, $\M_{0.5}$, $m=270$]}]{\label{fig: er_dc_de_50_30_270}\includegraphics[width=0.33\textwidth]{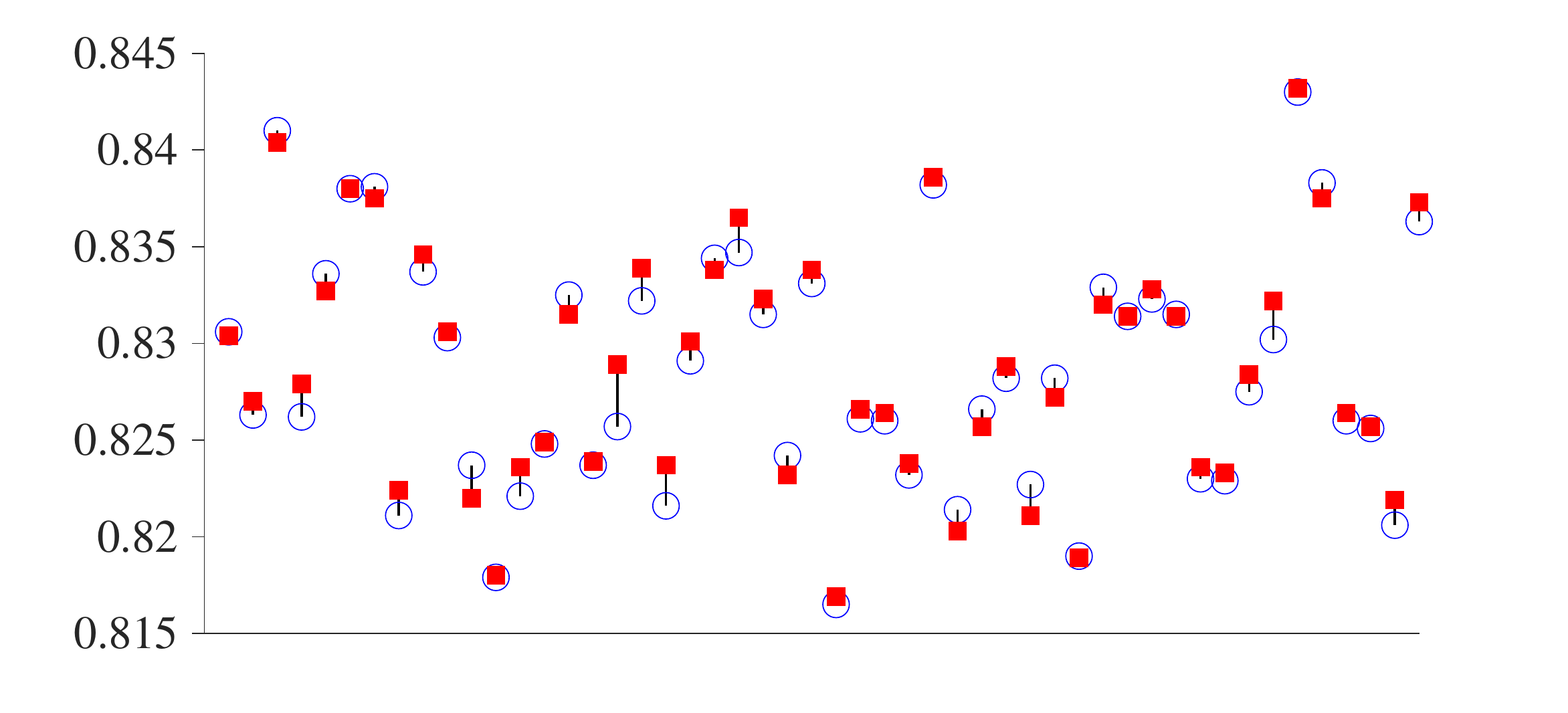}} 
	
	\subfloat[{[DE-DC, $\M_{0.5}$, $m=270$]}]{\label{fig: er_de_dc_50_30_270}\includegraphics[width=0.33\textwidth]{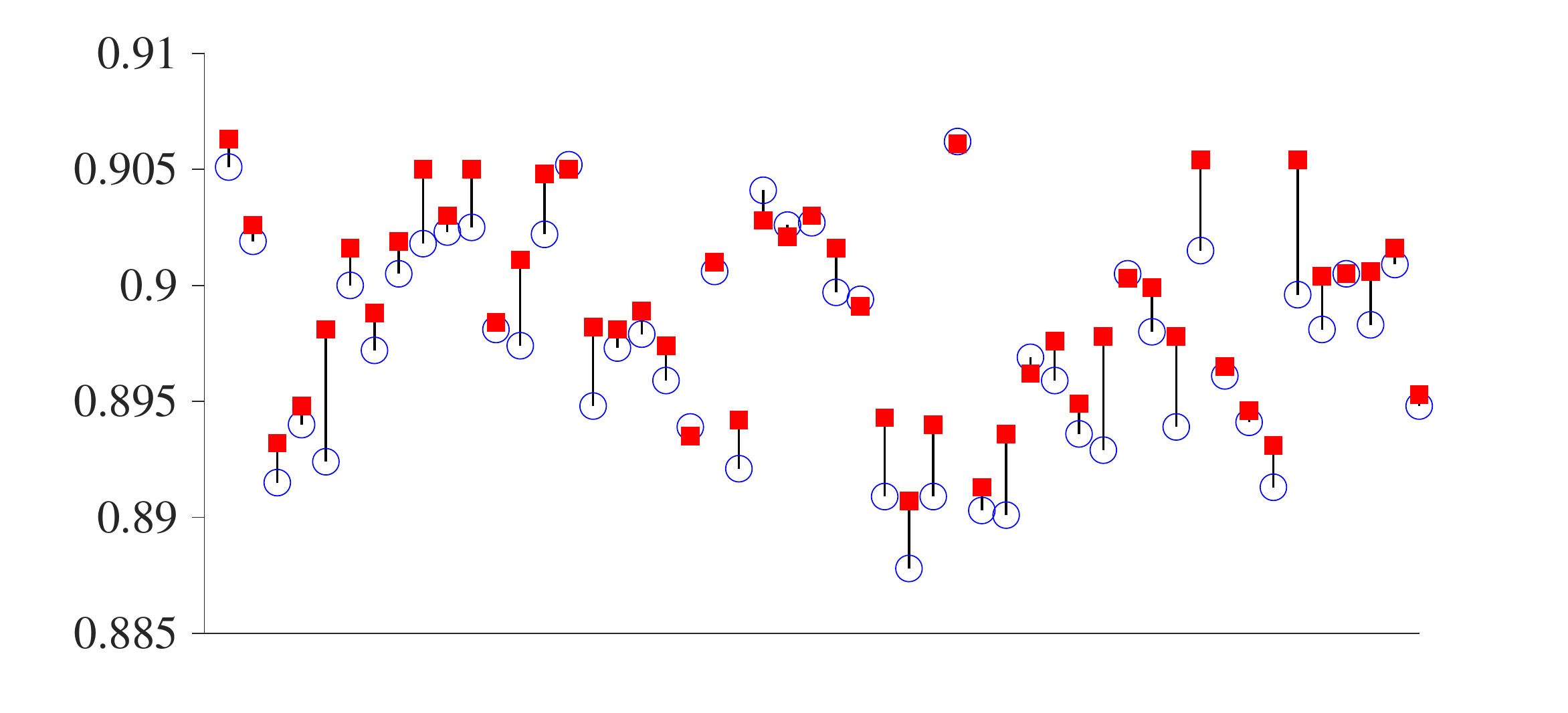}}	
	\subfloat[{[DC-DE, $\M_{\infty}$, $m=270$]}]{\label{fig: er_dc_de_uni_30_270}\includegraphics[width=0.33\textwidth]{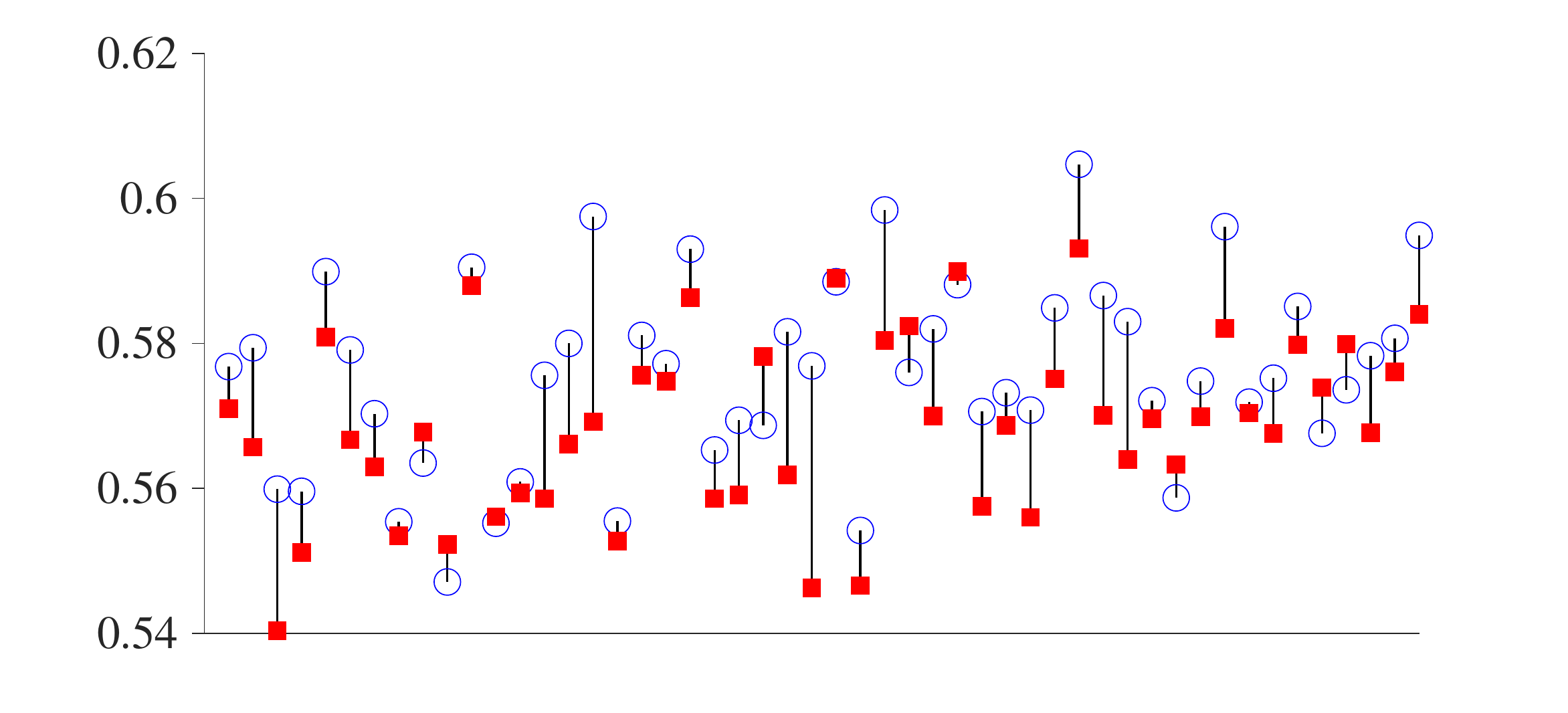}} 
	\subfloat[{[DE-DC, $\M_{\infty}$, $m=270$]}]{\label{fig: er_de_dc_uni_30_270}\includegraphics[width=0.33\textwidth]{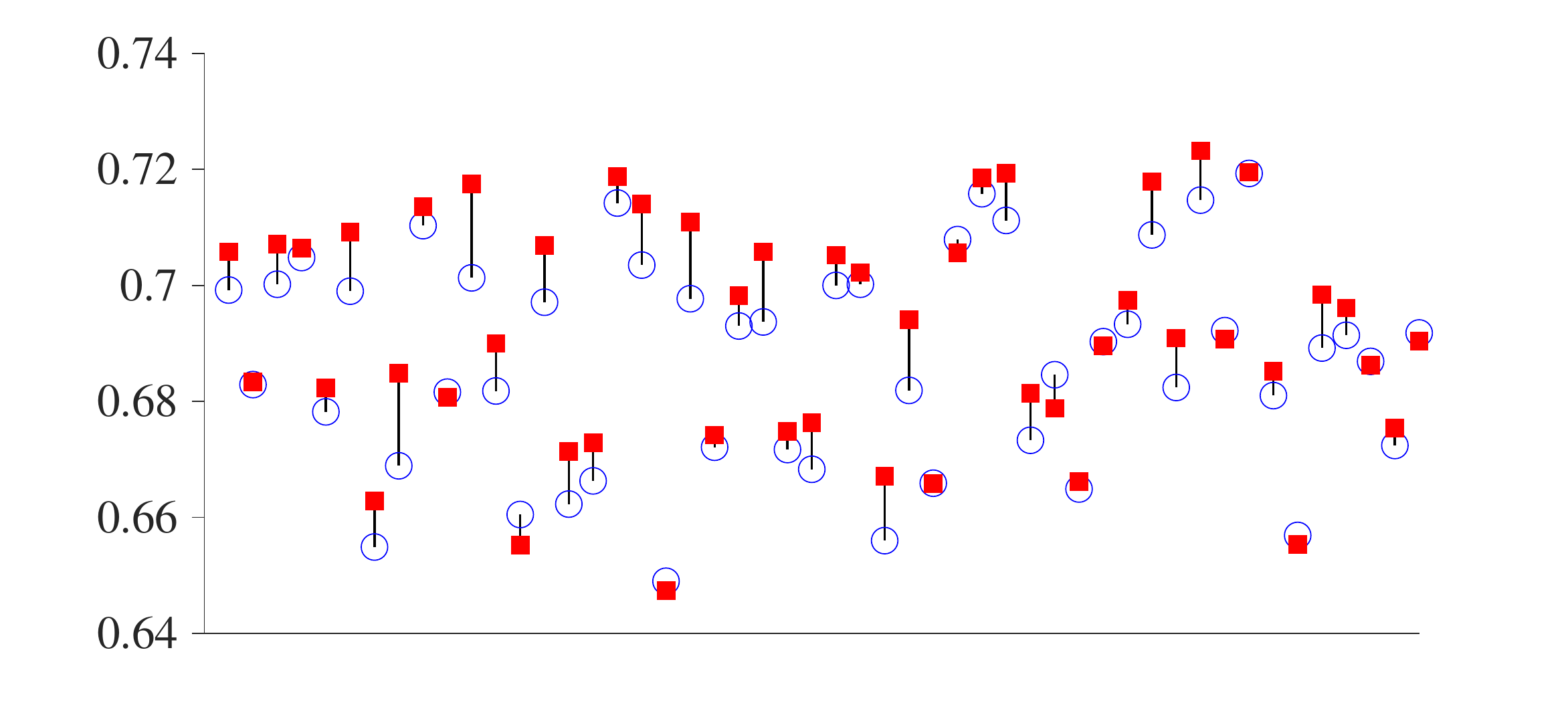}}
	\caption{Each subgraph shows the results of 50 testing queries on the Erdős-Rényi graph with $K=30$. For each query, we report the quality (i.e. performance ratio) of the predictions made based on the initial weight (before optimization) and final weight (after optimization).}
	\label{fig: er}
	\vspace{-5mm}
\end{figure}

The results on the Erdős-Rényi graph are given in Table \ref{table: er} and Figure \ref{fig: er}. Similar to the observations on the Kronecker graph, it is clear that Social-Inverse performs better when $K$ becomes large or $\M_{em}$ is closer to $\M_{true}$. One interesting observation is that the methods based on neural networks (i.e., GNN and DSPN) can hardly have any optimization effect except for DE-DE, and NB only works when the source task is the same as the target task, which suggests that they cannot mitigate the task migrations on the Erdős-Rényi graph. In addition, according to Figure \ref{fig: er}, the efficacy of the samples is evident for solving DC-DE with $\M_{\infty}$,  but it is less significant in other cases. 
\begin{table}[h]
	\renewcommand{\arraystretch}{1.3} 
	\small
	\caption{\textbf{Results on Erdős-Rényi.} Each cell presents the mean of performance ratio with std.}
	\centering
	\label{table: er}
	\begin{tabular}{  @{}  c   @{\hspace{4mm}}   c @{\hspace{2mm}}  c @{\hspace{1mm}}    c  @{\hspace{1mm}}  c  @{\hspace{1mm}} c  @{\hspace{4mm}}   c @{\hspace{1mm}} c @{\hspace{2mm}} c    @{}  }
		\toprule
		&\multicolumn{5}{@{} c @{\hspace{0mm}} }{\textbf{Social-Inverse}} &  \multicolumn{3}{c}{ \multirow{1}{*}{\textbf{Other Methods}} } \\
		&  $K$ &  $5$ &  $15$   & $30$ &  $60$ &&&GNN	\\
		\midrule
		\midrule
		\multirow{4}{*}{\textbf{DE-DE}} & $\M_{\infty}$  &	 0.421{\tiny (0.009)} &    0.551{\tiny (0.28)} 	  &  0.579{\tiny (0.023)} 	&   0.587{\tiny (0.010)}	   & DSPN	& NB &    0.339 (11)	\\
		& $\M_{1}$ &  0.715{\tiny (0.011)}  &   0.746{\tiny (0.011)} & 	 0.750{\tiny (0.005)} 		&   0.747{\tiny (0.003)}    &  0.378{\tiny (0.177)}     &  0.496{\tiny (0.005)}      &   0.381 (17)\\
		& $\M_{0.5}$ &  0.784{\tiny (0.013)}  &   0.816{\tiny (0.006)} & 	 0.828{\tiny (0.004)} 	&	 0.830{\tiny (0.006)}    & Random	  &    {HD} &      0.317 (13)   \\
		& $\M_{0.1}$ &   0.806{\tiny (0.006)} &    0.825{\tiny (0.006)} 	  &  0.828{\tiny (0.007)} 	 &	0.833{\tiny (0.006)}      &   0.063{\tiny (0.002)}      &      0.317{\tiny (0.006)}  &    0.397 (43)   \\
		
		\cmidrule{1-9}
		
		\multirow{4}{*}{\textbf{DC-DE}} & $\M_{\infty}$  &	 0.424{\tiny (0.017)} &    0.541{\tiny (0.010)} 	  &  0.585{\tiny (0.018)} 	   	   & 0.577{\tiny (0.007)} 	   	   &  	DSPN	& NB & 0.028 (11)	\\
		& $\M_{1}$ &  0.724{\tiny (0.014)}  &   0.744{\tiny (0.005)} & 	 0.752{\tiny (0.003)} 		     &  0.754{\tiny (0.003)} 		     &    0.060{\tiny (0.028)}     &   0.077{\tiny (0.005)}   &  0.068 (13)    \\
		& $\M_{0.5}$ &  0.784{\tiny (0.005)}  &   0.825{\tiny (0.005)} & 	 0.830{\tiny (0.005)} 	&	  0.830{\tiny (0.005)}     & Random   &    {HD}       &     0.057 (43)   \\
		& $\M_{0.1}$ &   0.798{\tiny (0.004)}  &   0.829{\tiny (0.002)} & 	 0.833{\tiny (0.004)} 	 &	0.833{\tiny (0.004)}	     &   0.063{\tiny (0.002)}      & 0.317{\tiny (0.006)}        & 0.035 (17)      \\
		
		\cmidrule{1-9}
		
		\multirow{4}{*}{\textbf{DE-DC}} & $\M_{\infty}$  &	 0.263{\tiny (0.012)} &    0.526{\tiny (0.020)} 	  &  0.677{\tiny (0.016)} 	&  0.750{\tiny (0.010)} 	   	     & DSPN	& NB  &  0.049 (11)	\\
		& $\M_{1}$ &  0.725{\tiny (0.018)}  &   0.773{\tiny (0.016)} & 	 0.796{\tiny (0.010)} 	& 	 0.800{\tiny (0.009)} 		      &    0.042{\tiny (0.007)}  &   0.041{\tiny (0.002)}     & 0.053 (13)	  \\
		& $\M_{0.5}$ &  0.823{\tiny (0.006)}  &   0.879{\tiny (0.002)} & 	 0.900{\tiny (0.002)} 		     & 0.892{\tiny (0.002)}  &   Random  &    {HD}   &  0.054 (43)  \\
		& $\M_{0.1}$ &   0.837{\tiny (0.002)} &    0.886{\tiny (0.005)} 	  &  0.902{\tiny (0.003)} 	 &  0.895{\tiny (0.003)}	     &  0.043{\tiny (0.002)}   & 0.094{\tiny (0.006)}       &     0.051 (17)  \\
		
		\cmidrule{1-9}

	\multirow{4}{*}{\textbf{DC-DC}} & $\M_{\infty}$  &	 0.273{\tiny (0.019)} &    0.528{\tiny (0.022)} 	  &  0.682{\tiny (0.019)} 	&  0.748{\tiny (0.019)}    	      & DSPN	& NB   &  0.065 (11)	\\
	& $\M_{1}$ &  0.741{\tiny (0.005)}  &   0.796{\tiny (0.013)} & 	 0.807{\tiny (0.009)}  & 	 0.807{\tiny (0.011)} 		     &    0.083{\tiny (0.013)}   &     0.576{\tiny (0.028)}  &  0.071 (13)    \\
	& $\M_{0.5}$ &  0.820{\tiny (0.009)}  &   0.879{\tiny (0.007)} & 	 0.901{\tiny (0.006)} 	   & 0.895{\tiny (0.005)} 	   & Random   &    {HD}      &    0.073 (43)     \\
	& $\M_{0.1}$ &   0.830{\tiny (0.008)} &    0.887{\tiny (0.009)} 	  &  0.902{\tiny (0.001)} 	 	&  0.895{\tiny (0.002)} 	     & 0.043{\tiny (0.002)}     &   0.094{\tiny (0.006)}        &    0.059 (17)   \\
		
		\bottomrule
	\end{tabular}
\end{table}

\newpage
\subsection{Results on Higgs and Hep}

\begin{figure}[t]
	\centering
	\subfloat{\includegraphics[width=0.4\textwidth]{images/legend.pdf}}
	\addtocounter{subfigure}{-1}
	
	\subfloat[{[Higgs, DE-DC, $\M_{\infty}$, 270]}]{\label{fig: higgs_de_dc_uni_5_270}\includegraphics[width=0.33\textwidth]{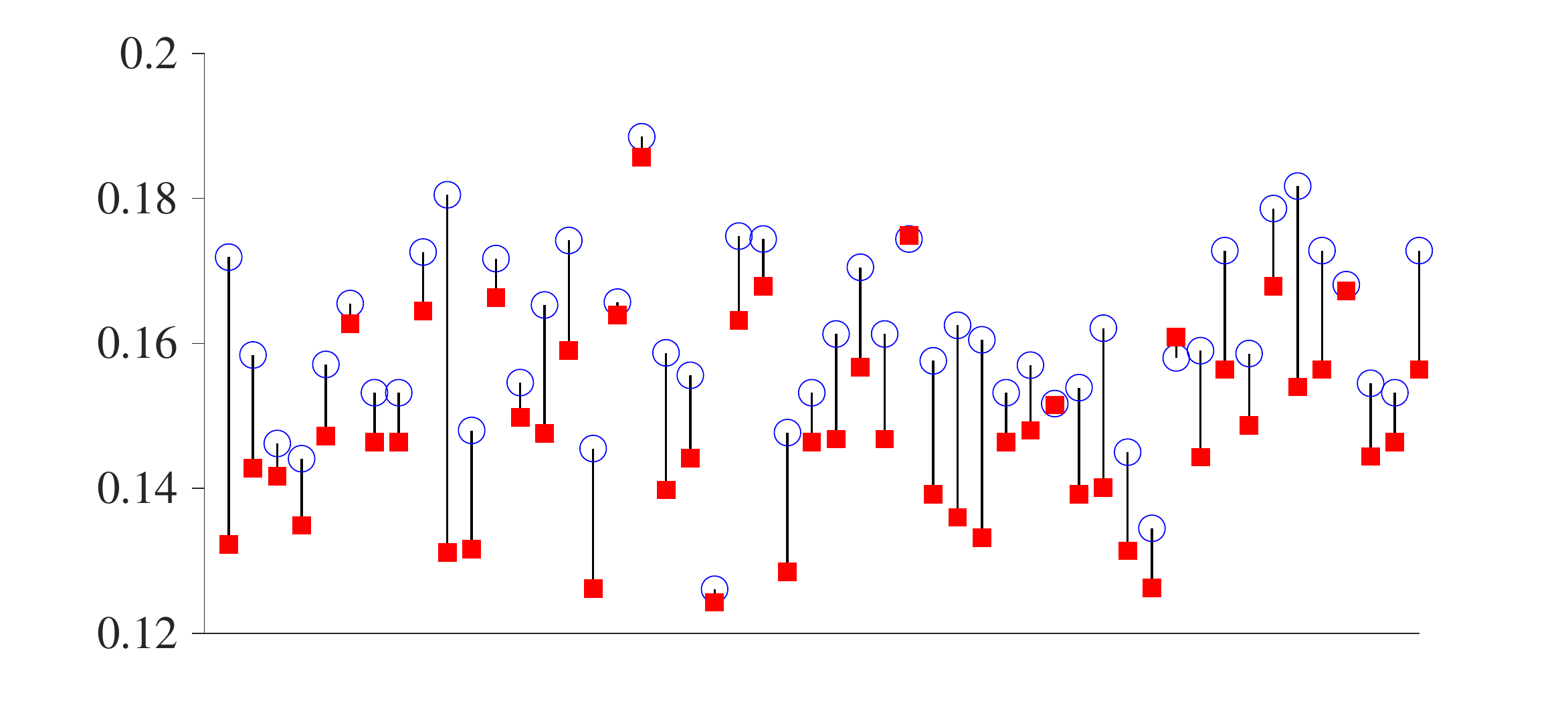}} 
	\subfloat[{[Higgs, DE-DC, $\M_{\infty}$, 1350]}]{\label{fig: higgs_de_dc_uni_5_1350}\includegraphics[width=0.33\textwidth]{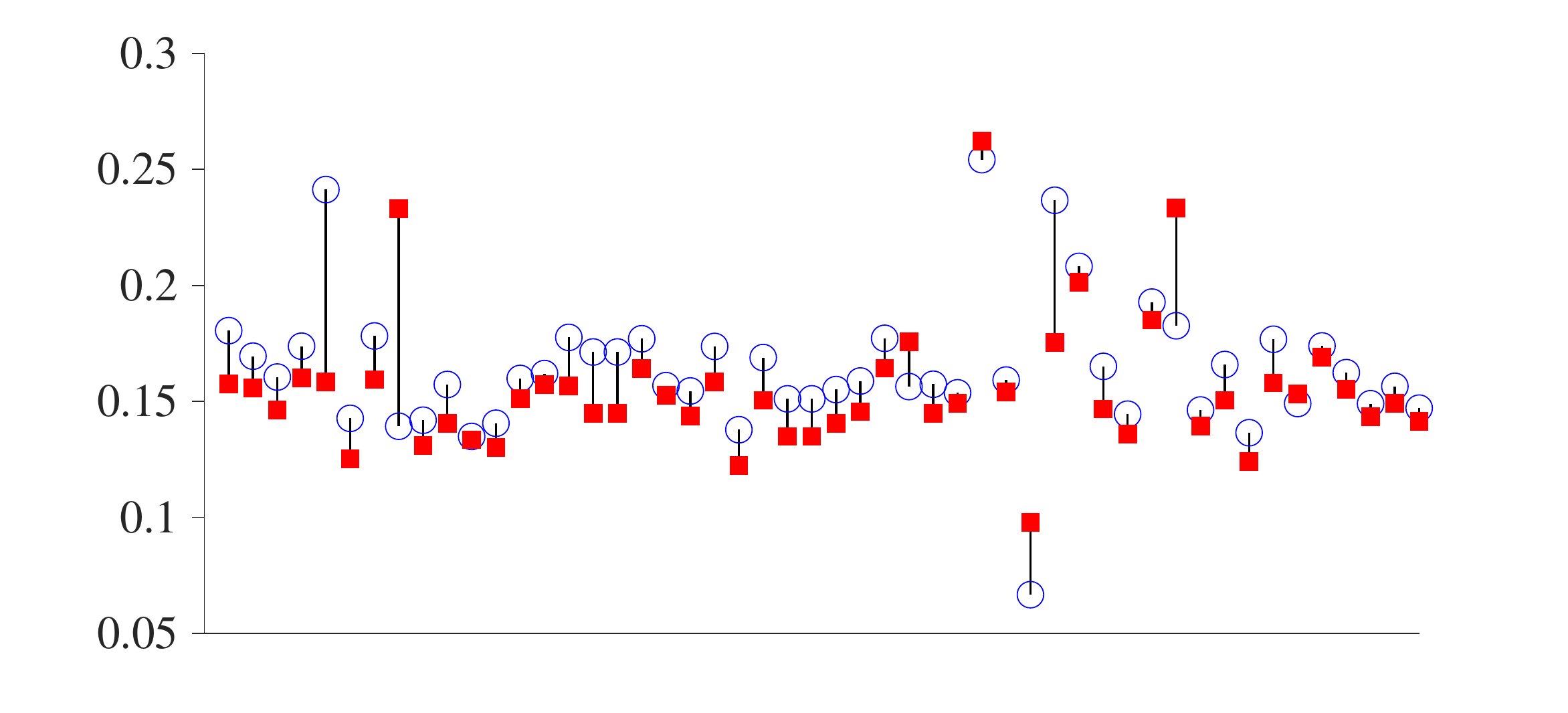}}
	\subfloat[{[Higgs, DC-DE, $\M_{\infty}$, 270]}]{\label{fig: higgs_dc_de_uni_5_270}\includegraphics[width=0.33\textwidth]{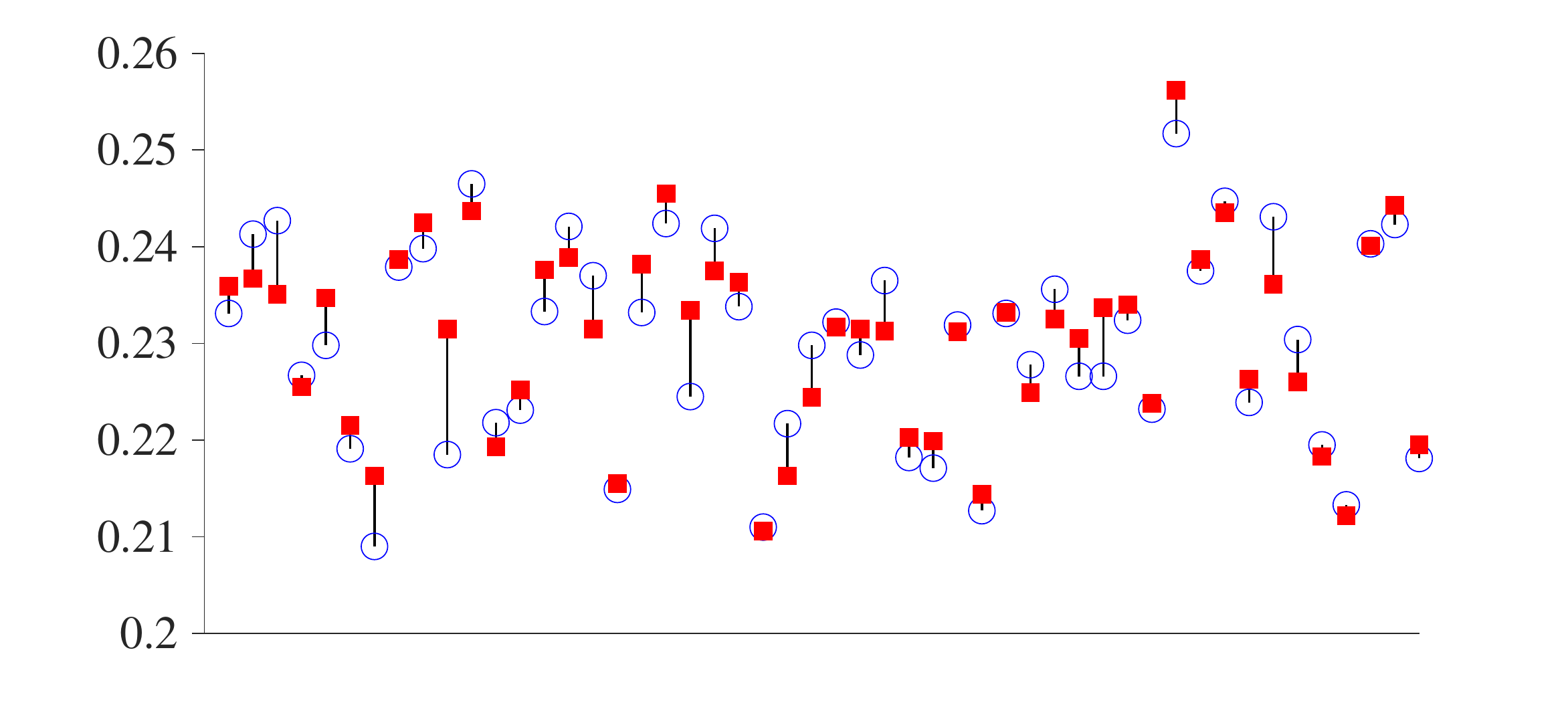}} 
	
	\subfloat[{[Higgs, DC-DE, $\M_{\infty}$, 1350]}]{\label{fig: higgs_dc_de_uni_5_1350}\includegraphics[width=0.33\textwidth]{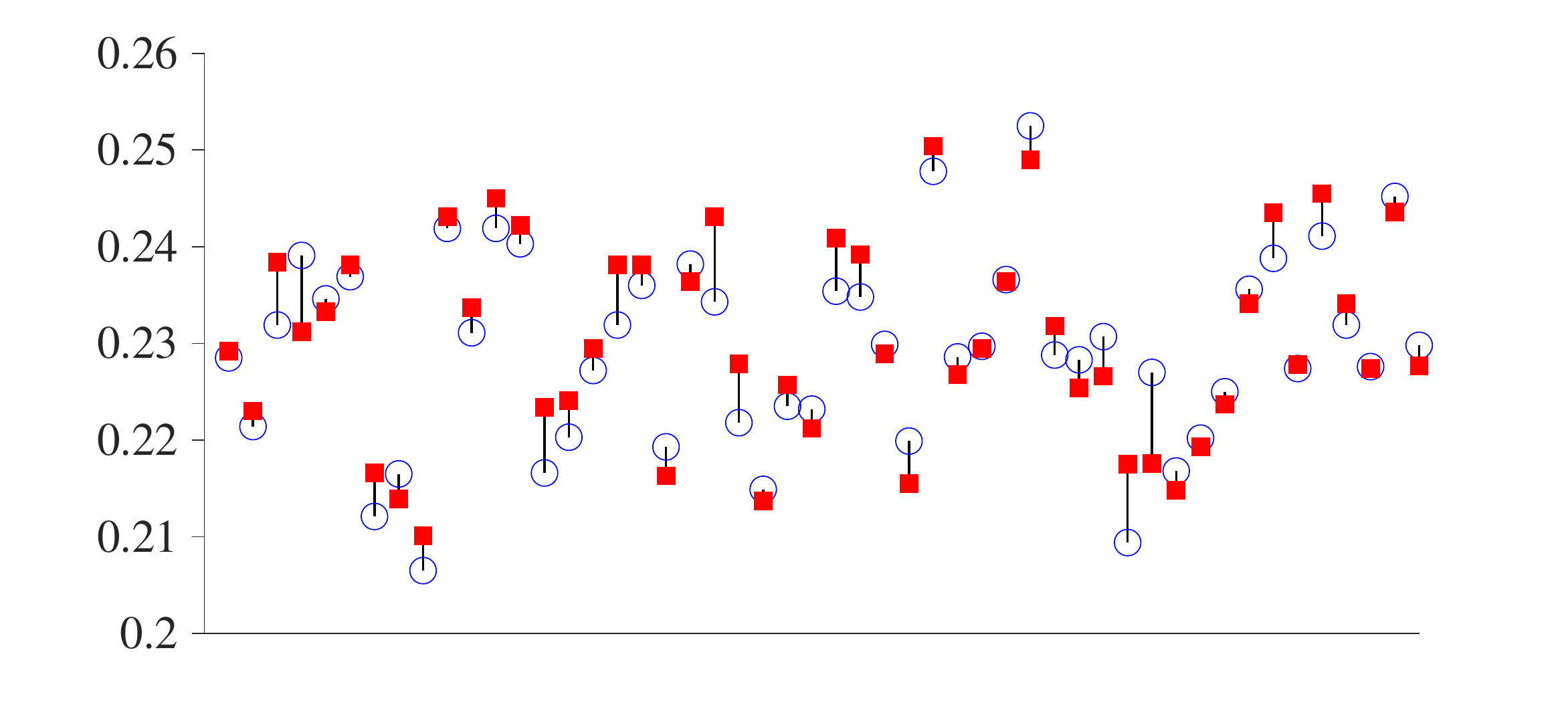}} 
	\subfloat[{[Hep, DE-DC, $\M_{0.1}$, 1350]}]{\label{fig: hep_de_dc_10_5_1350}\includegraphics[width=0.33\textwidth]{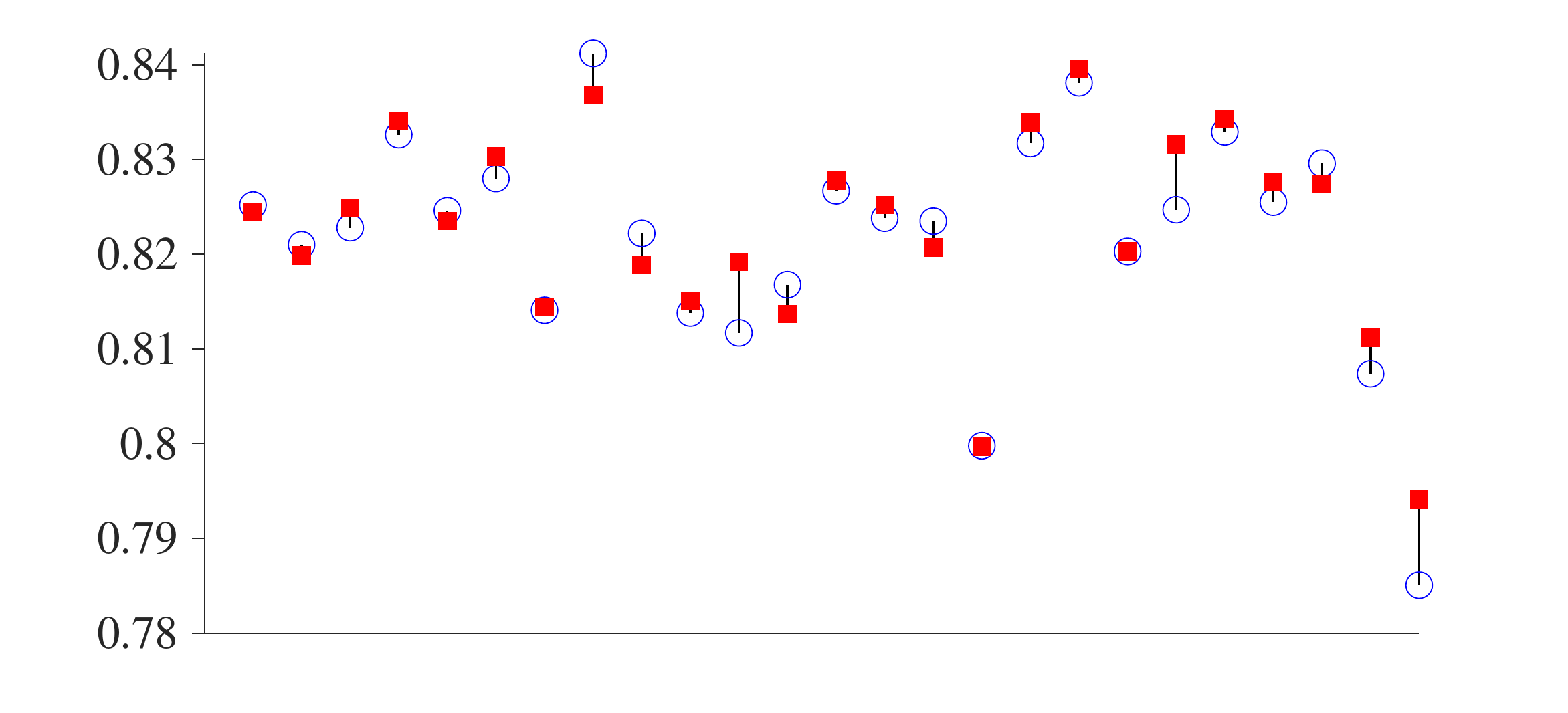}}
	\subfloat[{[Hep, DE-DC, $\M_{0.5}$, 1350]}]{\label{fig: hep_de_dc_50_5_1350}\includegraphics[width=0.33\textwidth]{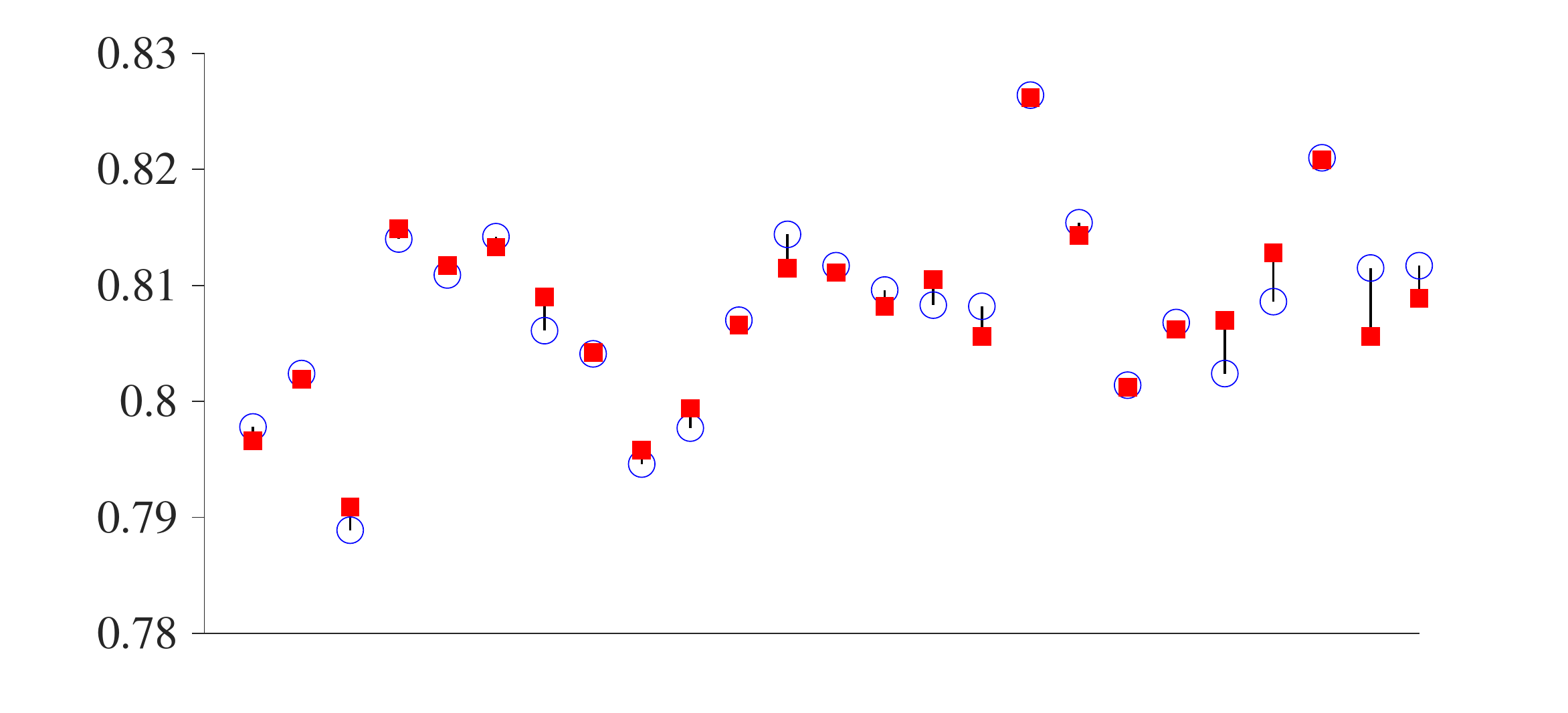}}
	
	\subfloat[{[Hep, DE-DC, $\M_{1}$, 1350]}]{\label{fig: hep_de_dc_100_5_1350}\includegraphics[width=0.33\textwidth]{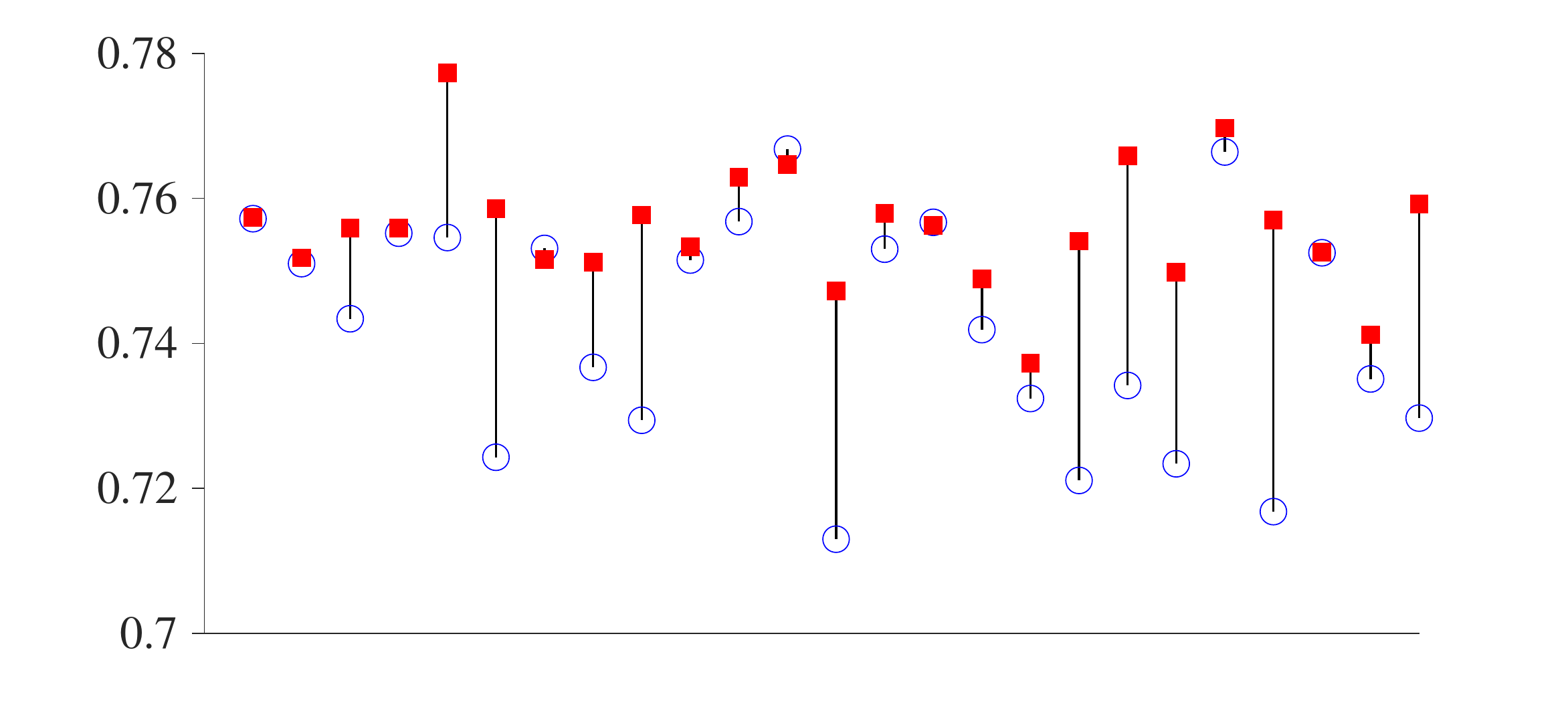}}
	\subfloat[{[Hep, DE-DC, $\M_{\infty}$, 1350]}]{\label{fig: hep_de_dc_uni_5_1350}\includegraphics[width=0.33\textwidth]{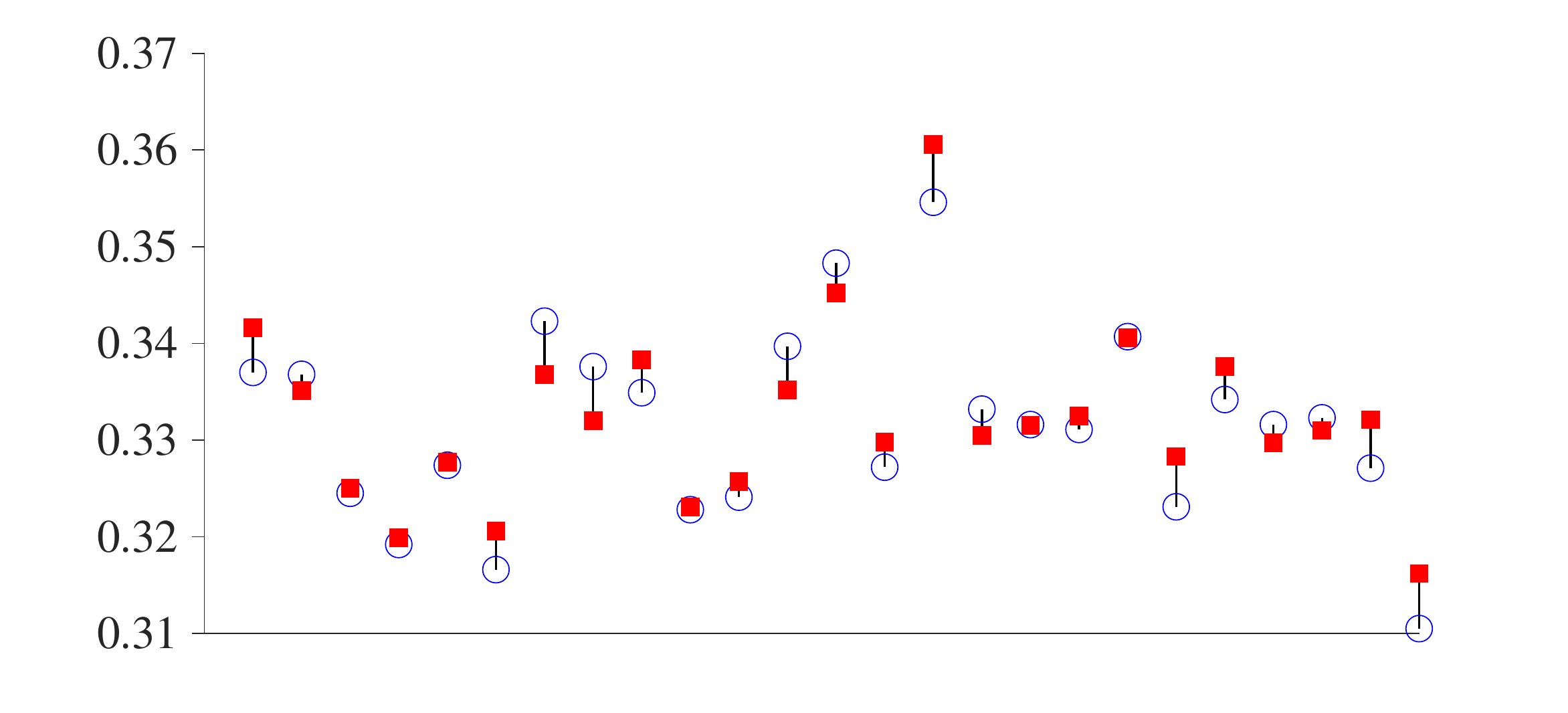}}
	\caption{Each subgraph is labeled by [graph, source-target, empirical distribution, training size] showing the results under one setting with $K=5$. For each testing query, we report the quality (i.e. performance ratio) of the predictions made based on the initial weight (before optimization) and final weight (after optimization).}
	\label{fig: higgs_hep}
\end{figure}

The results on Higgs and Hep can be found in Table \ref{table: higgs_hep} and Figure \ref{fig: higgs_hep}. GNN is not feasible for such large graphs due to the memory limit.  In addition to the aforementioned observations, we observe that none of the learning-based methods or HD can offer decent performance on large graphs like Higgs and Hep. Second, there is a clear gap between $\M_{\infty}$ and other empirical distributions, which suggests that prior knowledge is critical when the graph is large. According to Figure \ref{fig: higgs_hep}, the samples are clearly useful on Higgs for the case of DE-DC but less effective for DC-DE; on Hep, the task migration seems to be essential, and therefore, the efficacy of Social-Inverse depends entirely on the realizations but not on the samples at all. We believe that further investigation is needed to explore the conditions in which samples could be leveraged for solving DE-DC or DC-DE on large graphs.
\begin{table}[h]
	\renewcommand{\arraystretch}{1.3} 
	\small
	\caption{\textbf{Results on Higgs and Hep} Each cell presents the mean of performance ratio with std.}
	\centering
	\label{table: higgs_hep}
	\begin{tabular}{  @{}  c   @{\hspace{4mm}} c   @{\hspace{8mm}}   c @{\hspace{2mm}}  c @{\hspace{2mm}}    c  @{\hspace{2mm}} c @{\hspace{2mm}} c @{\hspace{8mm}}   c @{\hspace{2mm}} c @{}  }
		\toprule
		& &\multicolumn{5}{@{} c @{\hspace{0mm}} }{\textbf{Social-Inverse}} &  \multicolumn{2}{c}{ \multirow{2}{*}{\textbf{Other Methods}} } \\
		& &  $K$ &  $5$ &  $15$   & $30$ & $60$&   	\\
		\midrule
		\midrule
		\multirow{17}{*}{\textbf{Higgs}} & \multirow{4}{*}{\textbf{DE-DE}} & $\M_{\infty}$  &	 0.221{\tiny (0.008)} &    0.350{\tiny (0.008)} 	  &  0.417{\tiny (0.059)} 	   	  & 0.477{\tiny (0.010)}  & 	 DSPN	& NB 	\\
		& & $\M_{1}$ &  0.802{\tiny (0.007)}  &   0.844{\tiny (0.007)} & 	 0.862{\tiny (0.011)} 		&   0.876{\tiny (0.006)}     &  0.015{\tiny (0.013)}     &  0.122{\tiny (0.011)}       \\
		& & $\M_{0.5}$ &  0.882{\tiny (0.010)}  &   0.940{\tiny (0.003)} & 	 0.950{\tiny (0.002)} 		    &  0.956{\tiny (0.003)}  &    {HD} & Random  \\
		& & $\M_{0.1}$ &   0.919{\tiny (0.008)} &    0.950{\tiny (0.001)} 	  &  0.960{\tiny (0.003)} 	   	  & 0.960{\tiny (0.001)}      &       0.0257{\tiny (0.002)} &   0.002{\tiny (0.001)}   \\
		
		\cmidrule{2-9}
		
		&\multirow{4}{*}{\textbf{DC-DE}} & $\M_{\infty}$  &	 0.233{\tiny (0.013)} &    0.358{\tiny (0.007)} 	  &  0.419{\tiny (0.010)} 	   	   &   0.481{\tiny (0.008)}	 & DSPN	& NB 	\\
		& & $\M_{1}$ &  0.805{\tiny (0.011)}  &   0.860{\tiny (0.007)} & 	 0.872{\tiny (0.003)} 		     &   0.878{\tiny (0.005)}  &  0.004{\tiny (0.002)}     &   0.034{\tiny (0.001)}      \\
		& & $\M_{0.5}$ &  0.897{\tiny (0.010)}  &   0.939{\tiny (0.006)} & 	 0.951{\tiny (0.002)} 		     &   0.960{\tiny (0.001)}  &    {HD} & Random   \\
		& & $\M_{0.1}$ &   0.919{\tiny (0.006)}  &   0.952{\tiny (0.001)} & 	 0.960{\tiny (0.001)} 	 		     &  0.960{\tiny (0.001)} 	    & 0.0257{\tiny (0.002)} &   0.002{\tiny (0.001)}    \\
		
			\cmidrule{2-9}
		
		& \multirow{4}{*}{\textbf{DE-DC}} & $\M_{\infty}$  &	 0.139{\tiny (0.012)} &    0.173{\tiny (0.017)} 	  &  0.214{\tiny (0.010)} 	   	    &  0.228{\tiny (0.028)} & DSPN	& NB 	\\
		& & $\M_{1}$ &  0.809{\tiny (0.007)}  &   0.891{\tiny (0.010)} & 	 0.934{\tiny (0.009)} 		     & 0.951{\tiny (0.008)} 	  &    0.030{\tiny (0.015)}  &   0.019{\tiny (0.006)}     \\
		& & $\M_{0.5}$ &  0.874{\tiny (0.009)}  &   0.960{\tiny (0.007)} & 	 0.991{\tiny (0.006)} 		     &1.010{\tiny (0.006)} &    {HD} & Random    \\
		& & $\M_{0.1}$ &   0.877{\tiny (0.017)} &    0.975{\tiny (0.008)} 	  &  	 	1.010{\tiny (0.002)}      &   1.036{\tiny (0.006)}   & 0.075{\tiny (0.058)}  & 0.004{\tiny (0.004)}  \\
		
			\cmidrule{2-9}

		& \multirow{4}{*}{\textbf{DC-DC}} & $\M_{\infty}$  &	 0.141{\tiny (0.011)} &    0.184{\tiny (0.015)} 	  &  0.222{\tiny (0.018)}  &   0.279{\tiny (0.018)} & DSPN	& NB  	\\
		& & $\M_{1}$ &  0.804{\tiny (0.011)}  &   0.906{\tiny (0.009)} & 	 0.931{\tiny (0.007)} 		     &   0.968{\tiny (0.004)}  &   0.010{\tiny (0.007)}   &     0.110{\tiny (0.010)}   \\
		& & $\M_{0.5}$ &  0.879{\tiny (0.007)}  &   0.965{\tiny (0.007)} & 	 1.004{\tiny (0.003)} 	   &   1.026{\tiny (0.007)}  &    {HD} & Random    \\
		& & $\M_{0.1}$ &   0.897{\tiny (0.007)} &    0.977{\tiny (0.007)} 	  &  1.006{\tiny (0.005)} 	 		     &   1.030{\tiny (0.004)} 	   &   0.075{\tiny (0.058)}  & 0.004{\tiny (0.003)}    \\
		
		\midrule
		
		\multirow{17}{*}{\textbf{Hep}} &\multirow{4}{*}{\textbf{DE-DE}} & $\M_{\infty}$  &	 0.340{\tiny (0.008)} &    0.396{\tiny (0.007)} 	  &  0.400{\tiny (0.007)} 	   	  & 0.411{\tiny (0.006)}  & 	 DSPN	& NB    	\\
		& & $\M_{1}$ &  0.728{\tiny (0.005)}  &   0.763{\tiny (0.003)} & 	 0.773{\tiny (0.002)} 		&   0.774{\tiny (0.002)}     &  0.038{\tiny (0.020)}     &  0.344{\tiny (0.017)}          \\
		& & $\M_{0.5}$ &  0.797{\tiny (0.005)}  &   0.846{\tiny (0.004)} & 	 0.856{\tiny (0.001)} 		    &  0.862{\tiny (0.002)}  &         {HD} & Random    \\
		& & $\M_{0.1}$ &   0.844{\tiny (0.003)} &    0.889{\tiny (0.002)} 	  &  0.899{\tiny (0.002)} 	   	  & 0.901{\tiny (0.002)}      &      0.067{\tiny (0.001)} &   0.003{\tiny (0.001)}    \\
		
		\cmidrule{2-9}
		
		& \multirow{4}{*}{\textbf{DC-DE}} & $\M_{\infty}$  &	 0.341{\tiny (0.007)} &    0.400{\tiny (0.003)} 	  &  0.402{\tiny (0.004)} 	   	   &   0.404{\tiny (0.004)}	 & DSPN	& NB 	\\
		& & $\M_{1}$ &  0.727{\tiny (0.004)}  &   0.758{\tiny (0.002)} & 	 0.765{\tiny (0.008)} 		     &   0.774{\tiny (0.005)}  &  0.013{\tiny (0.005)}     &   0.036{\tiny (0.001)}       \\
		& & $\M_{0.5}$ &  0.795{\tiny (0.004)}  &   0.836{\tiny (0.007)} & 	 0.841{\tiny (0.006)} 		     &   0.845{\tiny (0.009)}  &         {HD} & Random   \\
		& & $\M_{0.1}$ &   0.837{\tiny (0.011)}  &   0.886{\tiny (0.005)} & 	 0.890{\tiny (0.003)} 	 		     &  0.897{\tiny (0.004)} 	        &  0.067{\tiny (0.001)} &   0.003{\tiny (0.001)}    \\
		
		\cmidrule{2-9}
		
		&\multirow{4}{*}{\textbf{DE-DC}} & $\M_{\infty}$  &	 0.329{\tiny (0.009)} &    0.517{\tiny (0.010)} 	  &  0.608{\tiny (0.010)} 	   	    &  0.684{\tiny (0.026)} & DSPN	& NB 	\\
		& & $\M_{1}$ &  0.725{\tiny (0.013)}  &   0.809{\tiny (0.008)} & 	 0.849{\tiny (0.008)} 		     & 0.895{\tiny (0.014)} 	  &    0.020{\tiny (0.009)}  &   0.005{\tiny (0.001)}    \\
		& & $\M_{0.5}$ &  0.782{\tiny (0.011)}  &   0.882{\tiny (0.008)} & 	 0.908{\tiny (0.011)} 		     &0.984{\tiny (0.005)}   &    {HD} & Random    \\
		& & $\M_{0.1}$ &   0.785{\tiny (0.008)} &    0.889{\tiny (0.004)} 	  &  	 	0.923{\tiny (0.011)}      &   0.990{\tiny (0.007)}   & 0.046{\tiny (0.002)}  & 0.004{\tiny (0.002)}   \\
		
		\cmidrule{2-9}

		& \multirow{4}{*}{\textbf{DC-DC}} & $\M_{\infty}$  &	 0.321{\tiny (0.004)} &    0.534{\tiny (0.016)} 	  &  0.612{\tiny (0.009)}  &   0.713{\tiny (0.011)} & DSPN	& NB	\\
		& & $\M_{1}$ &  0.711{\tiny (0.011)}  &   0.800{\tiny (0.025)} & 	 0.843{\tiny (0.012)} 		     &   0.911{\tiny (0.011)}  &   0.023{\tiny (0.011)}   &     0.046{\tiny (0.002)}   \\
		& & $\M_{0.5}$ &  0.776{\tiny (0.013)}  &   0.858{\tiny (0.009)} & 	 0.898{\tiny (0.003)} 	   &   0.926{\tiny (0.012)}   &    {HD} & Random    \\
		& & $\M_{0.1}$ &   0.789{\tiny (0.014)} &    0.865{\tiny (0.015)} 	  &  0.935{\tiny (0.013)} 	 		     &   0.932{\tiny (0.014)} 	   &   0.046{\tiny (0.002)}  & 0.004{\tiny (0.002)}   \\
		
		\bottomrule
	\end{tabular}
\end{table}

\newpage
~
\newpage
\section{Related works}
\label{sec: appendix_related}
\textbf{Social contagion management.} The study of lunching information cascades for maximizing the cascade influence is popularized by the seminal work of Kempe \textit{et al.} \cite{kempe2003maximizing}, following which various social contagion management tasks have been proposed and studied, such as scalable influence maximization \cite{chen2009efficient,chen2010scalable,tang2014influence}, competitive influence maximization \cite{budak2011limiting,tong2020stratlearner}, and epidemics source detection \cite{dawkins2021diffusion,dong2013rooting,shah2011rumors,yu2018rumor}. The early studies focus on algorithmic developments based on the setting that the diffusion models are presumably known, where submodular optimization \cite{kempe2003maximizing,budak2011limiting} and stochastic optimization \cite{borgs2014maximizing} are critical in solving these problems. Different from these works, our paper considers the case where the contagion management task is defined over unknown models. To better support real-world applications, another branch of research seeks to develop methodologies for learning diffusion model from data, with a particular focus on solving decision-making tasks based on the learned model (e.g., \cite{goyal2010learning,fang2013predicting,saito2010selecting,bonchi2011influence,kalimeris2018learning,he2016learning,du2014influence}). 

\textbf{Learn-and-optimize methods.} When presented with unknown decision-making models, one natural choice is the learn-and-optimize process.  In the initial studies, the learning stage and the optimization stage are treated separately \cite{wang2006cope,fang2016deploying,mukhopadhyay2017prioritized,xue2016avicaching}. Such methods are justified when the hidden distribution $\M_{true}$ can be perfectly recovered and the optimization problem can be solved effectively; however, learning errors are not avoidable, either in theory or in practice, thereby leading to unsatisfactory optimization performance (as observed in a few works \cite{beygelzimer2009offset,ford2015beware}). A growing body of research has been focused on the idea of learning-for-optimization aiming to design end-to-end training methods that can push the learning phase towards the final objectives \cite{donti2017task,wilder2019melding,bertsimas2020predictive}; such an idea can be dated back to Bengio's work twenty years ago \cite{bengio1997using}. While these approaches focus on the optimization effects, they still attempt to learn the decision model by using input-value pairs, which are $((X, Y), f_{\M_{true}}^T(X, Y))$ under our context. In contrast, we seek to solve the problem using data of query-decision pairs $(X, Y_{ \M_{true}, T, X}^{\dal})$. This means that our data has encoded an optimization process, which we believe is a novel way of thinking.

\textbf{Inverse engineering.} Our work is also related to the study of inverse engineering. One branch of such study focuses on finding a minimal adjustment of the objective functions such that a given set of feasible solutions becomes efficient \cite{roland2013inverse,zhang1999solution,chan2021inverse}; another relevant topic is to recover decision-making models or parameterized representations from optimal or suboptimal decisions \cite{tan2020learning,shahmoradi2021quantile,ahmadi2020inverse, dong2020wasserstein,dong2018generalized}. A key differentiating factor is that our work seeks to directly compute future decisions using query-decision pairs, rather than accurately inferring the latent diffusion model. Under the setting of reverse decision making, one recent work \cite{tong2020stratlearner} explores the case of DC-DC, and later, such settings are generalized to abstract stochastic combinatorial optimizations \cite{tong2021usco}; comparing to these works, we generalize the existing settings and introduce the concept of task migration, which allows the source task to be different from the target task, thereby addressing the case where the data availability of the target task is limited; in addition, this paper introduces a parameterized margin together with new analysis and training schemes to resolve the task migrations, and therefore, our framework generalizes the existing ones (e.g., StratLearner \cite{tong2020stratlearner} and USCO-Solver \cite{tong2021usco}) and can be further fine-tuned.

\end{document}